\documentclass[table]{ai2style/ai2}
\usepackage{amssymb}
\usepackage{multirow}
\usepackage{bigdelim}
\usepackage{todonotes}
\usepackage{longtable}
\usepackage{tabularray}
\usepackage{wrapfig}
\usepackage[most]{tcolorbox} 
\usepackage{url}
\usepackage{xspace}
\usepackage{svg}
\usepackage[absolute]{textpos} %

\usepackage{fdsymbol}   %

\usepackage{csvsimple}

\newcommand{\taskitem}[4]{%
    \item \textbf{Task Name:} #1 \\
    \textbf{Task Description:} #2 \\
    \textbf{Language Description:} #3 \\
    \textbf{Task Progression Score Metrics:} #4
}

\usepackage[utf8]{inputenc} %
\usepackage[T1]{fontenc}    %
\usepackage{url}            %
\usepackage{booktabs}       %
\usepackage{amsfonts}       %
\usepackage{nicefrac}       %
\usepackage{microtype}      %
\usepackage[table]{xcolor}         %
\usepackage{amsmath}
\usepackage[most]{tcolorbox}
\usepackage{csquotes}

\usepackage{siunitx}
\usepackage{graphicx}
\usepackage{arydshln}
\usepackage{wrapfig}
\usepackage{enumitem}
\usepackage{soul} %

\usepackage{multirow}
\usepackage{xspace}
\usepackage{adjustbox}
\usepackage{pifont}
\usepackage{caption}
\usepackage{makecell}
\usepackage{subcaption}
\usepackage{bold-extra}
\usepackage{url}

\usepackage{booktabs,rotating,xcolor}

\usepackage{hyperref}
\definecolor{linkcolor}{RGB}{0, 0, 128}
\hypersetup{
     colorlinks   = true,
     citecolor    = linkcolor,
     linkcolor    = linkcolor,
     urlcolor     = linkcolor,
}
\usepackage{pifont}%
\usepackage{listings}

\setlist[itemize]{leftmargin=*,itemsep=0em,parsep=0.3em,topsep=0.3em}

\definecolor{maroon}{HTML}{F26035}
\definecolor{yellow}{HTML}{FDBC42}
\definecolor{lavender}{HTML}{734f96}
\definecolor{darkergrey}{HTML}{444444}
\definecolor{midgrey}{HTML}{e6eded}
\definecolor{ai2pink}{HTML}{f0529c}%
\definecolor{ai2midpink}{HTML}{fad3e5}
\definecolor{ai2lightpink}{HTML}{fbecf3}
\definecolor{ai2midwhite}{HTML}{f2e5d9}
\definecolor{ai2offwhite}{HTML}{fbf4ee}
\definecolor{ai2green}{HTML}{0fcb8c}
\definecolor{ai2lightgreen}{HTML}{e7f9f3}
\definecolor{ai2darkgreen}{HTML}{105257}
\definecolor{ai2purple}{HTML}{B932EB}
\definecolor{ai2lightpurple}{HTML}{f7e8fc}
\definecolor{neutralEight}{HTML}{343434}
\definecolor{neutralFive}{HTML}{838383}
\definecolor{neutralThree}{HTML}{bebebe}
\definecolor{neutralOne}{HTML}{dedede}
\definecolor{lightgrey}{HTML}{fafcfc}

\usepackage{tikz}

\definecolor{maroon}{HTML}{F26035}
\definecolor{yellow}{HTML}{FDBC42}
\definecolor{darkred}{RGB}{156, 39, 33}
\definecolor{darkblue}{RGB}{31, 90, 153}
\definecolor{forestgreen}{rgb}{0.13, 0.55, 0.13}

\newcommand{\arm}{\textsc{ARM}\xspace}
\newcommand{\arms}{\textsc{ARM}s\xspace}
\newcommand{\molmoact}{\textsc{MolmoAct}\xspace}
\newcommand{\molmoacto}{\textsc{MolmoAct-7B-O}\xspace}
\newcommand{\molmoactd}{\textsc{MolmoAct-7B-D}\xspace}
\newcommand{\molmoactdpre}{\textsc{MolmoAct-7B-D-Pretrain}\xspace}
\newcommand{\molmoactdata}{\textsc{MolmoAct Dataset}\xspace}

\newcommand{\groot}{\textsc{GR00T N1.5}\xspace}
\newcommand{\pizero}{$\pi_0$\xspace}
\newcommand{\pizerofast}{$\pi_0$-FAST\xspace}

\newcommand{\simpler}{SimplerEnv\xspace}
\newcommand{\libero}{LIBERO\xspace}

\newcommand{\huggingface}{\raisebox{-1.5pt}{\includegraphics[height=1.05em]{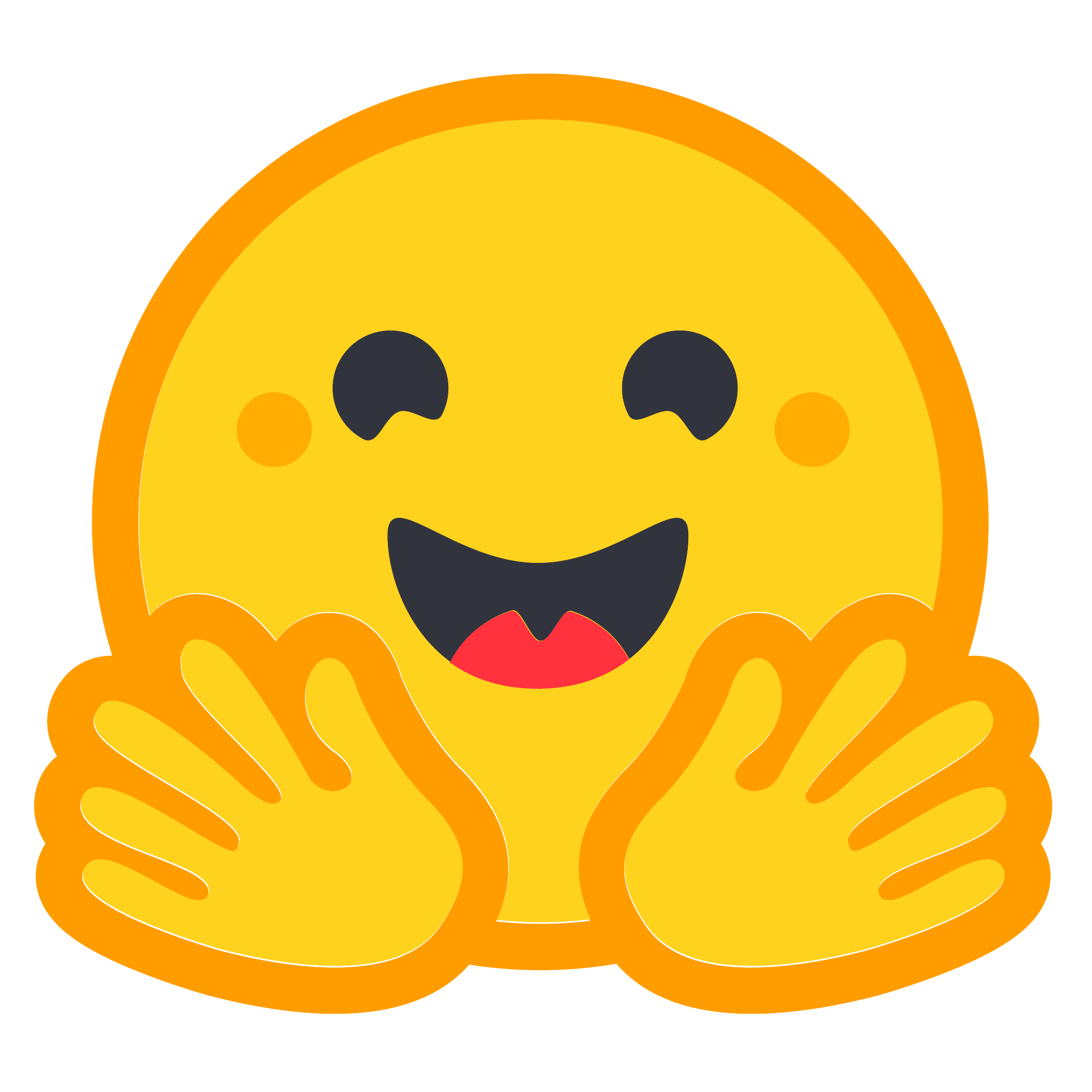}}\xspace}

\newcommand{\hfdataset}{\raisebox{-1.5pt}{\includegraphics[height=1.05em]{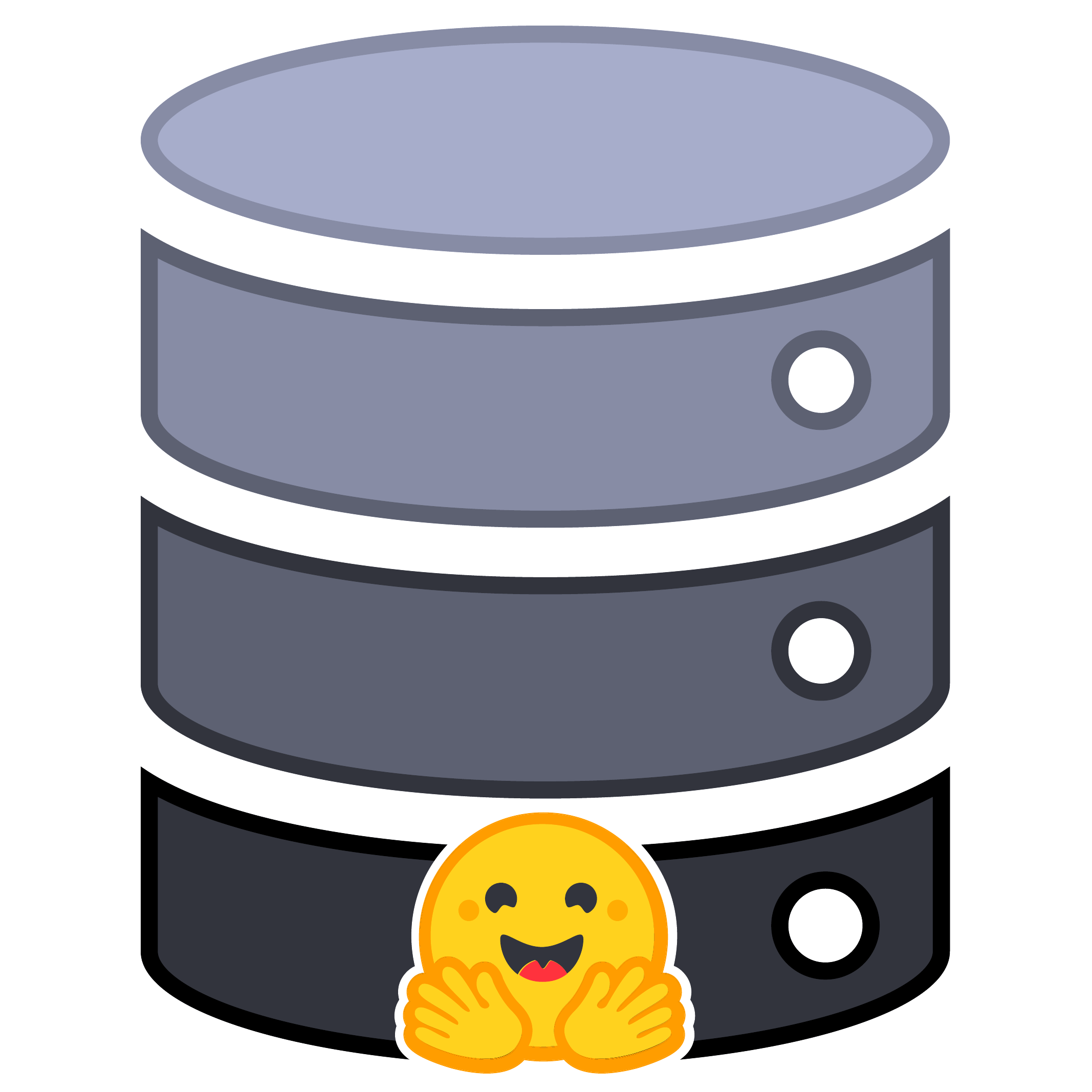}}\xspace}

\newcommand{\aitoo}{\raisebox{-1.5pt}{\includegraphics[height=1.05em]{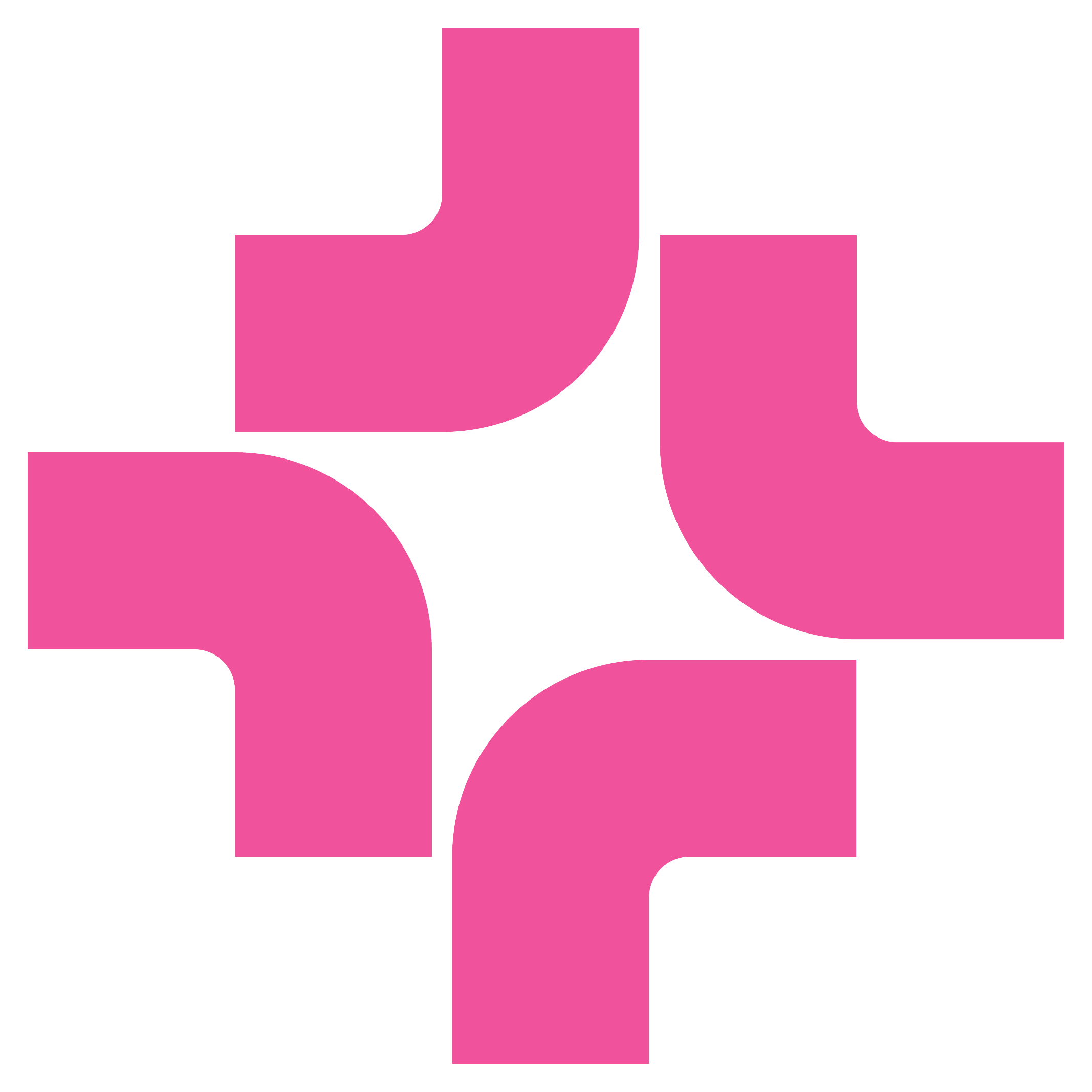}}\xspace}

\usepackage{setspace}
\usepackage{quoting}      % nice quote environment

\usepackage{nicematrix}
\newcolumntype{L}[1]{>{\raggedright\let\newline\\\arraybackslash\hspace{0pt}}m{#1}}
\newcolumntype{C}[1]{>{\centering\let\newline\\\arraybackslash\hspace{0pt}}m{#1}}
\newcolumntype{R}[1]{>{\raggedleft\let\newline\\\arraybackslash\hspace{0pt}}m{#1}}
\newcolumntype{P}[1]{>{\centering\let\newline\\\arraybackslash\columncolor{ai2lightpink}}m{#1}}
\newcolumntype{W}[1]{>{\columncolor{white}}c}  %
\title{MolmoAct\\{\fontsize{18pt}{12pt}\selectfont Action Reasoning Models that can Reason in Space}}

\newcommand{\core}{\textsuperscript{\textcolor{ai2pink}{\ding{170}}}}

\authorOne[1,2*]{Jason Lee\core}
\authorOne[1,2*]{Jiafei Duan\core}
\authorOne[1,2*]{Haoquan Fang\core}

\authorTwo[1]{Yuquan Deng\core}
\authorTwo[1,2]{Shuo Liu\core}
\authorTwo[2]{Boyang Li\core}
\authorTwo[2]{Bohan Fang\core}
\authorTwo[1,2]{Jieyu Zhang\core}
\authorTwo[1,2]{Yi Ru Wang\core}

\authorThree[1]{Sangho Lee}
\authorThree[1]{Winson Han}
\authorThree[1]{Wilbert Pumacay}
\authorThree[2]{Angelica Wu}
\authorThree[1]{Rose Hendrix\core}
\authorThree[1]{Karen Farley}
\authorThree[1]{Eli~VanderBilt}

\authorFour[1,2]{Ali Farhadi}
\authorFour[1,2]{Dieter Fox\core}
\authorFour[1,2]{Ranjay Krishna\core}

\affiliation[1]{Allen Institute for AI}
\affiliation[2]{University of Washington}

\contribution[*]{denotes equal contribution in no particular order. $\textcolor{ai2pink}{\varheartsuit}$ marks core contributors. See full author contributions \hyperref[sec:contrib]{here}.}

\abstract{
Reasoning is central to purposeful action, yet most robotic foundation models map perception and instructions directly to control, which limits adaptability, generalization, and semantic grounding. We introduce Action Reasoning Models (\arms), a class of robotic foundation models that integrates perception, planning, and control through a structured three-stage pipeline. Our model, \molmoact, encodes observations and instructions into depth-aware perception tokens, generates mid-level spatial plans as editable trajectory traces, and predicts precise low-level actions, enabling explainable and steerable behavior. \molmoactd achieves strong performance across simulation and real-world settings: $70.5\%$ zero-shot accuracy on \simpler Visual Matching tasks, surpassing closed-source \pizero and \groot; $86.6\%$ average success on \libero, including a $+6.3\%$ gain over ThinkAct on long-horizon tasks; and in real-world fine-tuning, $+10\%$ (single-arm) and $+22.7\%$ (bimanual) task progression over \pizerofast. It also outperforms baselines by $+23.3\%$ on out-of-distribution generalization and achieves top human-preference scores for open-ended instruction following and trajectory steering. Furthermore, we release, for the first time, the \molmoactdata—a mid-training robot dataset comprising over 10,000 high-quality robot trajectories across diverse scenarios and tasks. Training with this dataset yields an average 5.5\% improvement in general performance over the base model. We release all model weights, training code, \molmoactdata and our action reasoning dataset, establishing \molmoact as both a state-of-the-art robotics foundation model and an open blueprint for building ARMs that transform perception into purposeful action through grounded reasoning.
}

\metadata[\quad\huggingface MolmoAct:]{
\href{https://huggingface.co/allenai/MolmoAct-7B-D-Pretrain-0812}{\texttt{MolmoAct-7B-D-Pretrain-0812}} \quad \href{https://huggingface.co/allenai/MolmoAct-7B-D-0812}{\texttt{MolmoAct-7B-D-0812}} \quad
\href{https://huggingface.co/allenai/MolmoAct-7B-0-0812}{\texttt{MolmoAct-7B-O-0812}}}

\metadata[\vspace{.5em}\quad\hfdataset MolmoAct Data:]{
\href{https://huggingface.co/datasets/allenai/MolmoAct-Dataset}{\texttt{MolmoAct-Dataset}} \quad \href{https://huggingface.co/datasets/allenai/MolmoAct-Pretraining-Mixture}{\texttt{MolmoAct-Pretraining-Datasets}} \quad
\href{https://huggingface.co/datasets/allenai/MolmoAct-Midtraining-Mixture}{\texttt{MolmoAct-Midtraining-Datasets}}}

\metadata[\vspace{.5em}\quad\aitoo Blog:]{\href{https://allenai.org/blog/molmoact}{\texttt{allenai.org/blog/molmoact}}}

\begin{document}

\maketitle

\setcounter{tocdepth}{2}%

%%%%%%%%% MAIN PAPER %%%%%%%%%

\section{Introduction}
\label{sec:intro}

\begin{figure*}[t]  % spans both columns
  \centering
  \includegraphics[width=\textwidth]{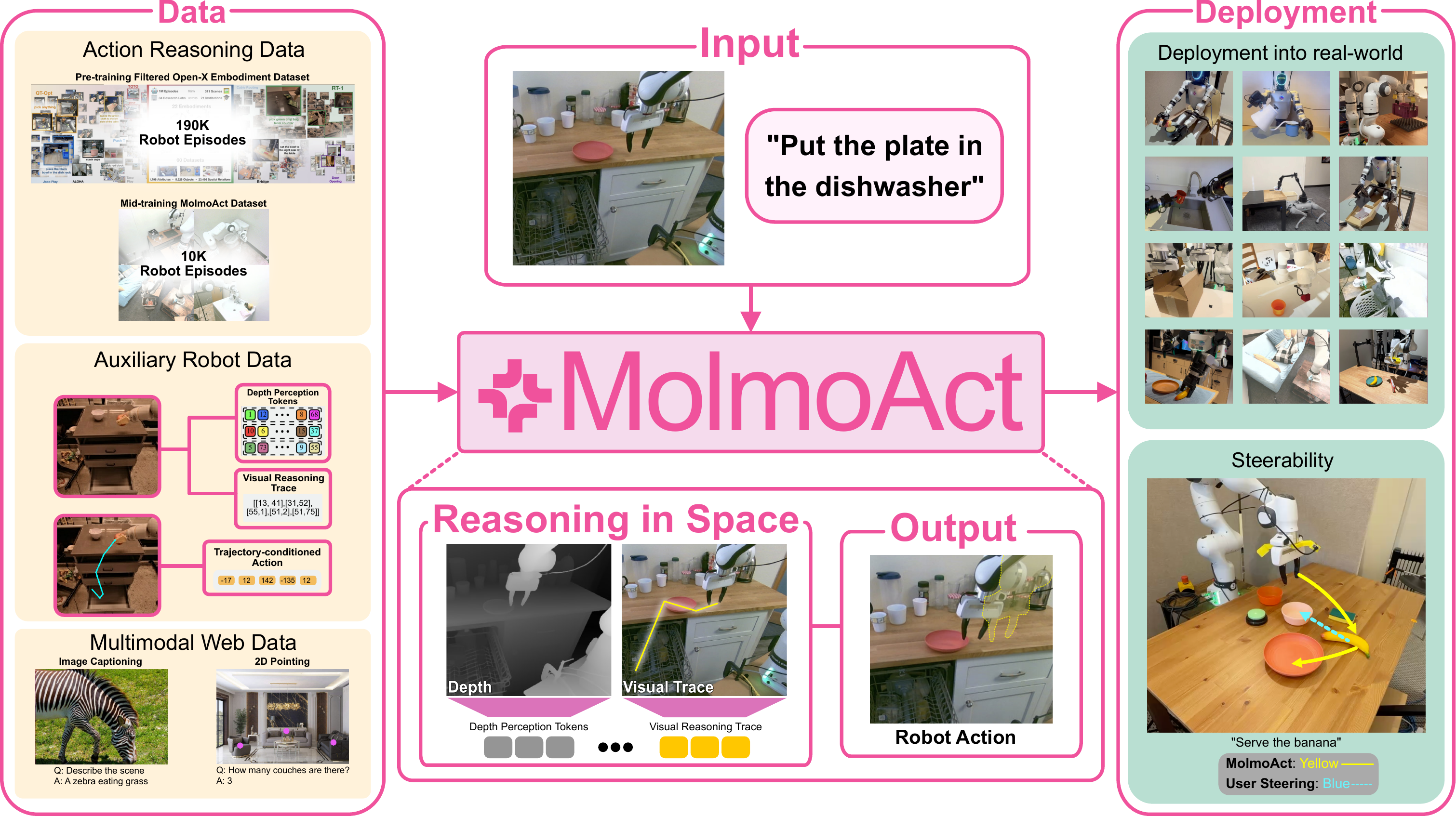}
  \caption{\textbf{Overview}. \molmoact is an open action reasoning model that, given a user’s language instruction, reasons in space and autoregressively predicts three structured reasoning chains: \textsc{Depth Perception Tokens} for sensing and reconstructing the 3D environment, \textsc{Visual Reasoning Trace Tokens} for representing its planned trajectory in the scene, and \textsc{Action Tokens} for generating the corresponding robot control commands. Each explainable reasoning chain can be independently decoded—yielding a depth map of the scene, a 2D trajectory overlay on the image plane, and executed actions in the physical world—providing explicit, spatially grounded reasoning at every stage.}
  \label{fig:fig1_overview}
\end{figure*}

\begin{quote}
  \textit{Thinking is embodied, spatial, and outside your head.}\\[2pt]
  --- Barbara Tversky, \emph{Emerita Professor of Psychology at Stanford}~\citep{tversky2025yourbody}
\end{quote}

Reasoning allows us to act with intention. Before reaching for a cup or moving through a room, we subconsciously weigh context, goals, and constraints—transforming perception into purpose. This process, grounded in our physical experience of the world, makes our actions coherent, adaptable, and explainable. For robots to operate with the same fluency, they must do more than map images and instructions to robot control. They must learn to reason.

In contrast to the rapid generalization gains seen in large language and vision models, progress in robotics has lagged behind \citep{duan2022survey,xu2024survey,firoozi2025foundation}. Vision-Language-Action (VLA) models \citep{black2410pi0,kim2024openvla,team2025gemini,nvidia2025gr00tn1openfoundation,yang2025magma} aim to bring similar capabilities to physical agents, but have yet to reach the same level of flexibility or robustness. Despite massive efforts in dataset collection and model scaling, today's VLAs remain brittle and opaque—struggling to transfer across tasks, scenes, or embodiments, and offering little insight into why a robot chose one action over another \citep{liu2025towards,pumacay2024colosseum}.

This gap stems not just from limited data, but from a lack of structure. While language and vision tasks benefit from abundant, loosely labeled web-scale data, robotics demands fine-grained, embodied interaction—data that is costly, ambiguous, and difficult to scale. Yet in parallel, language models have begun to shift away from brute-force scaling toward \textit{structured learning}: building intermediate representations that support reasoning, abstraction, and control \citep{wei2022chain,zelikman2022star,huang2022large}. We believe robotics can—and must—do the same.

We introduce \textsc{\molmoact} (\textbf{M}ultimodal \textbf{O}pen \textbf{L}anguage \textbf{Mo}del for \textbf{Act}ion), a family of completely open \textbf{A}ction \textbf{R}easoning \textbf{M}odels (\arm) that integrate perception, planning, and control through a structured reasoning pipeline. \molmoact learns to interpret language instructions, sense its environment, generate spatial plans, and execute them as smooth, goal-directed trajectories. The model first encodes observations and instructions into structured 2.5D representations via autoregressive prediction of depth-aware perception tokens. These tokens condition the generation of mid-level planning representations, which, when visualized as visual traces in image space, guide the prediction of precise, low-level robot actions. This three-stage reasoning architecture enables \molmoact to produce explainable and steerable behavior as shown in Figure \ref{fig:fig1_overview}.

\molmoact{}’s structured design delivers both strong performance and high explainability. On standard benchmarks such as \libero and \simpler (Google Robot), \molmoact consistently outperforms competitive baselines including \groot \citep{nvidia2025gr00tn1openfoundation}, \pizero and \pizerofast \citep{black2410pi0}, RT-1 \citep{brohan2022rt}, and TraceVLA \citep{zheng2024tracevla}. In arena-style human evaluations for open-ended language instruction following, \molmoact is preferred over baselines, achieving significantly higher Elo ratings. The model adapts to novel tasks more effectively through lightweight fine-tuning, surpassing other strong baselines in efficiency. Moreover, it generalizes well to diverse environments and task perturbations in both simulation and real-world settings. Its visual reasoning traces offer an explainable view into the model’s decision-making, while also enabling direct action steering by editing trajectory lines—an approach we find more reliable than language commands, which can suffer from ambiguity.

\molmoact is fully open in every aspect: we release the model weights, training code, and all components of our action reasoning dataset. We aim for \molmoact to be more than a high-performing robotics foundation model that serves as a blueprint for building agents that reason, transforming perception into purposeful action through reasoning.

\section{MolmoAct}
\label{sec:model}

\molmoact is a fully open-source action reasoning model (\arm) for robotic manipulation. It builds on Molmo~\citep{deitke2024molmo}, reusing its vision--language backbone composed of a vision encoder, a vision--language connector, and a large language model (LLM). While Molmo supports chain-of-thought reasoning for language and vision, \molmoact extends this capability to generate grounded action sequences. The model is trained on a subset of the Open X-Embodiment (OXE) \citep{o2024open} mixture consisting of BC-Z \citep{jang2022bc}, BridgeData V2 \citep{walke2023bridgedata}, and RT-1 \citep{brohan2022rt}, and our in-house collected \molmoactdata.  

In the following sections, we describe the VLM preliminaries (\autoref{sec:model:vlm}), our method to adapt VLMs for action prediction via action tokenization(\autoref{sec:model:vla}), how we transform Molmo into an action reasoning model (\arm, \autoref{sec:model:arm}), and our approach to steer action by visual reasoning traces (\autoref{sec:model:steer}).

\subsection{Vision Language Model}
\label{sec:model:vlm}

To equip an action model with visual and linguistic world knowledge, we build upon vision–language models (VLMs). Most modern VLMs share a three-component structure: (i) a visual encoder that transforms an image into patch-level embeddings, (ii) a projection module that maps these visual features into the input space of a language model, and (iii) a large language model (LLM) backbone. These components are typically trained with a next-token prediction objective on paired or interleaved image–text data.

Our work builds on Molmo, the Multimodal Open Language Model, which follows this standard design. It employs a Vision Transformer (ViT) visual encoder, a two-layer MLP connector for projecting vision features into the language embedding space, and a decoder-only LLM backbone. In our implementation, we use vision encoders such as OpenAI ViT-L/14 336px CLIP~\citep{radford2021learning} and ViT-SO400M/14 384px SigLIP2~\citep{tschannen2025siglip}, paired with open LLMs including OLMo2-7B~\citep{olmo20242} and Qwen2.5-7B~\citep{qwen2025qwen25technicalreport}. We trained \molmoacto with a VLM backbone based on OpenCLIP and OLMo2-7B, and \molmoactd with a backbone based on SigLIP2 and Qwen2.5-7B. For full details of our model architecture and implementation, please refer to \autoref{supp:model}.

The degree of openness in the backbone components varies. SigLIP2 and Qwen2.5 do not disclose the full details of their pre-training, and are presumed to use large-scale Internet-sourced multimodal data. In contrast, OLMo2 provides open training datasets (e.g., LAION-2B/5B~\citep{schuhmann2022laion}, Dolma~\citep{soldaini2024dolma}), model weights, and complete training code. Although OpenAI CLIP also uses closed training data, it can be reproduced from scratch, as shown by MetaCLIP~\citep{metaclip}. We use the model from OpenAI because it was trained for higher resolution images, and also following the previous choice from Molmo~\citep{deitke2024molmo}. In \molmoact, we initialize vision and language components from these open checkpoints whenever available, and use the pre-training procedure from Molmo~\citep{deitke2024molmo} to train the VLM on dense captioning data. Then, after vision-language alignment, we start to fine-tune \molmoact with a subset of Open X-Embodiment (OXE) mixture and the \molmoactdata. This enables full reproducibility and supports community-driven re-training and ablation studies on data curation and scaling.

\subsection{Vision-Language-Action Model}
\label{sec:model:vla}

A standalone VLM—even when expertly prompted—cannot directly control a robot: it lacks a representation of the robot’s action space and dynamics, and thus can only provide high-level planning over the current observation. To produce accurate, executable commands, we follow prior work \citep{brohan2022rt,kim2024openvla,zitkovich2023rt} in formulating action prediction as a vision–language sequence modeling task. For each action dimension, we normalize using dataset quantiles and discretize into 256 uniform-width bins between the first and ninety-ninth percentiles, which reduces the influence of outliers while preserving effective granularity. This yields an $N$-dimensional action represented as $N$ integers in $[0,255]$. The model is trained end-to-end with a next-token prediction objective, and the loss is computed only on the action tokens.

Prior work represents the 256 discretized action bins with 256 distinct language tokens taken from the tail of the vocabulary. Continuous action bins, however, possess ordinal structure and local correlation, whereas arbitrary language tokens are effectively unrelated. This mismatch yields a poor initialization for learning an action codebook. We adopt a simple alternative that better reflects the geometry of the action space. We first identify the final 256 tokens in the Qwen2 tokenizer and, for each, use its underlying byte-level BPE symbol. We then assign them monotonically to the 256 bins so that adjacent bins map to adjacent symbols, and this becomes our action token vocabulary $V_{\mathrm{action}}$. We notice that these BPE symbols have a better initialization for action token embeddings by sharing similar characters between adjacent action bins. This similarity-preserving initialization offers a smoother starting point for optimization and, in practice, substantially reduces training time. In contrast to \groot{}’s \citep{nvidia2025gr00tn1openfoundation} training time of 50,000 GPU hours, \molmoact achieves pre-training with only 9,216 GPU hours: over a 5x reduction.

\subsection{Action Reasoning Model}
\label{sec:model:arm}

\begin{figure*}[t]  % spans both columns
  \centering
  \includegraphics[width=\textwidth]{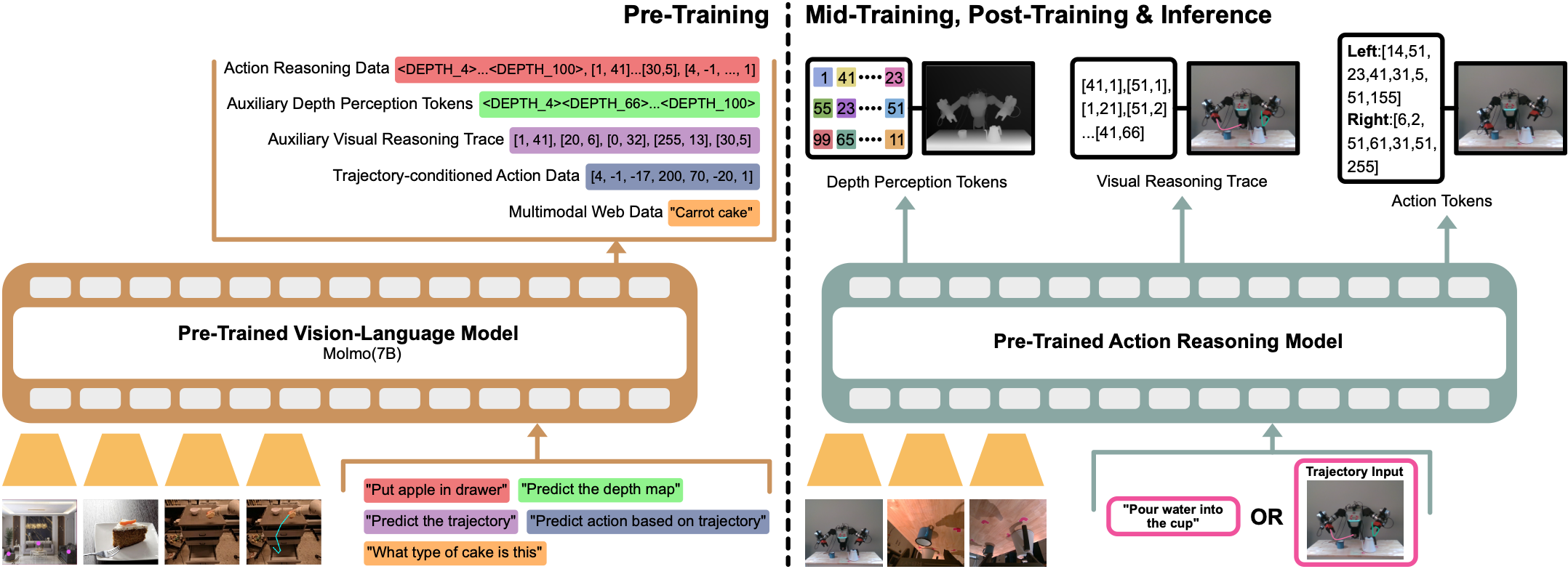}
  \caption{\textbf{Training process of \molmoact.} The model training process consists of two stages: \textbf{Pre-training} (left) and \textbf{Post-training, Mid-training \& Inference} (right). During pre-training, the vision–language backbone (Molmo) is trained on multimodal and robot reasoning data for diverse objectives, including discretized robot control, 2D pointing, trajectory drawing, open-vocabulary question answering, and perception token prediction. In post-training, the action reasoning model consumes multi-view camera images and either natural language instructions or visual trajectory inputs, generating perception tokens, visual reasoning trace tokens, and action tokens for execution.}
  \label{fig:fig2_model_and_training}
\end{figure*}

Chain-of-Thought (CoT) \citep{wei2022chain} has been shown to significantly improve Language Models' performance on complex tasks. Likewise, Multimodal Language Models (MLLM) also benefit from multimodal Chain of Thought (MCoT) \citep{lai2024mcot} in processing long multimodal contexts. However, this "think-before-you-act" paradigm is rarely present in robotic control policies. While some work attempts to incorporate reasoning to VLAs, they focus on high-level language reasoning \citep{sun2024emma,zawalski2024robotic,intelligence2025pi05visionlanguageactionmodelopenworld} such as decomposing a high-level semantic task into subtasks. While useful, they ignore two crucial aspects for precise control: depth perception and precise motion planning. First, most VLMs are trained solely on RGB images and hence lack the ability of depth estimation and 3D understanding, which is critical for robotic manipulation. Moreover, attempting to distill complex 3D trajectories into linguistic descriptions often results in significant loss of spatial and temporal information.

Contrast to previous approaches, \molmoact does not incorporate intermediate reasoning through language; rather, we teach our models to reason in space. Conditioned on images and instructions, the model autogregressively generates a sequence of \emph{depth perception tokens}, followed by \emph{visual reasoning trace} of the intended end-effector motion, before predicting the action tokens.

\begin{figure*}[t]  
  \centering
  \includegraphics[width=0.8\textwidth]{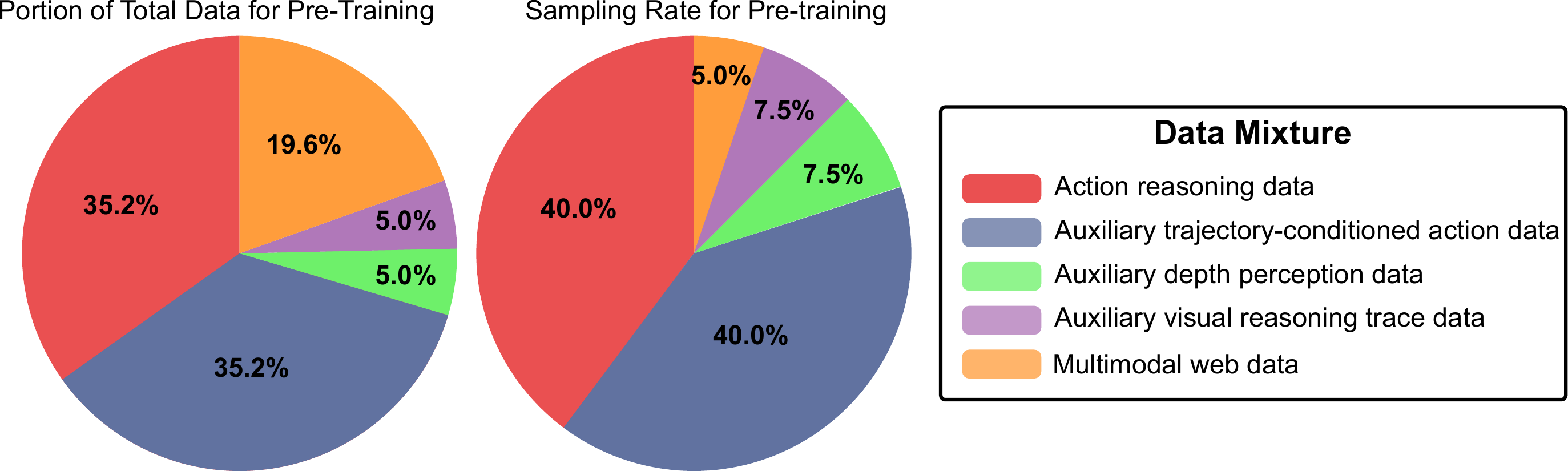}
  \caption{Distribution of data mixture in the overall pre-training mixture (left) and in the sampled subset used for \molmoact pre-training (right). The mixture contains primarily action reasoning data (38.7\%), trajectory-conditioned data (38.7\%), and multimodal web data (21.5\%), with small fractions of auxiliary depth and trace data (0.5\% each). The sampled subset increases the proportion of auxiliary data (7.5\% each for depth and line) while reducing multimodal web data to 5\%.}

  \label{fig:fig3_datamix}
\end{figure*}

\paragraph{Depth Perception Tokens}
Depth estimation is a key aspect for spatial understanding and robot action prediction. Most conventional VLMs and VLAs condition their outputs solely on the RGB image input and text instruction. Lacking depth estimation, they fail on tasks that require spatial understanding. Prior work \citep{bigverdi2024perceptiontokensenhancevisual} has shown that depth perception tokens are effective in enhancing chain-of-thought reasoning for visual–spatial tasks. Building on this insight, we leverage depth estimation as a key component to enable fine-grained robotics control in 3D environments. We now examine how each intermediate step in the reasoning chain is formulated:

We begin by defining the auxiliary vocabulary that contains depth perception tokens. Let
\begin{equation}
\label{eq:depth_vocab}
V_{\mathrm{depth}}=\{\langle\mathrm{DEPTH\_START}\rangle,\langle\mathrm{DEPTH\_END}\rangle\}
\;\cup\;
\{\langle\mathrm{DEPTH}\_k\rangle\}_{k=1}^{N}
\end{equation}
be the discrete token set used to represent depth, with $N=128$. For each input image, the target depth string is defined as
\begin{equation}
\label{eq:depth_string}
\mathbf{d}
= \big( \langle\mathrm{DEPTH\_START}\rangle\;
\langle\mathrm{DEPTH}\_{z^{\mathrm{depth}}_{1}}\rangle\;
\ldots\;
\langle\mathrm{DEPTH}\_{z^{\mathrm{depth}}_{M}}\rangle\;
\langle\mathrm{DEPTH\_END}\rangle \big)
\in V_{\mathrm{depth}}.
\end{equation}
with $M=100$, and each $z^{\mathrm{depth}}_{i}\in\{1,\ldots,N\}$ indexing a code in a VQVAE \citep{van2017neural} codebook $\mathcal{C}=\{c_1,\ldots,c_{N}\}$. The codebook is produced by a pre-trained specialist depth estimator (trained on depth maps from Depth Anything V2), which quantizes the dense depth map into a fixed-length sequence of $M$ indices. There is a deterministic one-to-one correspondence between each codebook index and a depth token in $V_{\mathrm{depth}}$ (i.e., index $k$ maps to $\langle\mathrm{DEPTH}\_k\rangle$), so $\mathbf{d}$ is a discrete, explainable summary of the depth map of the scene. We employ a specialist-to-generalist distillation strategy to ground \molmoact's depth perception tokens prediction: the specialist produces $\mathbf{d}$ as the ground-truth depth string, and the VLA is trained to predict this string autoregressively from the original RGB observation, thereby internalizing depth in a form that can condition downstream trajectory and action generation.

\paragraph{Visual Reasoning Trace}
Planning is a crucial component in robotics. Instead of planning through subtask decomposition in language, we predict intermediate two-dimensional representation to better align visual inputs and control outputs across diverse robots and tasks. In particular, we generate the 2D trajectory of the end-effector and train the model to predict this trajectory alongside the next action command, a strategy shown to be effective in prior work \citep{li2025hamster,zheng2024tracevla,gu2023rttrajectory,niu2024llarva}. These predicted waypoints help align each action to precise end-effector locations, enabling the model to focus on fine-grained localization and thereby improving action-prediction accuracy. To achieve this, we introduce Visual Reasoning Trace.

Given a certain image observation, we define the end-effector visual reasoning trace on that image as a polyline with $L$ points, $1\!\le\!L\!\le\!5$ (i.e., 0 to 4 line segments), 
\begin{equation}
\label{eq:visual_trace}
\boldsymbol{\tau} \;=\; (p_1,\ldots,p_L), \qquad p_i = (u_i,v_i).
\end{equation}
where for every $p_i = (u_i,v_i)$, the coordinates of the point are normalized with respect to the given image dimension so that $u_i,v_i \in \{0, ..., 255\}$.

Note that $p_1$ corresponds to the coordinate where the robot end-effector is located in the given image, and all other points are the location of the end-effector in later frames. For the full polyline, points are subsampled evenly from the future horizon of the episode between the current frame and the terminal frame.

\begin{figure*}[t]  
  \centering
  \includegraphics[width=\textwidth]{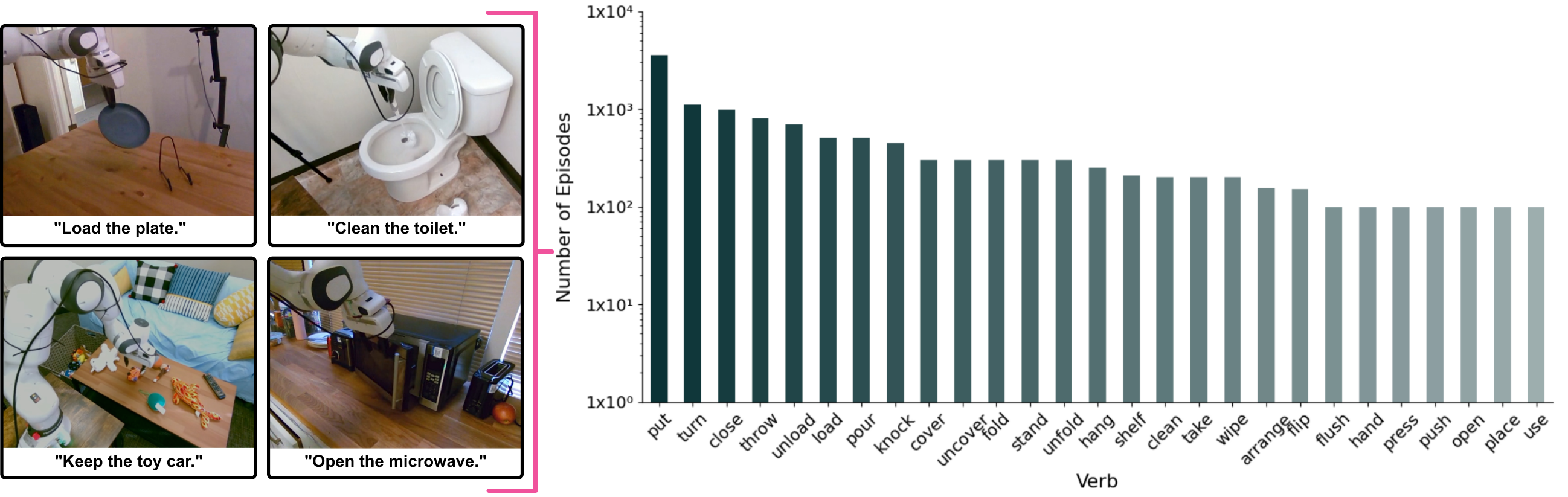}
  \caption{\textbf{Examples and verb distribution in the \molmoactdata.} 
Left: Sample robot manipulation tasks paired with natural language instructions, spanning diverse household activities such as closing a laptop, loading a plate, cleaning a toilet, and opening a microwave. 
Right: Log-scale distribution of the top verbs in the dataset, showing a long-tail pattern with ``put,'' ``turn,'' and ``close'' as the most frequent actions.}

  \label{fig:fig4_molmoact_dataset}
\end{figure*}

\paragraph{Action Reasoning Procedure}
With depth perception tokens and visual reasoning trace, \molmoact can finally perform action reasoning in space with the following procedure: given an RGB image observation \(I\) and a language instruction \(T\) (which includes an action CoT prompt), the model autoregressively generates three token sequences in order: (i) depth perception tokens $\mathbf{d}$; (ii) visual reasoning trace $\boldsymbol{\tau}$; (iii) action tokens $\mathbf{a}$, where $\mathbf{a} = (a_{1},\dots,a_{D}) \in V_{\mathrm{action}}^{D}$ with $D$ degree of freedoms.
These are produced according to the factorization
\begin{equation}
\label{eq:action_reasoning}
p(\mathbf{d},\boldsymbol{\tau},\mathbf{a}\,\big\lvert\,I,T)
=\:\prod_{i=1}^{M+2} p\bigl(d_i \mid I,T,\mathbf{d}_{<i}\bigr)\;\times\;
\prod_{j=1}^{L} p\bigl(\tau_j \mid I,T,\mathbf{d},\boldsymbol{\tau}_{<j}\bigr)\;\times\;
\prod_{k=1}^{D} p\bigl(a_k \mid I,T,\mathbf{d},\boldsymbol{\tau},\mathbf{a}_{<k}\bigr).
\end{equation}
In other words, the depth string \(\mathbf{d}\) is generated first, then the visual reasoning trace \(\boldsymbol{\tau}\), and finally the action tokens \(\mathbf{a}\). By conditioning each stage on the preceding depth and trajectory tokens, we ensure that the final actions are spatially grounded in both the inferred depth and the planned motion sketch.

\subsection{Action Steerability via Visual Reasoning Trace}
\label{sec:model:steer}

We define \emph{steerability} as the ability to guide a policy at test time to perform different behaviors using user-provided instructions. Most prior VLA systems rely exclusively on language for steering. However, language-only steering faces three practical challenges: (i) it requires large and diverse corpora of high-quality language–action pairs to learn a reliable grounding between words and control, (ii) natural language is inherently ambiguous about magnitudes, scales, and endpoints, and (iii) post-trained models often exhibit narrow prompting habits, making them brittle to out-of-distribution phrasing. For manipulation, these issues translate into imprecise or inconsistent control. We therefore seek a steering modality that is both precise and scalable. Rather than relying on ambiguous language prompts, we allow the user to draw a visual reasoning trace \(\boldsymbol{\tau}\) directly on the camera image to indicate the desired end-effector path. A trace \(\boldsymbol{\tau}=(p_{1},\dots,p_{L})\), \(1\le L\le5\) (i.e., 0 to 4 line segments), is overlaid onto the RGB image \(I\) to form an augmented observation \(I^{+}=I\oplus\boldsymbol{\tau}\). visual reasoning traces are unambiguous, easily edited, and generalize across tasks without large text–action corpora or brittle language patterns.

At test time, given \(I\), instruction \(T\), and a user-drawn trace \(\boldsymbol{\tau}\), we construct \(I^{+}=I\oplus\boldsymbol{\tau}\) and generate the next-step action tokens \(\mathbf{a}=(a_{1},\dots,a_{D})\) autoregressively:
\begin{equation}
\label{eq:action_steer}
p(\mathbf{a}\mid I^{+},T)
=\;\prod_{k=1}^{D}p\bigl(a_{k}\mid I^{+},T,\mathbf{a}_{<k}\bigr).
\end{equation}
By conditioning directly on the overlaid trace, the model executes closed-loop control that follows the user’s sketch. Repeating this at each timestep yields precise, interactive steering that is both scalable and robust to out-of-distribution phrasing.

\section{Data Curation and Generation}
\label{sec:robotreasoning}

\molmoact is trained on a diverse set of datasets. During Pre-training, \molmoact is trained on Multimodal Web data, Auxiliary Robot Data as well as Action Reasoning Data. Furthermore, we collected and trained with the \molmoactdata for the Mid-training stage. Below, we describe each dataset and its collection process; further details and examples are provided in the \autoref{supp:data}.

\subsection{Action Reasoning Data}
\label{sec:data:gen}

As discussed in Section~\ref{sec:model:arm}, \molmoact frames visuomotor control as an autoregressive sequence modeling problem augmented with an action chain-of-thought (CoT). This formulation allows us to convert any conventional robot action dataset into action reasoning data by appending predicted depth perception tokens and visual reasoning traces to action tokens, conditioned on both language and robot observations. Section~\ref{sec:data:gen} details the process for generating ground-truth labels for the depth perception tokens and visual reasoning traces, and explains how these are combined with action labels to train \molmoact.

A robot episode typically consists of a sequence of timesteps, where each timestep is a tuple $(I, T, \textbf{a})_t$, containing an RGB observation image $I$, a language instruction $T$, and a ground-truth action $\textbf{a}$, specified either in end-effector space or joint space. To convert any robot data into the Action Reasoning data format, we generate ground-truth \textit{Depth Perception Tokens} and \textit{Visual Reasoning Traces} for each timestep in the episode.  We explain the details for generating ground-truth labels for Depth Perception Tokens and Visual Reasoning Trace below.

\paragraph{Depth Perception Tokens}
To generate \textit{Depth Perception Tokens} for each frame of a demonstration, we first train a VQVAE on 10 million depth maps of tabletop manipulation images collected from the RT-1, BridgeData V2, and BC-Z datasets. We use DepthAnything-v2 to obtain a depth map for each observation RGB image. The VQVAE is trained with a standard reconstruction objective to minimize reconstruction loss between input RGB images and their corresponding depth maps for 20 epochs. Once the VQVAE has been trained, we encode each observation image with the VQVAE to get their latent embeddings. We then represent the latent embedding with a learned codebook with 128 dimension based on a one-to-one index to depth token mapping. Note that all images are resized to 320$\times$320 px during training and inference to enforce the representation of 100 tokens per image. This allows us to express the depth map of each observation image as a tokenized string of 100 tokens, which we use for ground truth labeling for our depth perception token.

\paragraph{Visual Reasoning Trace}

To generate a \textit{Visual Reasoning Traces} for each frame of a demonstration, we employ Molmo, a vision-language model trained on diverse 2D pointing datasets, for data generation akin to synthetic data generation in NLP. For each timestep $t$, we extract the pixel coordinates $(u_t,v_t)$ of the robot’s gripper, and aggregate these across the episode to obtain a visual reasoning trace. For each observation frame, we prompt Molmo with the instruction \texttt{"point to the robot gripper"} for single-arm robots or \texttt{"point to the robot gripper on the left/right"} for bimanual embodiments. Molmo returns a 2D coordinate $(x_t,y_t) \in \mathbb{R}^2$ and $(x_t,y_t) \in [0, 100]$, corresponding to the predicted gripper location in image space. We rescale the coordinate values so that $(u_t,v_t) \in \mathbb{Z}^2$ and $(u_t,v_t) \in [0, 255]$. We then apply this query at every timestep in the episode, resulting in one gripper location per frame. Linking these predictions sequentially yields the full trajectory $\boldsymbol{\tau}$. In the case of bimanual robots, two separate prompts are issued per frame to obtain $\boldsymbol{\tau}_L$ and $\boldsymbol{\tau}_R$ for the left and right grippers, respectively. At each timestep $t$, we construct a visual reasoning trace by selecting a subsequence from $t$ to the episode end $e$. This includes the current point $(u_t,v_t)$, the final point $\big(u_e, v_e\big)$, and up to three intermediate points spaced uniformly between them. If fewer than three intermediate points are available (i.e., $e - t < 4$), we include all available points. If $t = e$, the trace contains only one point. This yields a visual reasoning trace between 1 to 5 points representing the future motion of the end effector.

\paragraph{Auxiliary Robot Data}
To strengthen \molmoact's ability to reason in space, we extend the same data generation pipeline used for depth perception token, visual reasoning trace to curate three auxiliary supervision dataset: (i) Auxiliary Depth Data—given an RGB observation and language instruction, the model only predicts the target Depth Perception Token sequence; (ii) Auxiliary Trace Data—given an RGB observation and language instruction, the model only predicts the corresponding Visual Reasoning Trace; and (iii) Trajectory-conditioned Action Data—given $o_t=(I, T, \boldsymbol{\tau})_t$, where $I$ is the current image, $T$ the instruction, and $\boldsymbol{\tau}=\big(p_1,\ldots,p_L\big)$ the ground truth Visual Reasoning Trace, the model predicts the next action by taking the language $T$ and the trace-overlaid image \(I^{+}=I\oplus\boldsymbol{\tau}\). Note that we curate the Trajectory-conditioned Action Data mainly for enabling the steerability feature of \molmoact.

Once we generate the ground truth label for each frame, we construct the action reasoning dataset by sequentailly aligning the ground-truth Depth Perception Tokens, Visual Reasoning Trace, and Action for instruction tuning. We also use the same data generation approach to obtain auxiliary robot data.

\subsection{MolmoAct Dataset}
We curated the \molmoactdata to improve the model’s general manipulation performance and spatial reasoning in real household environments. The dataset contains 10,689 high-quality trajectories of a single-arm Franka robot performing 93 unique manipulation tasks in both home and tabletop environments as shown in Figure \ref{fig:fig4_molmoact_dataset}. The average length of each trajectory spans 112 timesteps. Data collection spanned two months and involved five full-time operators following strict protocols. For further details, see \autoref{supp:data}. The \molmoactdata includes manipulation data from two primary settings: home environments and tabletops.

\begin{itemize}
    \item \textbf{Home Environment Data.} \\
    To collect diverse home environment data, we mounted a single-arm Franka robot on a lightweight, mobile platform similar to DROID \citep{khazatsky2024droid}, enabling us to transport the robot and capture scenes across living rooms, kitchens, bathrooms, and bedrooms. Each task was designed to reflect a specific household chore. For example, the long-horizon task “clean up the dishes” was decomposed into subtasks such as “put the bowl in the dishwasher,” “put the fork in the sink,” and “cover the pot.” This decomposition allows the policy to learn individual skill components in isolation before composing them into more complex behaviors. In total, we collected 7,730 trajectories spanning 73 distinct tasks and 20 verbs across a wide variety of scenes.
    
    \item \textbf{Tabletop Data.} \\
    We also collected 2,959 tabletop trajectories covering 20 atomic tasks, each performed with a diverse set of objects to promote robustness and generalization. Each task was decomposed into atomic motions and reinforced in a simplified tabletop environment. For example, the task “put the bowl in the dishwasher” consists of a sequence of motions such as opening the dishwasher, grasping the bowl, flipping it, and placing it inside. We isolated and separately collected data for each atomic motion—open, pick, flip, put, and close—to build a comprehensive set of motion primitives.
\end{itemize}

\begin{table*}[t]
  \centering
  \small
  \caption{\textbf{\simpler evaluation across different policies on Google Robot tasks.} 
  The zero-shot and fine-tuning results denote performance of OXE dataset~\citep{o2024open} pre-trained models and RT-1 dataset~\citep{brohan2022rt} fine-tuned models, respectively.}
  \label{tab:simplerenvgooglerobot}
  \resizebox{\textwidth}{!}{
  \begin{tabular}{l|cccc|c|cccc|c}
    \toprule
    \multirow{2}{*}{Model} &
    \multicolumn{4}{c|}{\textbf{Visual Matching}} &
    \multirow{2}{*}{Avg} &
    \multicolumn{4}{c|}{\textbf{Variant Aggregation}} &
    \multirow{2}{*}{Avg} \\
    \cline{2-5} \cline{7-10}
    & Pick Coke Can & Move Near & Open/Close Drawer & & & Pick Coke Can & Move Near & Open/Close Drawer & \\
    \midrule
    % RT-1 (Begin) & 2.7\% & 5.0\% & 13.9\% & & 6.8\% & 2.2\% & 4.0\% & 6.9\% & & 4.2\% \\
    % RT-1 (15\%) & 71.0\% & 35.4\% & 56.5\% & & 60.2\% & 81.3\% & 44.6\% & 26.7\% & & 56.2\% \\
    % RT-1 (Converged) & 85.7\% & 44.2\% & 73.0\% & & 74.6\% & 89.8\% & 50.0\% & 32.3\% & & 63.3\% \\
    % \midrule
    HPT \citep{wang2024scaling} & 56.0\% & 60.0\% & 24.0\% & & 46.0\% & \multicolumn{4}{c|}{---} & --- \\
    TraceVLA \citep{zheng2024tracevla} & 28.0\% & 53.7\% & 57.0\% & & 42.0\% & 60.0\% & 56.4\% & 31.0\% & & 45.0\% \\
    RT-1-X \citep{brohan2022rt} & 56.7\% & 31.7\% & 59.7\% & & 53.4\% & 49.0\% & 32.3\% & 29.4\% & & 39.6\% \\
    RT-2-X \citep{zitkovich2023rt} & 78.7\% & 77.9\% & 25.0\% & & 60.7\% & 82.3\% & 79.2\% & 35.3\% & & 64.3\% \\
    Octo-Base \citep{team2024octo} & 17.0\% & 4.2\% & 22.7\% & & 16.8\% & 0.6\% & 3.1\% & 1.1\% & & 1.1\% \\
    OpenVLA \citep{kim2024openvla} & 16.3\% & 46.2\% & 35.6\% & & 27.7\% & 54.5\% & 47.7\% & 17.7\% & & 39.8\% \\
    RoboVLM (zero-shot) \citep{liu2025towards} & 72.7\% & 66.3\% & 26.8\% & & 56.3\% & 68.3\% & 56.0\% & 8.5\% & & 46.3\% \\
    RoboVLM (fine-tuned) & 77.3\% & 61.7\% & 43.5\% & & 63.4\% & 75.6\% & 60.0\% & 10.6\% & & 51.3\% \\
    Emma-X \citep{sun2024emma} & 2.3\% & 3.3\% & 18.3\% & & 8.0\% & 5.3\% & 7.3\% & 20.5\% & & 11.0\% \\
    Magma \citep{yang2025magma} & 56.0\% & 65.4\% & 83.7\% & & 68.4\% & 53.4\% & 65.7\% & 68.8\% & & 62.6\% \\
    \pizero (fine-tuned) \citep{black2410pi0} & 72.7\% & 65.3\% & 38.3\% & & 58.7\% & 75.2\% & 63.7\% & 25.6\% & & 54.8\% \\
    \pizerofast (fine-tuned) & 75.3\% & 67.5\% & 42.9\% & & 61.9\% & 77.6\% & 68.2\% & 31.3\% & & 59.0\% \\
    \groot (fine-tuned) \cite{nvidia2025gr00tn1openfoundation} & 69.3\% & 68.7\% & 35.8\% & & 52.4\% & 46.7\% & 62.9\% & 17.5\% & & 43.7\% \\
    SpatialVLA \citep{qu2025spatialvla}  & 81.0\% & 69.6\% & 59.3\% & & 70.0\% & 89.5\% & 71.7\% & 36.2\% & & 65.8\% \\
    \rowcolor[HTML]{EFEFEF} \molmoact (zero-shot) & 71.3\% & 73.8\% & 66.5\% & & 70.5\% & 57.8\% & 43.8\% & 76.7\% & & 59.3\% \\
    \rowcolor[HTML]{EFEFEF} \molmoact (fine-tuned) & 77.7\% & 77.1\% & 60.0\% & & \textbf{71.6\%} & 76.1\% & 61.3\% & 78.8\% & & \textbf{72.1\%} \\
    \bottomrule

  \end{tabular}
  }
\end{table*}

\subsection{Multimodal Web Data}

Prior works have shown that co-training VLAs with the data mixture from the VLM training leads to more generalizable policies. These policies are more robust to perturbations such as lighting and background changes, and can generalize better to unseen environments and objects. We include a mixture of multimodal web data from Molmo’s Supervised fine-tuning stage involving academic datasets (VQA v2.0 \citep{goyal2017making}, Text VQA \citep{singh2019towards}, OK-VQA \citep{okvqa}, ChartQA \citep{masry-etal-2022-chartqa}, DocVQA \citep{mathew2021docvqa}, Infographic VQA \citep{mathew2022infographicvqa}, AI2D \citep{kembhavi2016diagram}, A-OKVQA \citep{schwenk2022okvqa}, AndroidControl \citep{li2024effects}, ScienceQA \citep{lu2022learn}, TabMWP \citep{lu2023dynamic}, ST-VQA \citep{biten2019scene}, TallyQA \citep{acharya2019tallyqa}, DVQA \citep{kafle2018dvqa}, FigureQA \citep{kahou2017figureqa}, and PlotQA \citep{methani2020plotqa}) for general visual skills and PixMo \citep{deitke2024molmo} for fine-grained understanding and pointing. Furthermore, we include LVIS \citep{gupta2019lvis} where the model is asked to predict the bounding box center of instances of a certain category to ground language to image regions.

\section{Training Recipe}
\label{sec:train}

\molmoact is first pre-trained on action reasoning data curated from a subset of the OXE dataset, along with the auxiliary robot data and multimodal web data. To further enhance its capabilities, we mid-train the model on the \molmoactdata before post-training it for specific downstream tasks and embodiments. In this section, we describe the different data mixtures and training configurations used at each stage of \molmoact’s training (as shown in Figure \ref{fig:fig2_model_and_training}.

\subsection{Pre-training}
\label{sec:train:pre}

In the first training stage, \molmoact is pre-trained on a mixture of action reasoning data, auxiliary robot data, and multimodal web data. For all robot data, we use a subset of OXE comprising RT-1, BridgeData V2, and BC-Z, totaling 10.5M samples, which we convert into action reasoning data using our reasoning in space formulation. We also include auxiliary robot data—auxiliary depth data (1.5M), auxiliary trace data (1.5M), and trajectory-conditioned action data (10.5M), and co-train with 2M samples of multimodal web data. During pre-training, data is sampled at the following rates: RT-1 (20\%), BridgeData V2 (12.5\%), BC-Z (7.5\%) for both action reasoning and trajectory-conditioned data, 7.5\% from the auxiliary depth and trace data, and 5\% from multimodal web data as shown in Figure \ref{fig:fig3_datamix}. The whole data mixture used for pre-training \molmoact totals up to 26.3M samples.

To pre-train \molmoact with the data mixture mentioned above, we use 256 H100s to train the model with 100k gradient steps using a batch size of 512, which takes around 9,728 GPU hours. At each training step, a batch of data pairs is drawn randomly from the entire data mixture by their assigned sampling rate defined above. Hyperparameter details are listed in \autoref{supp:training}.

\subsection{Mid-training}
\label{sec:train:mid}

Following the initial pre-training stage, we conduct a second stage of mid-training using high-quality action reasoning data closely aligned with our target domain of household manipulation. Specifically, we formulate the \molmoactdata into 1M action reasoning data samples and an additional 1M trajectory-conditioned action data samples, which we find beneficial for improving overall performance and action steering. To train on the \molmoactdata, we convert each sample—consisting of two side‐mounted camera views and one wrist camera view, all sharing the same instruction and ground‐truth action—into two paired‐view training examples by pairing each side view with the wrist view. For action reasoning data preparation (\autoref{sec:data:gen}), depth perception tokens and visual reasoning traces are generated only from the side views, while the wrist view is used solely for providing additional information. We train the model on this modified dataset for 50k gradient steps with a batch size of 128, using 128 H100 GPUs over around 2,304 GPU hours. Additional hyperparameters and optimization details are provided in \autoref{supp:training}.

\begin{table}[t]
  \centering
  \footnotesize           % smaller text; use \scriptsize if you need it even smaller
  \setlength{\tabcolsep}{4pt}  % tighten horizontal padding between columns
  \caption{\textbf{\libero benchmark success rates} across four task categories (Spatial, Object, Goal, and Long-horizon) along with the average performance. \molmoact achieves the highest overall average success rate of 86.6\%, outperforming all autoregressive baselines, with strong performance across all categories, particularly in long-horizon tasks.}

  \label{tab:liberobaselinessmall}
  \begin{tabular}{lccccc}
    \toprule
    \textbf{Baseline} & \textbf{Spatial} & \textbf{Object} & \textbf{Goal} & \textbf{Long} & \textbf{Avg} \\
    \midrule
    TraceVLA \citep{zheng2024tracevla}        & 84.6\% & 85.2\% & 75.1\% & 54.1\% & 74.8\%  \\
    Octo-Base  \citep{team2024octo}      & 78.9\% & 85.7\% & 84.6\% & 51.1\% & 75.1\% \\
    OpenVLA   \citep{kim2024openvla}      & 84.7\% & 88.4\% & 79.2\% & 53.7\% & 76.5\%  \\
    SpatialVLA  \citep{qu2025spatialvla}    & 88.2\% & 89.9\% & 78.6\% & 55.5\% & 78.1\% \\
    CoT-VLA     \citep{zhao2025cot}    & 87.5\% & 91.6\% & 87.6\% & 69.0\% & 83.9\% \\
    NORA-AC     \citep{hung2025nora}    & 85.6\% & 89.4\% & 80.0\% & 63.0\% & 79.5\% \\
    WorldVLA     \citep{cen2025worldvla}    & 87.6\% & 96.2\% & 83.4\% & 60.0\% & 79.1\% \\
    \pizerofast  \citep{black2410pi0}       & 96.4\% & 96.8\% & 88.6\% & 60.2\% & 85.5\%  \\
    ThinkAct  \citep{huang2025thinkact}      & 88.3\% & 91.4\% & 87.1\% & 70.9\% & 84.4\% \\
    \rowcolor[HTML]{EFEFEF} \molmoactd & 87.0\% & 95.4\% & 87.6\% & 77.2\% & \textbf{86.6\%} \\
    \bottomrule
  \end{tabular}
\end{table}

\subsection{Post-training} 
\label{sec:train:post}

After mid‐training, we conduct the final post‐training stage to rapidly adapt the model to new tasks and embodiments. For new tasks, we collect a small set of 30 to 50 tele‐operated demonstrations per task, then generate perception tokens and visual reasoning trace for each timestep. These demonstrations are converted into the action reasoning data and trajectory‐conditioned action data, following the same process as mid‐training data curation, with one key difference: we apply action chunking \citep{zhao2023learning} during post‐training, formatting action predictions in fixed‐length chunks ($N=8$). For each action chunk in all chunks, we tokenize each of them in the same way as we do for single action. After grouping them to a list, we train the model autoregressively for predicting all action chunks. We adapt \molmoact to post-train on single or multi-tasks via parameter-efficient LoRA fine-tuning. In all evaluations, we fix the LoRA rank at 32 and alpha at 16 to preserve the model’s pre-trained capabilities. For simulation benchmarks (e.g., \libero), we use a batch size of 128, and for real-world tasks we use 64. The number of gradient steps we train varies by task. Our post-training data generally consists of a front- or side-view image paired with a wrist-view image, although some setups provide multiple wrist views (e.g., bimanual scenarios). In all cases, we apply the same LoRA and training configuration described above. Additional details are available in \autoref{supp:training}.

\section{Experimental Evaluation}
\label{sec:exp}
Our experimental evaluation comprises a broad suite of studies that rigorously benchmark \molmoact against strong baselines. We assess \molmoact with \molmoactd version in (i) its pre‑training, “out‑of‑the‑box” capabilities, (ii) its post‑training adaptability across varied tasks, domains, and robotic embodiments, and (iii) its additional feature of being an interactive and steerable action reasoning model. By testing the model on a comprehensive range of scenarios both in simulation and real-world, we aim to answer the following research questions:
\begin{enumerate}[label=\textbf{\arabic*}]
    \item \textbf{How well does \molmoact perform, after pre-training, on tasks drawn from the same distribution as its training data?}\
    We address this question by benchmarking \molmoact against strong VLA models on the \simpler simulation benchmark \citep{li2024evaluating}.

    \item \textbf{How effectively does \molmoact adapt to novel tasks, domains, and embodiments through lightweight post-training fine-tuning?}\
    We fine-tune \molmoact with LoRA \citep{hu2022lora} and benchmark it against strong baselines on the \libero simulation suite \citep{liu2023libero}. We then validate its real-world performance on two hardware setups—a single and a bimanual Franka arm—to demonstrate adaptability across embodiments.

    \item \textbf{How effectively can \molmoact generalize beyond its training distribution?}\
    We investigate this through real-world out-of-distribution (OOD) tests and variant-aggregation experiments in \simpler.

    \item \textbf{How does mid-training on the \molmoactdata improve \molmoact's generalist performance?}\
    We address this through ablation experiments in the real-world evaluations to compare \molmoact’s performance with and without mid-training on the \molmoactdata.

    \item \textbf{How effectively does \molmoact follow language commands?}\
    We benchmark \molmoact against strong baselines in an open-ended simulation setup where human evaluators provide free-form prompts and assess each model’s resulting actions.

    \item \textbf{How steerable is \molmoact, and how can this steerability enhance user interaction?}\
    We perform extensive real-world ablations, guiding \molmoact by sketching trajectory cues on the interface and analyzing its responses, then examine the resulting human–robot interaction dynamics.

\end{enumerate}

\begin{figure*}[t]  % spans both columns
  \centering
  \includegraphics[width=\textwidth]{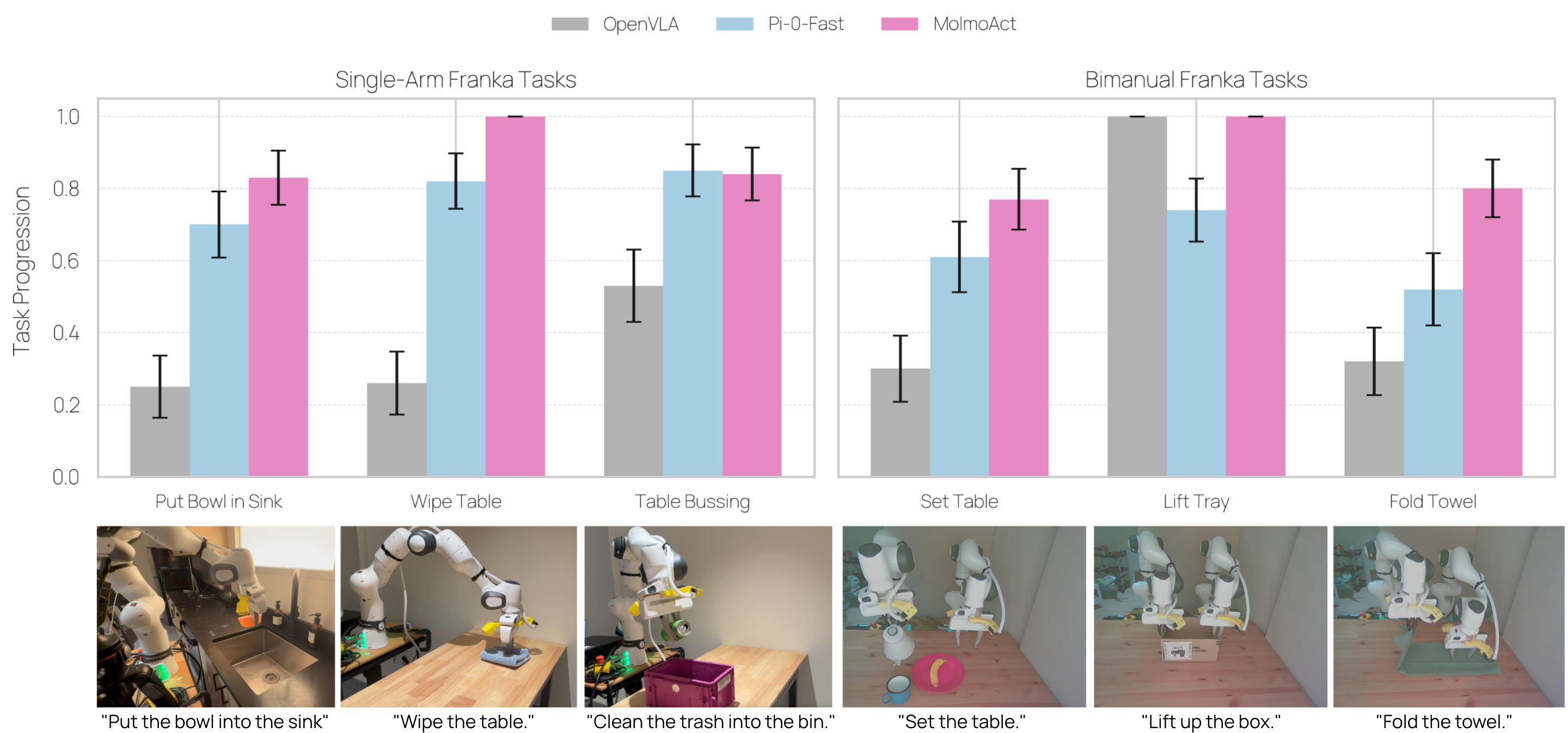}
  \caption{Real-world evaluation of OpenVLA, \pizerofast, and \molmoact on single-arm (left) and bimanual (right) Franka tasks. Bar plots report average task progression with standard error across 25 trials per task. \molmoact consistently outperforms baselines, particularly on single-arm tasks such as \emph{Wipe Table} and \emph{Table Bussing}, and maintains strong performance on bimanual tasks including \emph{Fold Towel} and \emph{Set Table}. Bottom row shows example task setups with corresponding natural language instructions.}

  \label{fig:fig5_real_eval_indist}
\end{figure*}

\subsection{\molmoact After Pre-training} 

\textbf{Evaluation Setup and Baselines.} 
We first evaluated \molmoact's zero-shot capability—its ability immediately after pre-training and before any task-specific fine-tuning. Unlike \pizero \citep{black2410pi0}, \groot \citep{nvidia2025gr00tn1openfoundation}, and other proprietary VLA models that rely on large-scale private robot datasets and the full OXE dataset for pre-training, \molmoact was trained exclusively on curated action reasoning data filtered from a subset of OXE (specifically BC-Z \citep{jang2022bc}, BridgeData V2 \citep{walke2023bridgedata}, and RT-1 \citep{brohan2022rt}), combined with multimodal web data and auxiliary robot data, as detailed in \autoref{sec:train:pre}. This amounts to approximately 26.3M samples—an order of magnitude smaller than \pizero, which uses at least 903M for pre-training. To evaluate \molmoact's out-of-the-box generalization, we used the \simpler benchmark, which features both visual-matching and variant-aggregation tasks across WidowX and Google Robot platforms (details in \autoref{supp:eval:simpler}). As \molmoact’s pre-training distribution is most aligned with the Google Robot visual-matching tasks, we focused our evaluation on this suite to best isolate in-distribution performance and capabilities of pre-training.

We compared \molmoactdpre against a set of generalist policies, including TraceVLA \citep{zheng2024tracevla}, RT-1X \citep{brohan2022rt}, OpenVLA \citep{kim2024openvla}, RoboVLM \citep{liu2025towards}, Emma-x \citep{sun2024emma}, \pizero and \pizerofast \citep{black2410pi0}, Octo \citep{team2024octo}, Magma \citep{yang2025magma}, HPT \citep{wang2024scaling}, SpatialVLA \citep{qu2025spatialvla} and \groot \citep{nvidia2025gr00tn1openfoundation}. Most baselines were evaluated in the zero-shot setting, with a subset also tested after fine-tuning. We additionally fine-tuned \molmoactdpre on the RT-1 subset of OXE to assess its capacity when given more pre-training data.

\textbf{Evaluation Results.} 
\molmoactdpre achieved strong zero-shot performance on the \simpler visual-matching suite, reaching 70.5\% success rate and outperforming baselines such as \groot, \pizero, \pizerofast, and Magma. With fine-tuning on the same RT-1 subset of OXE, \molmoactd improved to 71.6\%, exceeding Magma by 3.2\% as shown in Table \ref{tab:simplerenvgooglerobot}. These results indicate that \molmoact is both an effective zero-shot generalist and a strong initialization for fine-tuned deployment.

\subsection{Fast Adaptation of \molmoact in Post-training}

\textbf{Evaluation Setups and Baselines.}
We evaluate \molmoact in both simulation and real-world settings to assess its fast adaptation after post-training. In simulation, we evaluate on the \libero simulation benchmark (\cite{liu2023libero}), which consists of a Franka Emika Panda arm in simulation with demonstrations containing front and wrist view camera images (256$\times$256 px), language instructions, and delta end-effector pose actions. We follow prior works (\cite{kim2024openvla}) and evaluate on four task suites -- \libero-Spatial, \libero-Object, \libero-Goal, and \libero-Long -- each with 500 demonstrations across 10 tasks. Following (\cite{kim2024openvla}), we trained on a modified dataset that filtered out no-op actions and unsuccessful demonstrations. Moreover, we set action chunk size to $K = 8$ for evaluation on each task suite and execute full chunks before redoing action reasoning. We fine-tune \molmoactd using Low-Rank Adaptation (LoRA) and compared to state-of-the-art generalist autoregressive policies, such as TraceVLA \citep{zheng2024tracevla}, OpenVLA \citep{kim2024openvla}, SpatialVLA \citep{qu2025spatialvla}, \pizerofast \citep{black2410pi0}, CoT-VLA \citep{zhao2025cot}, WorldVLA \citep{cen2025worldvla}, ThinkAct \citep{huang2025thinkact}, and NORA-AC \citep{hung2025nora}.

In the real world, we evaluate \molmoact on six tasks across single-arm and bimanual Franka setups. The single-arm tasks include \texttt{put\_bowl\_in\_sink}, \texttt{wipe\_table}, and \texttt{table\_bussing}. The bimanual tasks include \texttt{set\_table}, \texttt{lift\_tray}, and \texttt{fold\_towel}. For each task, we collected 50 human tele-operated demonstrations and post-trained both \molmoactd and baseline models. We evaluate the task progress over 25 trials per task. We detail the task progress scores in \autoref{supp:eval:indist}. This setup enables a comprehensive comparison of adaptation efficiency across tasks and embodiments.

\textbf{Evaluation Results.} 
On the \libero benchmark, \molmoactd achieves an average success rate of 86.6\%, the highest among all compared methods. It performs particularly well on \libero-Long, a challenging long-horizon suite, where it exceeds the performance of ThinkAct—the second-best method in this setting—by 6.3\%. In the real world, \molmoact demonstrates effective fine-tuning and generalization across different embodiments. It outperforms \pizerofast by an average of 10\% in task progression on single-arm tasks and by 22.7\% on bimanual tasks, as shown in Figure \ref{fig:fig5_real_eval_indist}.

\begin{figure}[t]
  \centering
  \begin{subfigure}{0.75\columnwidth}
    \centering
    \includegraphics[width=\linewidth]{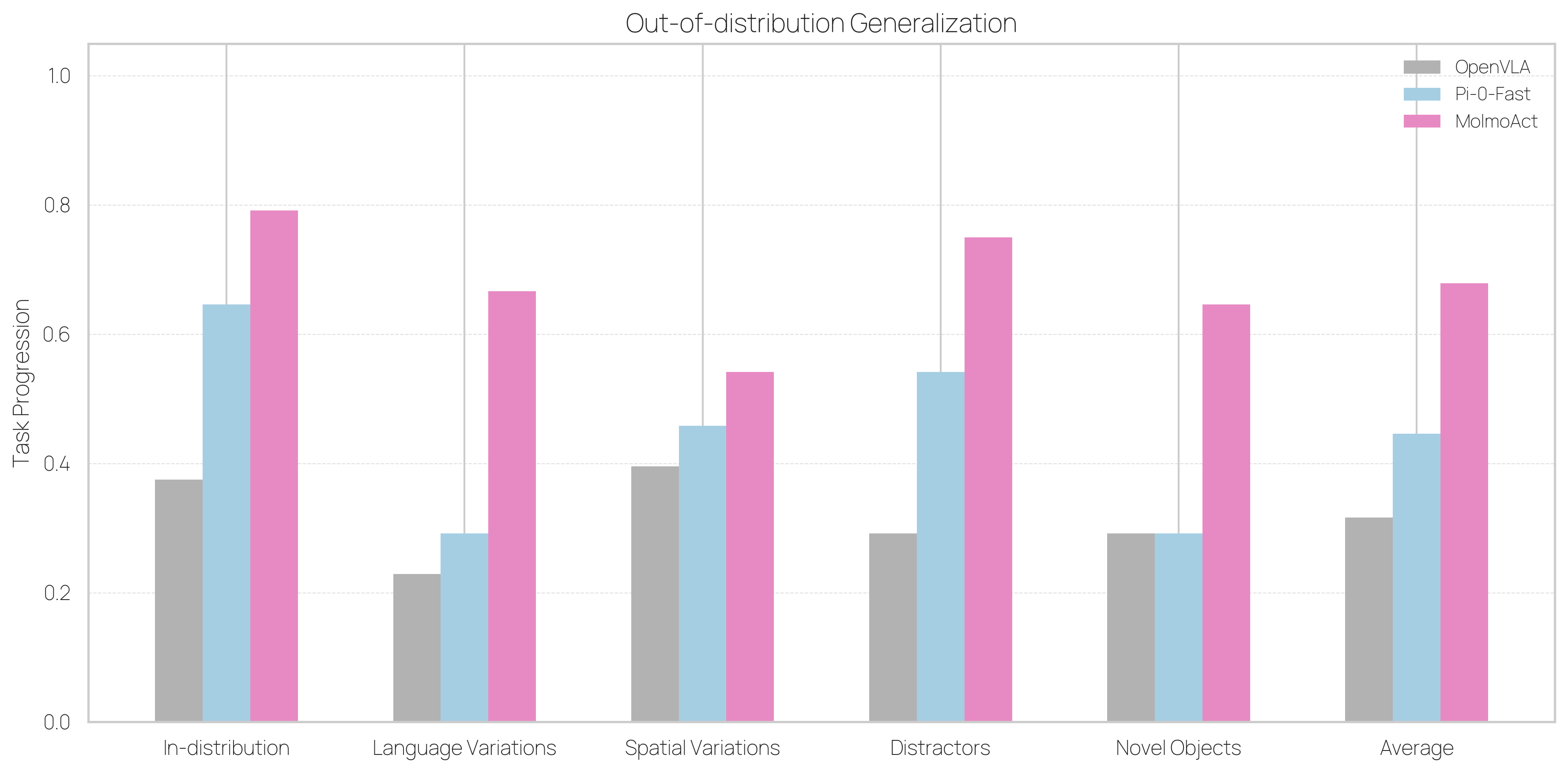}
    \caption{\molmoact generalizes beyond training distributions.}
    \label{fig:fig6_1_real_eval_outdist}
  \end{subfigure}
  \hspace{0.02\columnwidth}
  \begin{subfigure}{0.75\columnwidth}
    \centering
    \includegraphics[width=\linewidth]{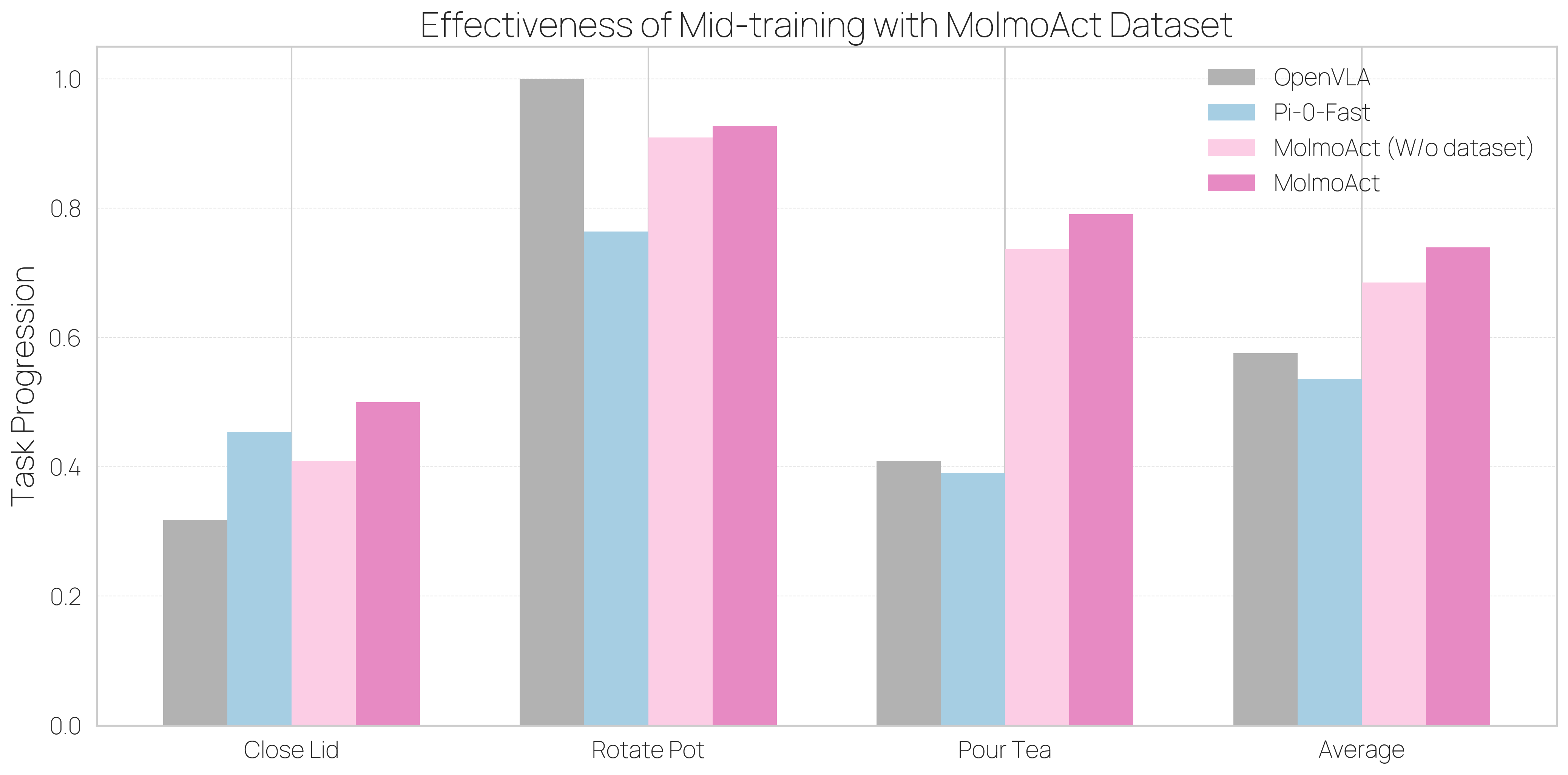}
    \caption{\molmoactdata improves task performance.}  % Edit as needed
    \label{fig:fig6_2_real_eval_molmoact_dataset}
  \end{subfigure}
  \caption{\textbf{\molmoact outperforms baselines across generalization and mid-training settings.} 
(a) Out-of-distribution generalization: Task progression scores for OpenVLA, \pizerofast, and \molmoact across in-distribution, language variation, spatial variation, distractors, and novel object conditions, showing consistent gains for \molmoact. 
(b) Effectiveness of mid-training with the \molmoactdata: Comparison of task progression on real-world tasks (Close Lid, Rotate Pot, Pour Tea) for \molmoact with and without the dataset, \pizerofast, and OpenVLA, demonstrating that mid-training with the dataset improves performance across tasks.
}

  \label{fig:molmoactcomparison2}
\end{figure}

\begin{figure*}[t]  % spans both columns
  \centering
  \includegraphics[width=0.9\textwidth]{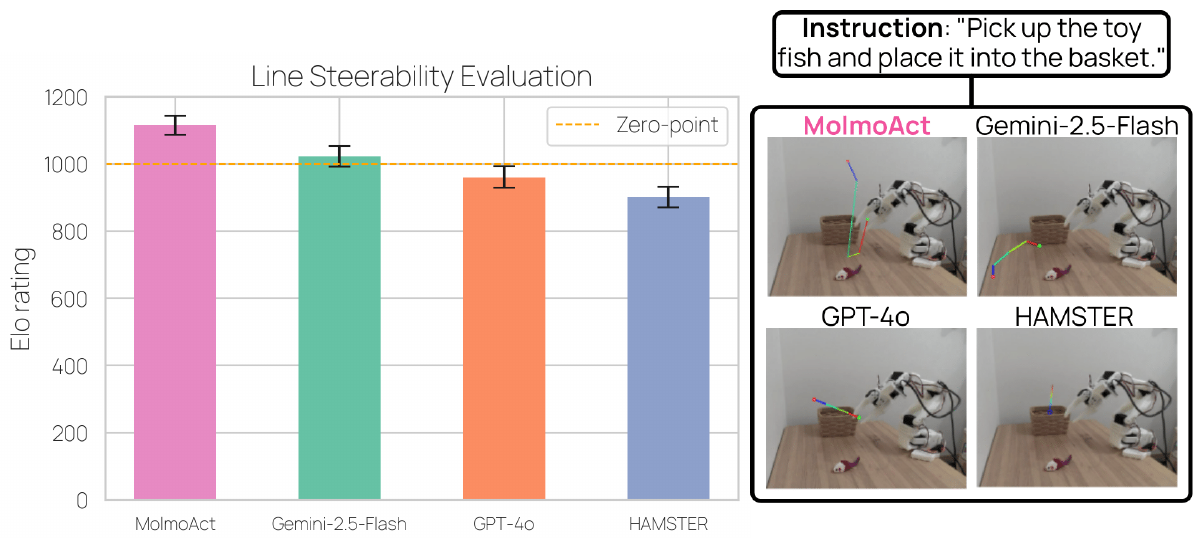}
  \caption{\textbf{Line steerability evaluation across models.}
\textbf{Left:} Elo ratings show that \molmoact achieves the highest performance, surpassing Gemini-2.5-Flash, GPT-4o, and HAMSTER, with error bars indicating 95\% confidence interval (CI).
\textbf{Right:} Example qualitative results showing predicted visual traces overlaid on robot camera views.}
  \label{fig:fig7_line_generalize}
\end{figure*}

\begin{figure*}[t]
  \centering
  \includegraphics[width=\textwidth]{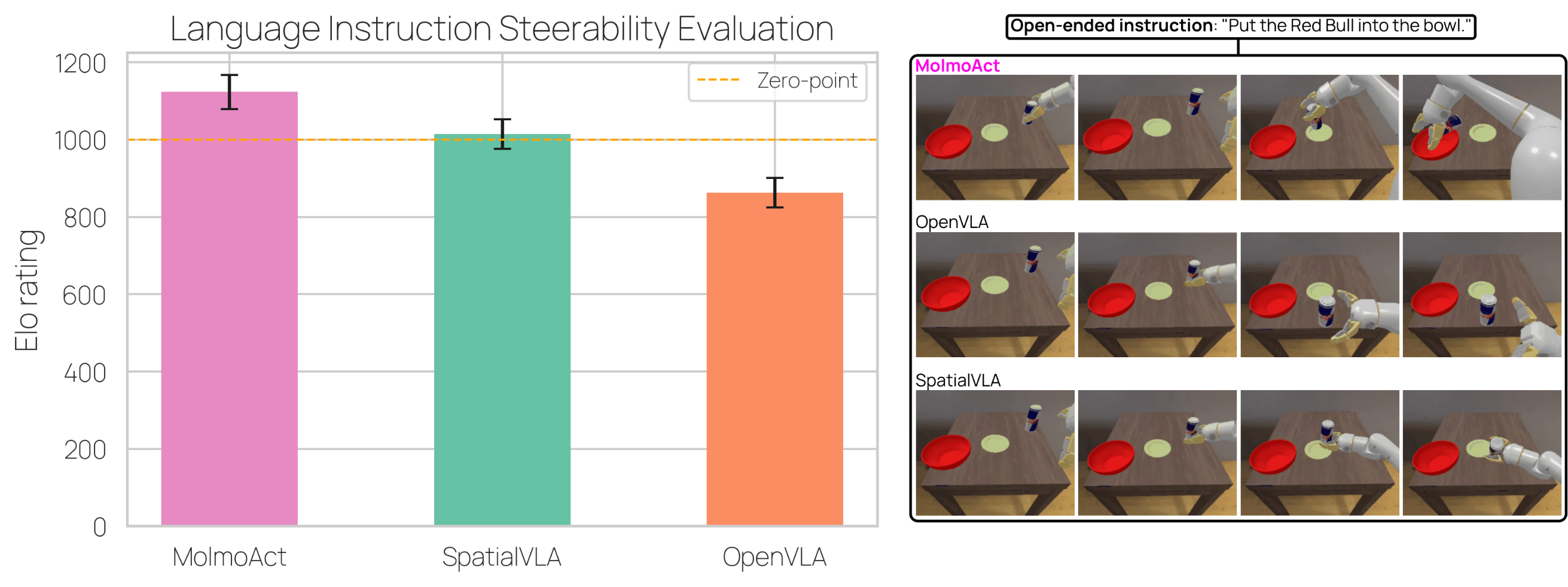}
  \caption{\textbf{Language Instruction Evaluation.} \textbf{Left:} Elo ratings for three models based on human votes in a head-to-head instruction-following evaluation. \textbf{Right:} Qualitative comparison of execution traces for the open-ended instruction ``Put the redbull into the bowl." \molmoact aligns more closely with the intended instruction than other models.}
  \label{fig:fig8_lang_generalize}
\end{figure*}

\subsection{Effectiveness of \molmoact in Out-of-Distribution Generalization } 

We evaluate \molmoact in both simulation and real-world settings to assess its ability to generalize beyond the training data distribution, both in zero-shot and fine-tuned regimes. In simulation, we follow the \simpler variant-aggregation protocol, which introduces distribution shifts through changes in lighting, textures, and camera viewpoints. We compare \molmoactdpre and its RT-1 fine-tuned variant against several state-of-the-art generalist policies—TraceVLA, RT-1X, OpenVLA, RoboVLM, Emma-X, \pizerofast, and SpatialVLA. For real-world evaluation, we test \molmoactd using a single Franka arm on a multi-task setup involving three objects and two different-colored plates arranged on a tabletop. We collect over 300 tele-operated demonstrations spanning three task types, then post-train \molmoactd and baselines in a multi-task setting. During evaluation, we test generalization in four aspects: (1) \textbf{Language Variation} — rephrased instructions, (2) \textbf{Spatial Variation} — changes in target object position, (3) \textbf{Distractors} — addition of irrelevant objects, and (4) \textbf{Novel Objects} — substitution of target objects with unseen ones. We benchmark \molmoactd against \pizerofast and OpenVLA, testing three variants per task and four trials per variant. Full task details are presented in \autoref{supp:eval:generalize}.

\textbf{Evaluation Results.}
In simulation, fine-tuned \molmoactdpre achieves 72.1\% on the variant aggregation tasks as shown in Table \ref{tab:simplerenvgooglerobot}, outperforming all baselines and exceeding the second-best model, RT-2-X, by 7.8\%. The performance difference between variant aggregation and visual matching is less than 1\%, highlighting \molmoact's robustness to visual and distributional shifts. In the real world, \molmoactd consistently surpasses all baselines across all generalization axes, achieving a 23.3\% average improvement in task progression over \pizerofast as shown in Figure \ref{fig:fig6_1_real_eval_outdist}.

\subsection{Effect of the \molmoactdata on \molmoact Performance}

\textbf{Evaluation Setups and Baselines.}
To assess the effectiveness of mid-training with the \molmoactdata, we conducted real-world experiments on three curated tasks that go beyond simple pick-and-place: \texttt{close\_lid}, \texttt{rotate\_pot}, and \texttt{pour\_tea}. For each task, we collected 50 demonstrations and trained four models: \molmoactd, \molmoactd without mid-training, \pizerofast, and OpenVLA. Each model was then evaluated over 10 trials per task. Task details are shown in \autoref{supp:eval:ablation}.

\textbf{Evaluation Results.} Based on the real-world ablation studies shown in Figure \ref{fig:fig6_2_real_eval_molmoact_dataset}, \molmoactd outperforms its counterpart without mid-training by an average margin of 5.5\% across the three tasks, demonstrating that mid-training on the \molmoactdata yields a consistent performance boost of around 5\%. Even without mid-training, \molmoactdpre surpasses \pizerofast and OpenVLA by 14.8\% and 10.9\%, respectively.

\subsection{Instruction Following of \molmoact}

We evaluated \molmoact's ability to follow natural language instructions in two settings: (i) executing tasks with open-ended commands in simulation, and (ii) generating visual traces conditioned on language prompts. For the first one, we curated five manipulation scenarios in the \simpler environment using a Google Robot, each involving novel out-of-distribution objects. Ten participants provided 29 open-ended instructions (e.g., \emph{``Put the redbull into the bowl."}). We compared \molmoactdpre to \textbf{SpatialVLA} and \textbf{OpenVLA}, both pre-trained on the OXE dataset. For each instruction, the models generated rollouts, which were evaluated in a head-to-head arena-style web interface. Human annotators ($n{=}100$) selected which rollout best matched the instruction. We collected over 1,500 votes, which were converted into Elo ratings (see Figure~\ref{fig:fig8_lang_generalize}). For visual trace generation, 10 participants wrote 87 language prompts for 30 internet-sourced images depicting tabletop and mobile manipulation scenarios. \molmoactdpre was evaluated against \textbf{Gemini-2.5-Flash}, \textbf{GPT-4o}, and \textbf{HAMSTER}—a VLM fine-tuned for trace generation. Participants voted in a similar blind arena interface, resulting in over 1,000 votes. Details with our curated manipulation scenes and instructions provided by participants are shown in \autoref{supp:eval:follow}.

\textbf{Evaluation Results.}  
\molmoactdpre achieved the highest Elo rating in the simulation instruction-following task, outperforming SpatialVLA by 109 points and OpenVLA by an even larger margin. Pairwise win rates also show that \molmoactdpre winning over SpatialVLA in 58\% of comparisons and over OpenVLA in 81\%. A sample rollout comparison for the instruction \emph{``Put the redbull into the bowl."} is shown on the right in \autoref{fig:fig8_lang_generalize}. In the visual trace task, \molmoactdpre again outperformed all baselines, achieving significantly higher Elo scores with non-overlapping 95\% confidence intervals, demonstrating strong language-grounded generalization in both action execution and trace generation as shown in Figure \ref{fig:fig7_line_generalize}.

\begin{figure*}[t]  % spans both columns
  \centering
  \includegraphics[width=\textwidth]{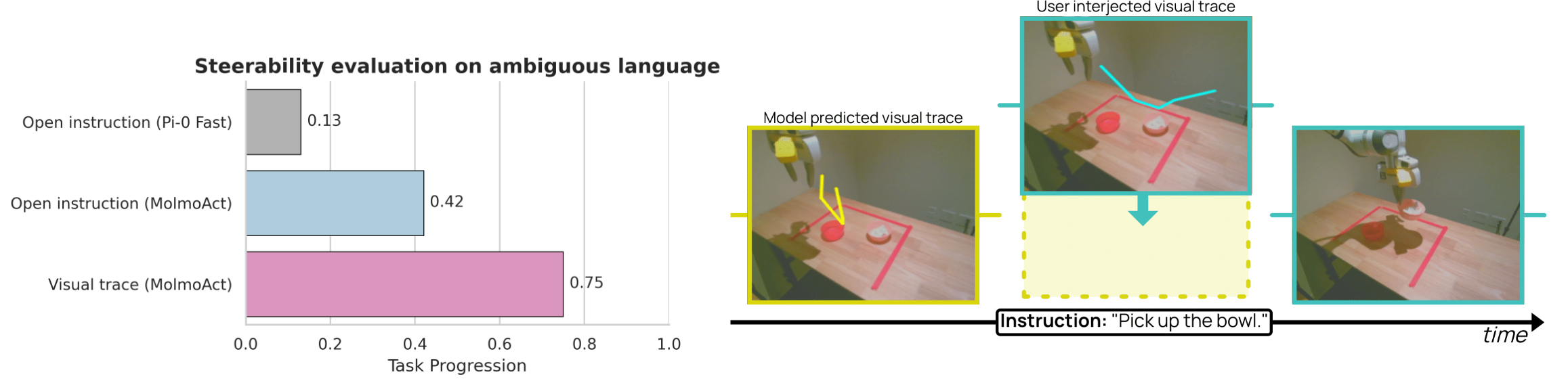}
  \caption{\textbf{Steerability evaluation with open instructions and visual traces.}
\textbf{Left:} Success rates for different steering modes, showing that \molmoact with visual trace steering achieves the highest success rate (0.75), outperforming its open-instruction variant and \pizerofast.
\textbf{Right:} Example of the \emph{"Pick up the bowl"} task: the model-predicted trajectory (yellow) is adjusted via a user-provided steering trajectory (cyan), resulting in the corrected task completion.}
  \label{fig:fig9_steerability}
\end{figure*}

\subsection{Steerability of \molmoact}

\textbf{Evaluation Setups and Baselines.} We aim to evaluate \molmoact's ability to steer robot actions, particularly when initial language instructions are ambiguous. Specifically, we investigate the effectiveness of different interaction mediums in guiding \molmoact toward user-intended targets during task execution. For this purpose, we set up a \texttt{pick\_up\_bowl} task, post-training \molmoactd and the baseline model (\pizerofast) with 100 collected demonstrations, each annotated with two distinct language instructions: one specifying the clean bowl and the other the dirty bowl, as depicted in Figure \ref{fig:fig9_steerability}. During evaluation, we first provide ambiguous instructions such as "pick up (the) bowl," prompting \molmoactd to predict an initial trajectory towards one of the bowls. Subsequently, we test two steering methods: visual trace sketches to visually instruct the model toward the alternative bowl, and open-ended natural language instructions provided interactively by participants (N=10) that are different from the ground-truth instruction. For comparison, we also attempt to steer the actions of \pizerofast by changing language instructions at test-time. Each model is evaluated in 15 trials, and the performance is evaluated according to the progression of the task. For more details about the setting, please refer to \autoref{supp:eval:steer}.

\textbf{Evaluation Results.} 
Based on our experiments, we observed that \molmoactd is notably more steerable via visual trace inputs, achieving a success rate of 75\%. Additionally, steering using visual traces significantly outperforms steering via open-ended natural language instructions by a margin of 33\%. Lastly, we demonstrate that \molmoactd exhibits superior instruction-following capabilities compared to the baseline model, \pizerofast. Specifically, when steering robot actions using open-ended language instructions, \molmoact surpasses \pizerofast by a substantial margin of 29\%, highlighting its enhanced instruction-following capabilities to user commands.

\section{Related Work}
\label{sec:related}

\subsection{Generalist robot manipulation policies}
Recent advances in robotic manipulation have shifted from training narrow, single-task specialists to learning from large, diverse datasets spanning many scenes, tasks, and embodiments\citep{berscheid2019robot,brohan2022rt,dasari2019robonet,ebert2021bridge,fang2023rh20t,jang2022bc,khazatsky2024droid,mandlekar2018roboturk,walke2023bridgedata,shafiullah2023bringing}. This shift has enabled policies that not only excel within their training distribution but also generalize to out-of-distribution scenes, environments, language instructions, and novel objects \citep{pumacay2024colosseum,xie2024decomposing,lin2024data}. Much of this progress has been fueled by Large Language Models (LLMs) \citep{o2024open,achiam2023gpt,groeneveld2024olmo,touvron2023llama} and Vision-Language Models (VLMs) \cite{team2024gemini, liu2023visual,deitke2024molmo}, giving rise to the paradigm of Vision-Language-Action models (VLAs) \citep{black2410pi0, brohan2022rt,zheng2024tracevla,zhao2025cot,team2025gemini,li2024cogact,qu2025spatialvla,kim2024openvla}. VLAs pretrain a VLM backbone on large-scale web data to capture rich semantic world knowledge, then fine-tune it for downstream robot control. While most share similar backbones, they differ in their action heads—employing flow matching\citep{wen2025tinyvla}, diffusion\citep{wen2025dexvla,liu2024rdt}, or advanced action tokenization schemes\citep{pertsch2025fast}. While some adopt hierarchical designs, where a robotics-focused VLM \citep{bjorck2025gr00t,li2025hamster,shentu2024llms} outputs intermediate representations for pre-trained, task-specific policies to improve generalization. However, a major bottleneck for these models is their heavy reliance on large amounts of robotics-specific data, often collected via tele-operation. In contrast, \molmoact aims to leverage reasoning in space to train an action reasoning model that achieves competitive or superior performance with only a fraction of the data required by existing methods.

\subsection{Robot reasoning and planning with language}
In recent years, numerous works have demonstrated that augmenting end-to-end robotic policies with high-level reasoning—either by integrating LLMs or VLMs directly into robotic systems, or by incorporating their reasoning outputs into policies—can substantially improve performance on long-horizon tasks and enhance generalization \citep{ahn2022can,huang2023voxposer,bharadhwaj2024roboagent,fang2025sam2act,liu2024ok,shi2024yell,wang2024inference,gu2023rttrajectory}.
An alternative line of research seeks to decouple perception and reasoning from low-level control, assigning VLMs the role of performing semantic prediction or generating intermediate representations such as task plans, scene graphs, or spatial layouts \citep{duan2024manipulate,liu2024moka,li2025hamster,huang2024rekep,liang2022code,singh2022progprompt,duan2024aha}. Execution is then handled by a separate low-level policy or control module that interprets these high-level outputs and converts them into robot actions.

Action prediction from \molmoact can be steered through both natural language and an interactive visual reasoning-trace sketch interface. This dual-modality control improves explainability and enables more effective diagnosis of model behavior. While prior methods such as RT-Trajectory \citep{gu2023rttrajectory}, HAMSTER \citep{li2025hamster}, and inference-time policy steering \citep{wang2024inference} also offer forms of policy steerability, they differ in important ways. RT-Trajectory and inference-time policy steering are tightly coupled to the architectural constraints and training regimes of robotics transformers or diffusion models, and therefore lack the broader semantic generalization provided by pre-training on a VLM backbone. HAMSTER enables language-conditioned trajectory steering but outputs only 2D trajectories by the high-level VLM, with execution handled by a low-level policy trained on a fixed set of tasks. In contrast, \molmoact generalizes its steering to novel spatial configurations, previously unseen objects, and even ambiguous language instructions for a diverse set of tasks, offering a more versatile and semantically grounded control interface with users.

\subsection{Embodied reasoning for robotic manipulation}

Chain-of-thought (CoT) prompting \citep{wei2022chain} has significantly improved the multi-step reasoning capabilities of LLMs across domains such as mathematics, programming, and question answering. This idea has also been extended to the visual domain through multimodal CoT \citep{bigverdi2025perception,zhang2023multimodal}, where visual information is processed iteratively and reasoned over in conjunction with images. Motivated by these advances, recent works in robotics have explored extending reasoning capabilities to embodied tasks within vision-language-action (VLA) models.

ECoT \citep{zawalski2024robotic} synthesizes intermediate subgoals via prompting and uses supervised fine-tuning to teach VLAs to reason before acting. CoT-VLA \citep{zhao2025cot} replaces linguistic CoT with visual subgoal frames generated prior to action prediction. RAD \citep{clark2025action} which leverages action-free human video to curate language-based reasoning to guide low-level actions. ThinkAct \citep{huang2025thinkact} leverages action-aligned reinforcement learning and visual latent planning to connect embodied reasoning with real-world action prediction in VLAs. Most similar to our reasoning-in-space approach are Emma-X \citep{sun2024emma}, which autoregressively fine-tunes OpenVLA with reasoning data formatted as subtasks, predicted future gripper states in 2D, and 3D spatial movement coordinates or \citep{yang2025bridging}, which focuses on different forms of mid-level representations, including a trajectory trace with depth awareness, however they only evaluated for diffusion policy on a small data region.

However, unlike ECoT, CoT-VLA, RAD, and ThinkAct—whose reasoning is represented as latent embeddings, generated sub-goals, or textual descriptions that are difficult to ground in the real world and lack the precision required for manipulation—\molmoact grounds every step of its reasoning chain directly in the scene. Compared to Emma-X, which reasons primarily over predicted gripper positions without leveraging the full 3D scene context, \molmoact performs reasoning in space, where each step can be decoded and visualized both on the image plane and within the 3D environment. This explicit spatial grounding improves explainability and enhances action prediction within a chain-of-thought prompting framework.

\section{Conclusion}
\label{sec:conclude}

We introduced \molmoact, a family of fully open action reasoning models that integrate perception, planning, and control by reasoning in space. By combining depth perception tokens, visual reasoning traces, and action prediction, \molmoact produces explainable, spatially coherent behaviors. The behaviors can be executed directly, or steered via trajectory editing. Our evaluations across simulation and real-world settings demonstrate that \molmoact consistently outperforms strong vision–language–action baselines, adapts efficiently to novel tasks and embodiments through lightweight fine-tuning, and generalizes robustly to out-of-distribution conditions. We release all model weights, code, and data, including the \molmoactdata, to enable reproducibility and foster community-driven research toward building foundation models that transform perception into purposeful action through structured reasoning.

%%%%% ACKKNOWLEDGEMENT %%%%%

\clearpage

\section*{Author Contributions}
\label{sec:contrib}
This project was made possible through the equal contributions of all three co-first authors in no particular order. Their individual contributions are as follows:
\begin{itemize}
    \item \textbf{Jason Lee}: Led the Action Reasoning data curation for pre-training, post-training and \molmoactdata; Developed simulation evaluation infrastructure for pre-training and post-training; Led the real-world evaluations and simulation evaluations; contributed to real-world data collection and paper writing.
    \item \textbf{Jiafei Duan}: Led the project and ideated the core method design; proposed and curated the \molmoactdata; Led the paper’s writing; and designed and conducted both simulation and real-world evaluations, including ablation studies.
    \item \textbf{Haoquan Fang}: Led the implementation of the model, training, and inference pipeline and infrastructure; designed and developed the steerability feature and evaluation framework; contributed to data curation, simulation/real-world evaluations, and paper writing.
\end{itemize}

All other contributors are also deeply appreciated for their effort, which is critical to the success of the \molmoact project. As not all of these can be captured, we indicate their primary contributing role in \molmoact:
\begin{itemize}
    \item For real-world robot infrastructure: Yuquan Deng, Shuo Liu, and Boyang Li.
    \item For real-world data collection and evaluation: Yuquan Deng, Bohan Fang, Shuo Liu, Boyang Li, and Angelica Wu.
    \item For paper writing and figures: Jiafei Duan, Yi Ru Wang, Jason Lee, Haoquan Fang, Jieyu Zhang, Winson Han, Eli
VanderBilt, and Ranjay Krishna.
    \item For project management: Karen Farley.
    \item For research advisory: Ranjay Krishna, Dieter Fox, Rose Hendrix, and Ali Farhadi.
    \item Project PI: Ranjay Krishna 
\end{itemize}

\section*{Acknowledgment}

This work would not be possible without the support of our colleagues at Ai2:

\begin{itemize}
    \item We thank Christopher Clark, Abhay Deshpande, Yejin Kim, Mahtab Bigverdi, Joel Jang and Rohun Tripathi for helpful research discussions and sharing of relevant findings across related projects.
    \item We thank for David Albright, Crystal Nam, Kristin Cha, Sophie Lebrecht, Taira Anderson, Kyle Wiggers, Kelsey MacMillan, Katie Morigi, and Megan Bartot for project management, support to robot room and publicity of \molmoact
    \item We thank Yoganand Chandrasekhar, Johann Dahm, Fangzhou Hu, and Caroline Wu for their work on the Ai2 cluster.

\end{itemize}

\molmoact would not have been possible without the support of many other institutions. In particular, we thank Google for their support in setting up the training environment for MolmoAct and to Cirrascale for their
on-going support of Ai2’s cluster. Jiafei Duan is supported by the Agency for Science, Technology and Research (A*STAR) National
Science Fellowship.

\clearpage

%%%%%%%%% BIB %%%%%%%%%
\bibliographystyle{abbrvnat}
\bibliography{neurips_2023}

\clearpage

%%%%%%%%% APPENDIX %%%%%%%%%
\appendix

\section*{Appendix}
The appendix includes the following sections:
\begin{itemize} 
\itemsep0em 
    \item \S\ref{supp:model} - Model Details
    \item \S\ref{supp:training} - Training Details
    \item \S\ref{supp:actionvocab} - Action Vocabulary
    \item \S\ref{supp:eval} - Evaluation Details
    \item \S\ref{supp:data} - Data Details
    \item \S\ref{supp:dataexamples} - Dataset Examples
    \item \S\ref{supp:limit} - Limitations and Potential Solutions

    % \item \S\ref{supp:related_work} - Related Work
\end{itemize}

\section{Model Details}
\label{supp:model}

This section summarizes the \molmoact model architecture, which inherits Molmo with slight modification. The design combines a pre-processor for multi-scale cropping and optionally image padding, a ViT image encoder, a vision–language connector, and a LLM.

\subsection{Backbone Overview}
\molmoact has the following parts:
\begin{enumerate}
    \item \textbf{Pre-processor:} converts each input image into one low-resolution crop and several high-resolution crops.
    \item \textbf{ViT Image Encoder:} encodes each crop independently into per-patch features.
    \item \textbf{Vision–language Connector:} pools and projects patch features into the LLM embedding space.
    \item \textbf{LLM:} autoregressively processes vision and text tokens.
\end{enumerate}
From this template \molmoact instantiates a family of models by selecting a vision encoder and an LLM while keeping the training recipe mainly consistent. Vision encoders include OpenAI ViT-L/14 336px CLIP and ViT-SO400M/14 384px SigLIP2. LLM backbones include fully open OLMo-2-1124-7B and open-weight Qwen2.5-7B. With the combination of ViT-SO400M/14 384px SigLIP2 with Qwen2.5-7B, we have \molmoactd, our best and demo model. With the combination of OpenAI ViT-L/14 336px CLIP with OLMo-2-1124-7B, we have \molmoacto, our most open model. Note that although OpenAI ViT-L/14 336px CLIP uses closed data, it can be reproduced from scratch, as shown by MetaCLIP~\citep{metaclip}.

\subsection{Image Encoding and Cropping}
Most ViTs accept square images at a fixed resolution, which is insufficient for fine-grained details. This mainly applies to the multimodal web data, as much of the robot data is not high-resolution, and there is also work~\citep{kim2024openvla} showing that image resolution doesn't affect much for robot control. To make \molmoact more general, it still inherits the way Molmo addresses the high-resolution problem by tiling each image with multiple square high-resolution crops plus a resized low-resolution full image. Cropping proceeds as follows.

\paragraph{Grid Selection and Overlap.}
A rectangular grid (e.g., \(2\times2\), \(3\times1\)) is chosen so each grid cell matches the ViT input size. The grid squares are then moved closer together to introduce a fixed overlap margin (default 4 patches, approximately 56 pixels), which supplies border patches with neighbor context. Features from overlapping pixels are \emph{not} forwarded to the connector or LLM, so the resulting tokens exactly tile the high-resolution image. Although overlap slightly reduces the effective tiled resolution, this can be offset by using more crops, and empirically improves performance.

\paragraph{Resizing and Padding.}
For OpenAI CLIP vison encoder, we follow the way Molmo~\citep{deitke2024molmo} does to resize and pad the image to keep its original aspect ratio before processing. The scheme is the following. The image is upscaled to fit the grid while preserving aspect ratio, choosing the scale that minimizes upscaling; ties are broken by minimizing the overall size. A maximum number of high-resolution crops is enforced. If covering the image would exceed this limit, the image is downscaled to fit. In all cases, the image is padded with black borders so each crop is square and aligned to the grid. The low-resolution crop is produced by resizing and padding the full image to the ViT’s native resolution. Each crop is encoded independently by the ViT and connector to produce patch features. A learned embedding indicating the crop’s padding status (no padding, some padding, or all padding) is added to the patch features so the model can distinguish natural black regions from artificial padding.

Note that for SigLIP2 vision encoder, we use the standard way to resize all image inputs to a square image without padding, which follows the original transform in SigLIP2 training.

\subsection{Vision–language Connector}
After ViT encoding, Molmo aggregates features in two steps:
\begin{enumerate}
    \item \textbf{Layer selection and concatenation:} features from the third-to-last (OpenAI CLIP) or fourth-to-last (SigLIP2) and the tenth-from-last ViT layers are concatenated for each patch; this slightly outperforms using a single layer as shown by Molmo~\citep{deitke2024molmo}.
    \item \textbf{Attention pooling in \(2\times2\) windows:} within each \(2\times2\) patch window, a multi-headed attention layer pools the four patches to a single vector, using the mean of the patches as the query. This pooling reduces sequence length while preserving local spatial structure and outperforms naive concatenation as shown by Molmo~\citep{deitke2024molmo}.
\end{enumerate}
Pooled features are then mapped to the LLM embedding space with a small MLP.

\subsection{Arranging Vision Tokens}
Pooled patch features (vision tokens) are serialized left-to-right and top-to-bottom. Tokens from the low-resolution full image appear first, followed by high-resolution crop tokens arranged in row-major order. Special tokens mark the start and end of both the low- and high-resolution sequences. Row-end tokens are inserted between rows to indicate row transitions.

\subsection{Multi-image Inputs}
Molmo~\citep{deitke2024molmo} itself doesn't provide the capability to take in multi-image inputs. We implement this in a straightforward way: we process all the images to vision tokens in the same way as mentioned above, then we append index tokens at the beginning of the vision tokens of each image, and we finally concatenate all images together as the input. The index tokens are just text tokens of \texttt{"Prefix i"}, where \texttt{i} stands for the ith image.

\subsection{Full Hyperparameters}

The full hyperparameters of \molmoact architecture are shown in \autoref{tab:hyperparams_model}. Note that for LoRA implementation, adapters are applied to all linear layers in the model.

\clearpage

\section{Training Details}
\label{supp:training}

\subsection{Implementation}
Our training implementation mainly follows Molmo. We train in PyTorch using Fully Sharded Data Parallel (FSDP), and use PyTorch’s Scaled Dot-Product Attention (SDPA) attention implementation. For numeric precision, we enable Automatic Mixed Precision (AMP) with bfloat16 for most operations. As an exception, we compute layer normalization and Rotary Position Embeddings (RoPE) in fp32.

With FSDP, each GPU forms a local mini-batch, computes gradients, and then we average the gradients across devices. When normalizing the loss on each device, we divide the device’s total loss by the global average number of loss-tokens per example across all devices, rather than by the device-local count. This avoids a subtle bias that can arise when examples with up-weighting also contain fewer loss-tokens (e.g., shorter responses) and happen to co-occur on devices with smaller token counts. Using the global average corrects this mismatch and is especially important when the global batch is much larger than any single device batch.

For parameter-efficient fine-tuning, we do not shard LoRA adapter parameters under FSDP. Instead, each GPU keeps a full copy of the LoRA parameters, and we register a gradient hook on those tensors to synchronize their gradients across ranks before the optimizer step. Because LoRA adds only a small fraction of the total parameters, this replication has negligible memory and communication overhead while simplifying the training setup and avoiding sharding edge cases for the adapters.

Batches mix examples from multiple tasks. We cap the sequence length at 2304 tokens for both pre-training and fine-tuning, truncating only when necessary (e.g., heavily annotated synthetic data or rare outliers). Training is stable under this recipe—without loss spikes or NaNs—which we attribute in part to initializing from pre-trained models.

To enable the model to learn to understand and output depth perception tokens, we follow the training scheme of LLaVA-AURORA~\citep{bigverdi2025perception} by unfreezing the tokenizer embedding and lm head. For \molmoactd, which uses Qwen2.5-7B, we simply replace the first 130 padding tokens with the depth perception tokens \(\{\langle\mathrm{DEPTH\_START}\rangle,\langle\mathrm{DEPTH\_END}\rangle\}\;\cup\;\{\langle\mathrm{DEPTH}\_k\rangle\}_{k=1}^{128}\). However, for \molmoacto, which uses Olmo2-7B, since it has less than 130 padding tokens, we first pad the tokenizer and lm head to its next multiple of 512, then replace the first 130 tokens with our depth perception tokens in the same way as \molmoactd.

All of our collected data used for the mid- and post-training stages is recorded at 640$\times$480 px, which triggers the high-resolution cropping procedure described in \autoref{supp:model}. By contrast, the OXE robot data used for pre-training has lower resolution, so no high-res crop is applied. To match OXE during pre-training, we downscale our collected images from 640$\times$480 to 320$\times$240 px while preserving the original aspect ratio. This alignment also reduces the number of vision tokens and accelerates training.

Full training hyperparameters and information are shown in \autoref{tab:hyperparams_train}. Note that for post-training, we train the model until it fully converges, which is determined by its training loss and evaluation performance. Therefore, training steps largely vary across different tasks and scenarios. We will show the training details for post-training in different tasks in later sections.

\subsection{GPU Cluster}

\molmoact was trained on Jupiter, an Ai2 GPU cluster in Austin, Texas. \molmoact workloads were scheduled using Beaker \citep{guerquin2022beaker}, a custom workload management system. Jupiter comprises 128 GPU nodes and is operated by Cirrascale Cloud Services\footnote{\href{https://www.cirrascale.com/}{\path{cirrascale.com}}}.

\paragraph{Compute} Jupiter provides 1{,}024 NVIDIA H100 GPUs (80GB HBM3, 700W) across 128 servers. Each server has 2,$\times$,Intel Xeon Platinum8468 CPUs, 2TB DDR5 system memory, and 18~TB local NVMe storage.

\paragraph{Storage} The servers are connected over an 800Gbps local network to a WEKA high-performance storage cluster\footnote{\href{https://www.weka.io/}{\path{weka.io}}}. The storage system provides 1PB of NVMe SSD across 11 storage servers and 5PB of HDD across 12 hosts. Each Jupiter server has two bonded 25Gbps Mellanox Ethernet NICs (50Gbps per host). In benchmarks, we achieved 761Gbps aggregate read/write throughput using 64 client machines.

\paragraph{Interconnect} Cross-node GPU communication uses RDMA over InfiniBand on a two-tier Rail-Optimized, balanced, full-bisection network \citep{wang2023rail}. Each server is equipped with eight 400Gbps InfiniBand adapters (3.2Tbps peak per host), supporting concurrent distributed jobs without topological scheduling.

\paragraph{Cooling} The servers are racked in \textit{Dynamic Density Cabinets}\footnote{\href{https://www.cirrascale.com/products-and-services/cabinet-technologies}{\path{cirrascale.com/products-and-services/cabinet-technologies}}}. Each cabinet houses five servers with dedicated cooling and power. Air circulates in a closed loop through an overhead plenum where it is cooled via heat transfer to water, enabling a datacenter PUE of~1.2. Under heavy utilization, H100 temperatures peak around $75^\circ\mathrm{C}$, with typical averages between $60^\circ\mathrm{C}$ and $65^\circ\mathrm{C}$.

% ---------- 1)MODEL ARCHITECTURE  ------------------------------------------
% Needs \usepackage{booktabs,xcolor,array}
\begin{table}[!htbp]
\centering
\footnotesize
\setlength\tabcolsep{4pt}
\renewcommand\arraystretch{1.1}

\begin{tabular}{l
                *{2}{>{\centering\arraybackslash}p{1.5cm}}
                *{2}{>{\centering\arraybackslash}p{1.5cm}}
                *{2}{>{\centering\arraybackslash}p{1.5cm}}}
\toprule
\multicolumn{1}{c}{} &
\multicolumn{2}{c}{\cellcolor{olive!5}\textbf{Image Encoder}} &
\multicolumn{2}{c}{\cellcolor{olive!5}\textbf{V/L Connector}} &
\multicolumn{2}{c}{\cellcolor{olive!5}\textbf{LLM}} \\[-0.2em]
\cmidrule(lr){2-3}\cmidrule(lr){4-5}\cmidrule(lr){6-7}
\textbf{Parameter} & 7B-D & 7B-O & 7B-D & 7B-O & 7B-D & 7B-O \\
\midrule
Params        & 383M & 278M & 121M & 75M & 7.6B & 7.3B \\
Dim           & 1152 & 1024  & — & —   & 3584 & 4096 \\
MLP Dim       & 4304 & 4096  & 37888 & 22016 & 37888 & 22016 \\
Activation    & GELU & GELU  & SwiGLU & SwiGLU & SwiGLU & SwiGLU \\
Heads         & 16 & 16 & 16 & 16 & 28 & 32 \\
KV Heads      & 16 & 16 & — & — & 4 & 32 \\
Layers        & 27 & 23 & — & — & 28 & 32 \\
Image Size    & $384{\times}384$ & $336{\times}336$ & — & — & — & — \\
Patch Size    & 14 & 14 & — & — & — & — \\
Pool Size     & — & — & $2{\times}2$ & $2{\times}2$ & — & — \\
Pool Dim      & — & — & 1152 & 1024 & — & — \\
Pool Heads    & — & — & 16 & 16 & — & — \\
Theta         & — & — & — & — & 1M & 0.5M \\
Dropout       & 0.0 & 0.0 & 0.0 & 0.0 & 0.1 & 0.1 \\
\addlinespace[2pt]  
\bottomrule
\end{tabular}
\caption{\textbf{\molmoact's Architecture Hyperparameters.} We specify all hyperparameter information for the different model architectures for \molmoactd and \molmoacto.}
\label{tab:hyperparams_model}
\end{table}

\begin{table}[!htbp]
\centering
\footnotesize
\setlength\tabcolsep{4pt}
\renewcommand\arraystretch{1.1}

\begin{tabular}{l
                *{2}{>{\centering\arraybackslash}p{1.7cm}}
                *{2}{>{\centering\arraybackslash}p{1.7cm}}
                *{2}{>{\centering\arraybackslash}p{1.7cm}}}
\toprule
\multicolumn{1}{c}{} &
\multicolumn{2}{c}{\cellcolor{olive!5}\textbf{Pre-train}} &
\multicolumn{2}{c}{\cellcolor{olive!5}\textbf{Mid-train}} &
\multicolumn{2}{c}{\cellcolor{olive!5}\textbf{Post-train}} \\[-0.2em]
\cmidrule(lr){2-3}\cmidrule(lr){4-5}\cmidrule(lr){6-7}
\textbf{Parameter} & 7B-D & 7B-O & 7B-D & 7B-O & 7B-D \\
\midrule
Warm-up ViT        & 200 & 200 & 200 & 200 & 200  \\
Warm-up Conn.      & 200 & 200 & 200 & 200 & 200  \\
Warm-up LLM        & 200 & 200 & 200 & 200 & 200  \\
LR ViT             & $1{\times}10^{-5}$ & $1{\times}10^{-5}$ & $5{\times}10^{-6}$ & $5{\times}10^{-6}$ & $5{\times}10^{-4}$ \\
LR Conn.           & $1{\times}10^{-5}$ & $1{\times}10^{-5}$ & $5{\times}10^{-6}$ & $5{\times}10^{-6}$ & $5{\times}10^{-4}$ \\
LR LLM             & $2{\times}10^{-5}$ & $2{\times}10^{-5}$ & $1{\times}10^{-5}$ & $1{\times}10^{-5}$ & $5{\times}10^{-4}$ \\
Cosine Decay       & 10\% & 10\% & 10\% & 10\% & 10\% \\
Eps.               & $10^{-6}$ & $10^{-6}$ & $10^{-6}$ & $10^{-6}$ & $10^{-6}$ \\
Betas              & 0.9/0.95 & 0.9/0.95 & 0.9/0.95 & 0.9/0.95 & 0.9/0.95 \\
LoRA Rank          & — & — & — & — & 32 \\
LoRA Alpha         & — & — & — & — & 16 \\
LoRA Dropout       & — & — & — & — & 0  \\
LoRA Bias         & — & — & — & — & None \\
Multi-image Input  & No & No & Yes & Yes & Yes \\
Steps              & 100k & 100k & 50k & 50k & Varies \\
Global Batch Size  & 512  & 512  & 256  & 256  & 64 (real) or 128 (sim) \\
GPUs (H100s)       & 256 & 256 & 128 & 128 & 32 (real) or 64 (sim) \\
Time (Hours)       & 38 & 32 & 18 & 15 & Varies \\
GPU Hours          & 9728 & 8192 & 2304 & 1920 & Varies \\
\addlinespace[2pt]  
\bottomrule
\end{tabular}
\caption{\textbf{\molmoact's Training Hyperparameters.} We specify all hyperparameter information for different training schemes for \molmoactd and \molmoacto. Note that for \molmoactdpre, we train the model with 150K steps, but it reaches better performance at 100K steps.}
\label{tab:hyperparams_train}
\end{table}

\clearpage

\section{Action Tokenization}
\label{supp:actionvocab}

We provide our full action vocabulary in \autoref{tab:action_vocab_1} and \autoref{tab:action_vocab_2}, which show the mapping from discrete bin index to its corresponding action token. Note that the string \texttt{\textbackslash{}u00e2\textbackslash{}u00bd\textbackslash{}u0139} is a sequence of Unicode escape codes. Each \texttt{\textbackslash{}uXXXX} gives one code point in hexadecimal. When decoded, those code points become the actual characters, concatenated in order.

% NOTE: Compile with XeLaTeX or LuaLaTeX to render UTF-8 token glyphs.
\begin{table*}[!htbp]
\centering
\footnotesize
\setlength{\tabcolsep}{1pt}
\setlength{\arrayrulewidth}{0.1pt}
\renewcommand{\arraystretch}{1.1}
\caption{\textbf{Action token vocabulary}: Mapping from discrete bin index (0 to 127) to the actual token string.}
\label{tab:action_vocab_1}
\begin{tabular}{r l r l r l r l}
\toprule
\textbf{Bin} & \textbf{| Action Token} & \textbf{Bin} & \textbf{| Action Token} & \textbf{Bin} & \textbf{| Action Token} & \textbf{Bin} & \textbf{| Action Token} \\
\midrule
0 & \texttt{\textbackslash{}u00e2\textbackslash{}u00bd\textbackslash{}u0139} & 1 & \texttt{\textbackslash{}u00e2\textbackslash{}u00ba\textbackslash{}u0141} & 2 & \texttt{\textbackslash{}u00e2\textbackslash{}u012f\textbackslash{}u00a8} & 3 & \texttt{\textbackslash{}u00e1\textbackslash{}u0137\textbackslash{}u00b7} \\
4 & \texttt{\textbackslash{}u00ef\textbackslash{}u00a8\textbackslash{}u012c} & 5 & \texttt{\textbackslash{}u00e3\textbackslash{}u0129\textbackslash{}u00bd} & 6 & \texttt{\textbackslash{}u00e3\textbackslash{}u0129\textbackslash{}u00ba} & 7 & \texttt{\textbackslash{}u00e2\textbackslash{}u00bd\textbackslash{}u00ba} \\
8 & \texttt{\textbackslash{}u00e2\textbackslash{}u0134\textbackslash{}u0142} & 9 & \texttt{\textbackslash{}u00e3\textbackslash{}u012c\textbackslash{}u00a5} & 10 & \texttt{\textbackslash{}u00e2\textbackslash{}u00bc\textbackslash{}u0143} & 11 & \texttt{\textbackslash{}u00e2\textbackslash{}u00b0\textbackslash{}u00a1} \\
12 & \texttt{\textbackslash{}u00e2\textbackslash{}u00b0\textbackslash{}u0142} & 13 & \texttt{\textbackslash{}u00e2\textbackslash{}u00b0\textbackslash{}u0141} & 14 & \texttt{\textbackslash{}u00e2\textbackslash{}u00b0\textbackslash{}u0133} & 15 & \texttt{\textbackslash{}u00e2\textbackslash{}u00b0\textbackslash{}u0132} \\
16 & \texttt{\textbackslash{}u00e2\textbackslash{}u00b0\textbackslash{}u0130} & 17 & \texttt{\textbackslash{}u00e2\textbackslash{}u00b0\textbackslash{}u012f} & 18 & \texttt{\textbackslash{}u00e2\textbackslash{}u00b0\textbackslash{}u0124} & 19 & \texttt{\textbackslash{}u00e2\textbackslash{}u0134\textbackslash{}u00a1} \\
20 & \texttt{\textbackslash{}u00e2\textbackslash{}u0134\textbackslash{}u0141} & 21 & \texttt{\textbackslash{}u00e2\textbackslash{}u0122\textbackslash{}u00b4} & 22 & \texttt{\textbackslash{}u00e2\textbackslash{}u0136\textbackslash{}u00b2} & 23 & \texttt{\textbackslash{}u00f0\textbackslash{}u0135\textbackslash{}u0131\textbackslash{}u00a7} \\
24 & \texttt{\textbackslash{}u00ef\textbackslash{}u00a8\textbackslash{}u00b7} & 25 & \texttt{\textbackslash{}u00e3\textbackslash{}u012a\textbackslash{}u00bc} & 26 & \texttt{\textbackslash{}u00e2\textbackslash{}u0140\textbackslash{}u00b6} & 27 & \texttt{\textbackslash{}u00e2\textbackslash{}u0138\textbackslash{}u00a4} \\
28 & \texttt{\textbackslash{}u00e2\textbackslash{}u0129\textbackslash{}u0140} & 29 & \texttt{\textbackslash{}u00e2\textbackslash{}u0128\textbackslash{}u00b7} & 30 & \texttt{\textbackslash{}u00e2\textbackslash{}u0128\textbackslash{}u00a4} & 31 & \texttt{\textbackslash{}u00e1\textbackslash{}u00a5\textbackslash{}u00a4} \\
32 & \texttt{\textbackslash{}u00e1\textbackslash{}u00a5\textbackslash{}u0136} & 33 & \texttt{\textbackslash{}u00e1\textbackslash{}u0127\textbackslash{}u00a3} & 34 & \texttt{\textbackslash{}u00e0\textbackslash{}u00ba\textbackslash{}u0124} & 35 & \texttt{\textbackslash{}u00ef\textbackslash{}u00b1\textbackslash{}u012c} \\
36 & \texttt{\textbackslash{}u00ea\textbackslash{}u00a6\textbackslash{}u0136} & 37 & \texttt{\textbackslash{}u00e3\textbackslash{}u012b\textbackslash{}u00ab} & 38 & \texttt{\textbackslash{}u00e3\textbackslash{}u0127\textbackslash{}u0138} & 39 & \texttt{\textbackslash{}u00e3\textbackslash{}u0126\textbackslash{}u00a7} \\
40 & \texttt{\textbackslash{}u00e3\textbackslash{}u0126\textbackslash{}u0135} & 41 & \texttt{\textbackslash{}u00e3\textbackslash{}u0126\textbackslash{}u012f} & 42 & \texttt{\textbackslash{}u00e2\textbackslash{}u0141\textbackslash{}u00b0} & 43 & \texttt{\textbackslash{}u00e2\textbackslash{}u013f\textbackslash{}u00ab} \\
44 & \texttt{\textbackslash{}u00e2\textbackslash{}u013f\textbackslash{}u00aa} & 45 & \texttt{\textbackslash{}u00e2\textbackslash{}u013d\textbackslash{}u0131} & 46 & \texttt{\textbackslash{}u00e2\textbackslash{}u013d\textbackslash{}u0129} & 47 & \texttt{\textbackslash{}u00e2\textbackslash{}u0137\textbackslash{}u012c} \\
48 & \texttt{\textbackslash{}u00e2\textbackslash{}u0136\textbackslash{}u00bd} & 49 & \texttt{\textbackslash{}u00e1\textbackslash{}u00b8\textbackslash{}u012c} & 50 & \texttt{\textbackslash{}u00e1\textbackslash{}u00a4\textbackslash{}u012c} & 51 & \texttt{\textbackslash{}u00e1\textbackslash{}u013d\textbackslash{}u0132} \\
52 & \texttt{\textbackslash{}u00e1\textbackslash{}u013d\textbackslash{}u0127} & 53 & \texttt{\textbackslash{}u00e1\textbackslash{}u013c\textbackslash{}u012e} & 54 & \texttt{\textbackslash{}u00e1\textbackslash{}u013b\textbackslash{}u00b3} & 55 & \texttt{\textbackslash{}u00e0\textbackslash{}u0142\textbackslash{}u012e} \\
56 & \texttt{\textbackslash{}u00c6\textbackslash{}u012a} & 57 & \texttt{\textbackslash{}u00f0\textbackslash{}u0141\textbackslash{}u0127\textbackslash{}u0135} & 58 & \texttt{\textbackslash{}u00f0\textbackslash{}u0141\textbackslash{}u0127\textbackslash{}u0127} & 59 & \texttt{\textbackslash{}u00f0\textbackslash{}u013f\textbackslash{}u013c\textbackslash{}u0131} \\
60 & \texttt{\textbackslash{}u00f0\textbackslash{}u013f\textbackslash{}u013c\textbackslash{}u0126} & 61 & \texttt{\textbackslash{}u00f0\textbackslash{}u013f\textbackslash{}u013b\textbackslash{}u00bf} & 62 & \texttt{\textbackslash{}u00f0\textbackslash{}u013f\textbackslash{}u013b\textbackslash{}u00bd} & 63 & \texttt{\textbackslash{}u00f0\textbackslash{}u013f\textbackslash{}u013b\textbackslash{}u00bc} \\
64 & \texttt{\textbackslash{}u00f0\textbackslash{}u013f\textbackslash{}u013b\textbackslash{}u00ba} & 65 & \texttt{\textbackslash{}u00f0\textbackslash{}u013f\textbackslash{}u013b\textbackslash{}u00b8} & 66 & \texttt{\textbackslash{}u00f0\textbackslash{}u013f\textbackslash{}u013b\textbackslash{}u00b0} & 67 & \texttt{\textbackslash{}u00f0\textbackslash{}u013f\textbackslash{}u013b\textbackslash{}u00ae} \\
68 & \texttt{\textbackslash{}u00f0\textbackslash{}u013f\textbackslash{}u013a\textbackslash{}u013c} & 69 & \texttt{\textbackslash{}u00f0\textbackslash{}u013f\textbackslash{}u013a\textbackslash{}u0132} & 70 & \texttt{\textbackslash{}u00f0\textbackslash{}u013f\textbackslash{}u013a\textbackslash{}u0131} & 71 & \texttt{\textbackslash{}u00f0\textbackslash{}u013f\textbackslash{}u0138\textbackslash{}u0138} \\
72 & \texttt{\textbackslash{}u00f0\textbackslash{}u013f\textbackslash{}u0137\textbackslash{}u00b1} & 73 & \texttt{\textbackslash{}u00f0\textbackslash{}u013f\textbackslash{}u0137\textbackslash{}u00a1} & 74 & \texttt{\textbackslash{}u00f0\textbackslash{}u013f\textbackslash{}u0137\textbackslash{}u012f} & 75 & \texttt{\textbackslash{}u00f0\textbackslash{}u013f\textbackslash{}u0136\textbackslash{}u0135} \\
76 & \texttt{\textbackslash{}u00f0\textbackslash{}u013f\textbackslash{}u0135\textbackslash{}u00be} & 77 & \texttt{\textbackslash{}u00f0\textbackslash{}u013f\textbackslash{}u0135\textbackslash{}u00b9} & 78 & \texttt{\textbackslash{}u00f0\textbackslash{}u013f\textbackslash{}u0135\textbackslash{}u00ac} & 79 & \texttt{\textbackslash{}u00f0\textbackslash{}u013f\textbackslash{}u0135\textbackslash{}u0137} \\
80 & \texttt{\textbackslash{}u00f0\textbackslash{}u013f\textbackslash{}u0133\textbackslash{}u00b3} & 81 & \texttt{\textbackslash{}u00f0\textbackslash{}u0138\textbackslash{}u00a5\textbackslash{}u00a8} & 82 & \texttt{\textbackslash{}u00f0\textbackslash{}u0138\textbackslash{}u00a5} & 83 & \texttt{\textbackslash{}u00f0\textbackslash{}u0132\textbackslash{}u00b1\textbackslash{}u0127} \\
84 & \texttt{\textbackslash{}u00f0\textbackslash{}u0132\textbackslash{}u0143\textbackslash{}u012c} & 85 & \texttt{\textbackslash{}u00ef\textbackslash{}u0143\textbackslash{}u00b2} & 86 & \texttt{\textbackslash{}u00ef\textbackslash{}u00a5\textbackslash{}u00b1} & 87 & \texttt{\textbackslash{}u00ef\textbackslash{}u00a5\textbackslash{}u0142} \\
88 & \texttt{\textbackslash{}u00ef\textbackslash{}u00a4\textbackslash{}u00a6} & 89 & \texttt{\textbackslash{}u00ed\textbackslash{}u0135\textbackslash{}u00bb} & 90 & \texttt{\textbackslash{}u00ed\textbackslash{}u0135\textbackslash{}u00b6} & 91 & \texttt{\textbackslash{}u00ed\textbackslash{}u0135\textbackslash{}u00ae} \\
92 & \texttt{\textbackslash{}u00ed\textbackslash{}u0135\textbackslash{}u00ac} & 93 & \texttt{\textbackslash{}u00ed\textbackslash{}u012d\textbackslash{}u012f} & 94 & \texttt{\textbackslash{}u00ec\textbackslash{}u00bc\textbackslash{}u0129} & 95 & \texttt{\textbackslash{}u00ec\textbackslash{}u0128\textbackslash{}u012c} \\
96 & \texttt{\textbackslash{}u00eb\textbackslash{}u00a1\textbackslash{}u00bc} & 97 & \texttt{\textbackslash{}u00ea\textbackslash{}u00b3\textbackslash{}u0124} & 98 & \texttt{\textbackslash{}u00ea\textbackslash{}u00b2\textbackslash{}u00b4} & 99 & \texttt{\textbackslash{}u00ea\textbackslash{}u00b2\textbackslash{}u013b} \\
100 & \texttt{\textbackslash{}u00e4\textbackslash{}u00b6\textbackslash{}u00b5} & 101 & \texttt{\textbackslash{}u00e3\textbackslash{}u012a\textbackslash{}u00aa} & 102 & \texttt{\textbackslash{}u00e2\textbackslash{}u00b2\textbackslash{}u00a2} & 103 & \texttt{\textbackslash{}u00e2\textbackslash{}u013c\textbackslash{}u00a3} \\
104 & \texttt{\textbackslash{}u00e2\textbackslash{}u013a\textbackslash{}u00b5} & 105 & \texttt{\textbackslash{}u00e2\textbackslash{}u0136\textbackslash{}u0140} & 106 & \texttt{\textbackslash{}u00e1\textbackslash{}u00b8\textbackslash{}u00bb} & 107 & \texttt{\textbackslash{}u00e1\textbackslash{}u00b8\textbackslash{}u0125} \\
108 & \texttt{\textbackslash{}u00e1\textbackslash{}u00a8\textbackslash{}u0123} & 109 & \texttt{\textbackslash{}u00e1\textbackslash{}u0142\textbackslash{}u0126} & 110 & \texttt{\textbackslash{}u00e1\textbackslash{}u0136\textbackslash{}u012c} & 111 & \texttt{\textbackslash{}u00e1\textbackslash{}u0136\textbackslash{}u0127} \\
112 & \texttt{\textbackslash{}u00e1\textbackslash{}u0134\textbackslash{}u012e} & 113 & \texttt{\textbackslash{}u00e1\textbackslash{}u0132\textbackslash{}u00a7} & 114 & \texttt{\textbackslash{}u00e1\textbackslash{}u012e\textbackslash{}u0136} & 115 & \texttt{\textbackslash{}u00e1\textbackslash{}u012e\textbackslash{}u0126} \\
116 & \texttt{\textbackslash{}u00e1\textbackslash{}u012d\textbackslash{}u00a9} & 117 & \texttt{\textbackslash{}u00e1\textbackslash{}u012c\textbackslash{}u0134} & 118 & \texttt{\textbackslash{}u00e1\textbackslash{}u012b\textbackslash{}u00a8} & 119 & \texttt{\textbackslash{}u00e1\textbackslash{}u0123\textbackslash{}u00bc} \\
120 & \texttt{\textbackslash{}u00e1\textbackslash{}u0122\textbackslash{}u0131} & 121 & \texttt{\textbackslash{}u00e0\textbackslash{}u00b2\textbackslash{}u0141} & 122 & \texttt{\textbackslash{}u00e0\textbackslash{}u00b0\textbackslash{}u00b5} & 123 & \texttt{\textbackslash{}u00e0\textbackslash{}u00b0\textbackslash{}u00b3} \\
124 & \texttt{\textbackslash{}u00e0\textbackslash{}u00ac\textbackslash{}u012b} & 125 & \texttt{\textbackslash{}u00e0\textbackslash{}u00a5\textbackslash{}u00b1} & 126 & \texttt{\textbackslash{}u00e0\textbackslash{}u00a4\textbackslash{}u0133} & 127 & \texttt{\textbackslash{}u00dd\textbackslash{}u00a5} \\
\addlinespace[4pt]  
\bottomrule
\end{tabular}
\end{table*}
% NOTE: Compile with XeLaTeX or LuaLaTeX to render UTF-8 token glyphs.
\begin{table*}[!htbp]
\centering
\footnotesize
\setlength{\tabcolsep}{1pt}
\setlength{\arrayrulewidth}{0.1pt}
\renewcommand{\arraystretch}{1.1}
\caption{\textbf{Action token vocabulary}: Mapping from discrete bin index (128 to 255) to the actual token string.}
\label{tab:action_vocab_2}
\begin{tabular}{r l r l r l r l}
\toprule
\textbf{Bin} & \textbf{| Action Token} & \textbf{Bin} & \textbf{| Action Token} & \textbf{Bin} & \textbf{| Action Token} & \textbf{Bin} & \textbf{| Action Token} \\
\midrule
128 & \texttt{\textbackslash{}u00dd\textbackslash{}u0135} & 129 & \texttt{\textbackslash{}u00d4\textbackslash{}u0133} & 130 & \texttt{\textbackslash{}u00d4\textbackslash{}u012a} & 131 & \texttt{\textbackslash{}u00ca\textbackslash{}u00b6} \\
132 & \texttt{\textbackslash{}u00c8\textbackslash{}u00b2} & 133 & \texttt{\textbackslash{}u00f0\textbackslash{}u0141\textbackslash{}u0131\textbackslash{}u0129} & 134 & \texttt{\textbackslash{}u00f0\textbackslash{}u0141\textbackslash{}u0127\textbackslash{}u00a2} & 135 & \texttt{\textbackslash{}u00f0\textbackslash{}u013f\textbackslash{}u013c\textbackslash{}u0123} \\
136 & \texttt{\textbackslash{}u00f0\textbackslash{}u013f\textbackslash{}u013b\textbackslash{}u013e} & 137 & \texttt{\textbackslash{}u00f0\textbackslash{}u013f\textbackslash{}u0135\textbackslash{}u00b0} & 138 & \texttt{\textbackslash{}u00f0\textbackslash{}u013f\textbackslash{}u0135\textbackslash{}u0140} & 139 & \texttt{\textbackslash{}u00f0\textbackslash{}u0132\textbackslash{}u00b0\textbackslash{}u00bc} \\
140 & \texttt{\textbackslash{}u00f0\textbackslash{}u0132\textbackslash{}u0143\textbackslash{}u0135} & 141 & \texttt{\textbackslash{}u00f0\textbackslash{}u0132\textbackslash{}u00a4\textbackslash{}u0136} & 142 & \texttt{\textbackslash{}u00ef\textbackslash{}u00a8\textbackslash{}u0124} & 143 & \texttt{\textbackslash{}u00ef\textbackslash{}u00a7\textbackslash{}u00a9} \\
144 & \texttt{\textbackslash{}u00ef\textbackslash{}u00a6\textbackslash{}u0125} & 145 & \texttt{\textbackslash{}u00ef\textbackslash{}u00a4\textbackslash{}u0128} & 146 & \texttt{\textbackslash{}u00ef\textbackslash{}u00a4\textbackslash{}u0127} & 147 & \texttt{\textbackslash{}u00ed\textbackslash{}u013d\textbackslash{}u013e} \\
148 & \texttt{\textbackslash{}u00ed\textbackslash{}u0137\textbackslash{}u00b1} & 149 & \texttt{\textbackslash{}u00ed\textbackslash{}u0135\textbackslash{}u0143} & 150 & \texttt{\textbackslash{}u00ed\textbackslash{}u0135\textbackslash{}u0138} & 151 & \texttt{\textbackslash{}u00ed\textbackslash{}u0125\textbackslash{}u013b} \\
152 & \texttt{\textbackslash{}u00ed\textbackslash{}u0123\textbackslash{}u00bb} & 153 & \texttt{\textbackslash{}u00ec\textbackslash{}u00bb\textbackslash{}u0123} & 154 & \texttt{\textbackslash{}u00ec\textbackslash{}u00b3\textbackslash{}u0127} & 155 & \texttt{\textbackslash{}u00ec\textbackslash{}u013e\textbackslash{}u00be} \\
156 & \texttt{\textbackslash{}u00ec\textbackslash{}u013d\textbackslash{}u00a2} & 157 & \texttt{\textbackslash{}u00eb\textbackslash{}u00b1\textbackslash{}u0132} & 158 & \texttt{\textbackslash{}u00eb\textbackslash{}u00b1\textbackslash{}u012d} & 159 & \texttt{\textbackslash{}u00eb\textbackslash{}u00a7\textbackslash{}u0142} \\
160 & \texttt{\textbackslash{}u00eb\textbackslash{}u00a4\textbackslash{}u0124} & 161 & \texttt{\textbackslash{}u00eb\textbackslash{}u0138\textbackslash{}u00b0} & 162 & \texttt{\textbackslash{}u00e2\textbackslash{}u00a4\textbackslash{}u00a6} & 163 & \texttt{\textbackslash{}u00e2\textbackslash{}u00a1\textbackslash{}u00a2} \\
164 & \texttt{\textbackslash{}u00e2\textbackslash{}u013c\textbackslash{}u0139} & 165 & \texttt{\textbackslash{}u00e2\textbackslash{}u013c\textbackslash{}u0124} & 166 & \texttt{\textbackslash{}u00e2\textbackslash{}u013b\textbackslash{}u013b} & 167 & \texttt{\textbackslash{}u00e1\textbackslash{}u00bf\textbackslash{}u013c} \\
168 & \texttt{\textbackslash{}u00e1\textbackslash{}u00bf\textbackslash{}u0132} & 169 & \texttt{\textbackslash{}u00e1\textbackslash{}u00be\textbackslash{}u0136} & 170 & \texttt{\textbackslash{}u00e1\textbackslash{}u00b6\textbackslash{}u0131} & 171 & \texttt{\textbackslash{}u00e1\textbackslash{}u00a9\textbackslash{}u012d} \\
172 & \texttt{\textbackslash{}u00e1\textbackslash{}u00a8\textbackslash{}u00b8} & 173 & \texttt{\textbackslash{}u00e1\textbackslash{}u0142\textbackslash{}u00ac} & 174 & \texttt{\textbackslash{}u00e1\textbackslash{}u0142\textbackslash{}u0124} & 175 & \texttt{\textbackslash{}u00e1\textbackslash{}u0136\textbackslash{}u0143} \\
176 & \texttt{\textbackslash{}u00e1\textbackslash{}u012e\textbackslash{}u00bd} & 177 & \texttt{\textbackslash{}u00e1\textbackslash{}u012e\textbackslash{}u0125} & 178 & \texttt{\textbackslash{}u00e1\textbackslash{}u012b\textbackslash{}u0132} & 179 & \texttt{\textbackslash{}u00e1\textbackslash{}u012a\textbackslash{}u00be} \\
180 & \texttt{\textbackslash{}u00e1\textbackslash{}u012a\textbackslash{}u00a8} & 181 & \texttt{\textbackslash{}u00e1\textbackslash{}u012a\textbackslash{}u012c} & 182 & \texttt{\textbackslash{}u00e1\textbackslash{}u0128\textbackslash{}u00ba} & 183 & \texttt{\textbackslash{}u00e0\textbackslash{}u00bd\textbackslash{}u0127} \\
184 & \texttt{\textbackslash{}u00e0\textbackslash{}u00b4\textbackslash{}u00b4} & 185 & \texttt{\textbackslash{}u00d5\textbackslash{}u0125} & 186 & \texttt{\textbackslash{}u00ca\textbackslash{}u0135} & 187 & \texttt{\textbackslash{}u00c9\textbackslash{}u013a} \\
188 & \texttt{\textbackslash{}u00f0\textbackslash{}u0141\textbackslash{}u0137\textbackslash{}u012d} & 189 & \texttt{\textbackslash{}u00f0\textbackslash{}u0141\textbackslash{}u0128\textbackslash{}u0134} & 190 & \texttt{\textbackslash{}u00f0\textbackslash{}u0141\textbackslash{}u0127\textbackslash{}u00b1} & 191 & \texttt{\textbackslash{}u00ef\textbackslash{}u00ae\textbackslash{}u0131} \\
192 & \texttt{\textbackslash{}u00ed\textbackslash{}u0137\textbackslash{}u00ae} & 193 & \texttt{\textbackslash{}u00ed\textbackslash{}u012c\textbackslash{}u0143} & 194 & \texttt{\textbackslash{}u00ec\textbackslash{}u00a5\textbackslash{}u012b} & 195 & \texttt{\textbackslash{}u00ec\textbackslash{}u0142\textbackslash{}u00b0} \\
196 & \texttt{\textbackslash{}u00ec\textbackslash{}u0141\textbackslash{}u013b} & 197 & \texttt{\textbackslash{}u00ec\textbackslash{}u013f\textbackslash{}u00bf} & 198 & \texttt{\textbackslash{}u00ec\textbackslash{}u013f\textbackslash{}u00a9} & 199 & \texttt{\textbackslash{}u00ec\textbackslash{}u0139\textbackslash{}u00a4} \\
200 & \texttt{\textbackslash{}u00ec\textbackslash{}u0131\textbackslash{}u00b1} & 201 & \texttt{\textbackslash{}u00ec\textbackslash{}u012d\textbackslash{}u00b2} & 202 & \texttt{\textbackslash{}u00ec\textbackslash{}u012b\textbackslash{}u00a1} & 203 & \texttt{\textbackslash{}u00ec\textbackslash{}u0126\textbackslash{}u0132} \\
204 & \texttt{\textbackslash{}u00eb\textbackslash{}u00bc\textbackslash{}u013f} & 205 & \texttt{\textbackslash{}u00eb\textbackslash{}u00bb\textbackslash{}u0127} & 206 & \texttt{\textbackslash{}u00eb\textbackslash{}u00af\textbackslash{}u0133} & 207 & \texttt{\textbackslash{}u00eb\textbackslash{}u00a1\textbackslash{}u0133} \\
208 & \texttt{\textbackslash{}u00eb\textbackslash{}u0139\textbackslash{}u012f} & 209 & \texttt{\textbackslash{}u00eb\textbackslash{}u0136\textbackslash{}u012b} & 210 & \texttt{\textbackslash{}u00ea\textbackslash{}u00b8\textbackslash{}u0133} & 211 & \texttt{\textbackslash{}u00ea\textbackslash{}u013b\textbackslash{}u012d} \\
212 & \texttt{\textbackslash{}u00e3\textbackslash{}u00b3\textbackslash{}u00ac} & 213 & \texttt{\textbackslash{}u00e2\textbackslash{}u013d\textbackslash{}u00a4} & 214 & \texttt{\textbackslash{}u00e2\textbackslash{}u013c\textbackslash{}u00a7} & 215 & \texttt{\textbackslash{}u00e2\textbackslash{}u0126\textbackslash{}u00ac} \\
216 & \texttt{\textbackslash{}u00e1\textbackslash{}u00bd\textbackslash{}u013f} & 217 & \texttt{\textbackslash{}u00e1\textbackslash{}u00bc\textbackslash{}u00ae} & 218 & \texttt{\textbackslash{}u00e1\textbackslash{}u00ba\textbackslash{}u0122} & 219 & \texttt{\textbackslash{}u00e1\textbackslash{}u00b8\textbackslash{}u00b0} \\
220 & \texttt{\textbackslash{}u00e1\textbackslash{}u00a1\textbackslash{}u012e} & 221 & \texttt{\textbackslash{}u00da\textbackslash{}u0130} & 222 & \texttt{\textbackslash{}u00d1\textbackslash{}u00a8} & 223 & \texttt{\textbackslash{}u00f0\textbackslash{}u0141\textbackslash{}u0139\textbackslash{}u0123} \\
224 & \texttt{\textbackslash{}u00f0\textbackslash{}u0141\textbackslash{}u0138\textbackslash{}u00b6} & 225 & \texttt{\textbackslash{}u00f0\textbackslash{}u0141\textbackslash{}u0138\textbackslash{}u0133} & 226 & \texttt{\textbackslash{}u00f0\textbackslash{}u0141\textbackslash{}u0138\textbackslash{}u0129} & 227 & \texttt{\textbackslash{}u00f0\textbackslash{}u0141\textbackslash{}u0137\textbackslash{}u00b3} \\
228 & \texttt{\textbackslash{}u00f0\textbackslash{}u0141\textbackslash{}u0137\textbackslash{}u00a2} & 229 & \texttt{\textbackslash{}u00f0\textbackslash{}u0141\textbackslash{}u0137\textbackslash{}u0142} & 230 & \texttt{\textbackslash{}u00f0\textbackslash{}u0141\textbackslash{}u0137\textbackslash{}u0140} & 231 & \texttt{\textbackslash{}u00f0\textbackslash{}u0141\textbackslash{}u0137\textbackslash{}u013f} \\
232 & \texttt{\textbackslash{}u00f0\textbackslash{}u0141\textbackslash{}u0137\textbackslash{}u013e} & 233 & \texttt{\textbackslash{}u00f0\textbackslash{}u0141\textbackslash{}u0137\textbackslash{}u013c} & 234 & \texttt{\textbackslash{}u00f0\textbackslash{}u0141\textbackslash{}u0137\textbackslash{}u0138} & 235 & \texttt{\textbackslash{}u00f0\textbackslash{}u0141\textbackslash{}u0136\textbackslash{}u00a9} \\
236 & \texttt{\textbackslash{}u00f0\textbackslash{}u0141\textbackslash{}u0136\textbackslash{}u00a4} & 237 & \texttt{\textbackslash{}u00f0\textbackslash{}u0141\textbackslash{}u0136\textbackslash{}u00a2} & 238 & \texttt{\textbackslash{}u00f0\textbackslash{}u0141\textbackslash{}u0136\textbackslash{}u0135} & 239 & \texttt{\textbackslash{}u00f0\textbackslash{}u0141\textbackslash{}u0136\textbackslash{}u0129} \\
240 & \texttt{\textbackslash{}u00f0\textbackslash{}u0141\textbackslash{}u0136\textbackslash{}u0125} & 241 & \texttt{\textbackslash{}u00f0\textbackslash{}u0141\textbackslash{}u0136\textbackslash{}u0124} & 242 & \texttt{\textbackslash{}u00f0\textbackslash{}u0141\textbackslash{}u0136\textbackslash{}u0122} & 243 & \texttt{\textbackslash{}u00f0\textbackslash{}u0141\textbackslash{}u0135\textbackslash{}u00bc} \\
244 & \texttt{\textbackslash{}u00f0\textbackslash{}u0141\textbackslash{}u0135\textbackslash{}u00aa} & 245 & \texttt{\textbackslash{}u00f0\textbackslash{}u0141\textbackslash{}u0135\textbackslash{}u0141} & 246 & \texttt{\textbackslash{}u00f0\textbackslash{}u0141\textbackslash{}u0134\textbackslash{}u00ba} & 247 & \texttt{\textbackslash{}u00f0\textbackslash{}u0141\textbackslash{}u0134\textbackslash{}u00b9} \\
248 & \texttt{\textbackslash{}u00f0\textbackslash{}u0141\textbackslash{}u0133\textbackslash{}u013f} & 249 & \texttt{\textbackslash{}u00f0\textbackslash{}u0141\textbackslash{}u0132\textbackslash{}u0122} & 250 & \texttt{\textbackslash{}u00f0\textbackslash{}u0141\textbackslash{}u0131\textbackslash{}u00af} & 251 & \texttt{\textbackslash{}u00f0\textbackslash{}u0141\textbackslash{}u0131\textbackslash{}u00a9} \\
252 & \texttt{\textbackslash{}u00f0\textbackslash{}u0141\textbackslash{}u0131\textbackslash{}u0134} & 253 & \texttt{\textbackslash{}u00f0\textbackslash{}u0141\textbackslash{}u0131\textbackslash{}u0131} & 254 & \texttt{\textbackslash{}u00f0\textbackslash{}u0141\textbackslash{}u0130\textbackslash{}u00bf} & 255 & \texttt{\textbackslash{}u00f0\textbackslash{}u0141\textbackslash{}u0130\textbackslash{}u0133} \\
\addlinespace[4pt]  
\bottomrule
\end{tabular}
\end{table*}
\clearpage

\section{Evaluation Details}
\label{supp:eval}

\begin{figure}[t]
  \centering
  \includegraphics[width=\linewidth]{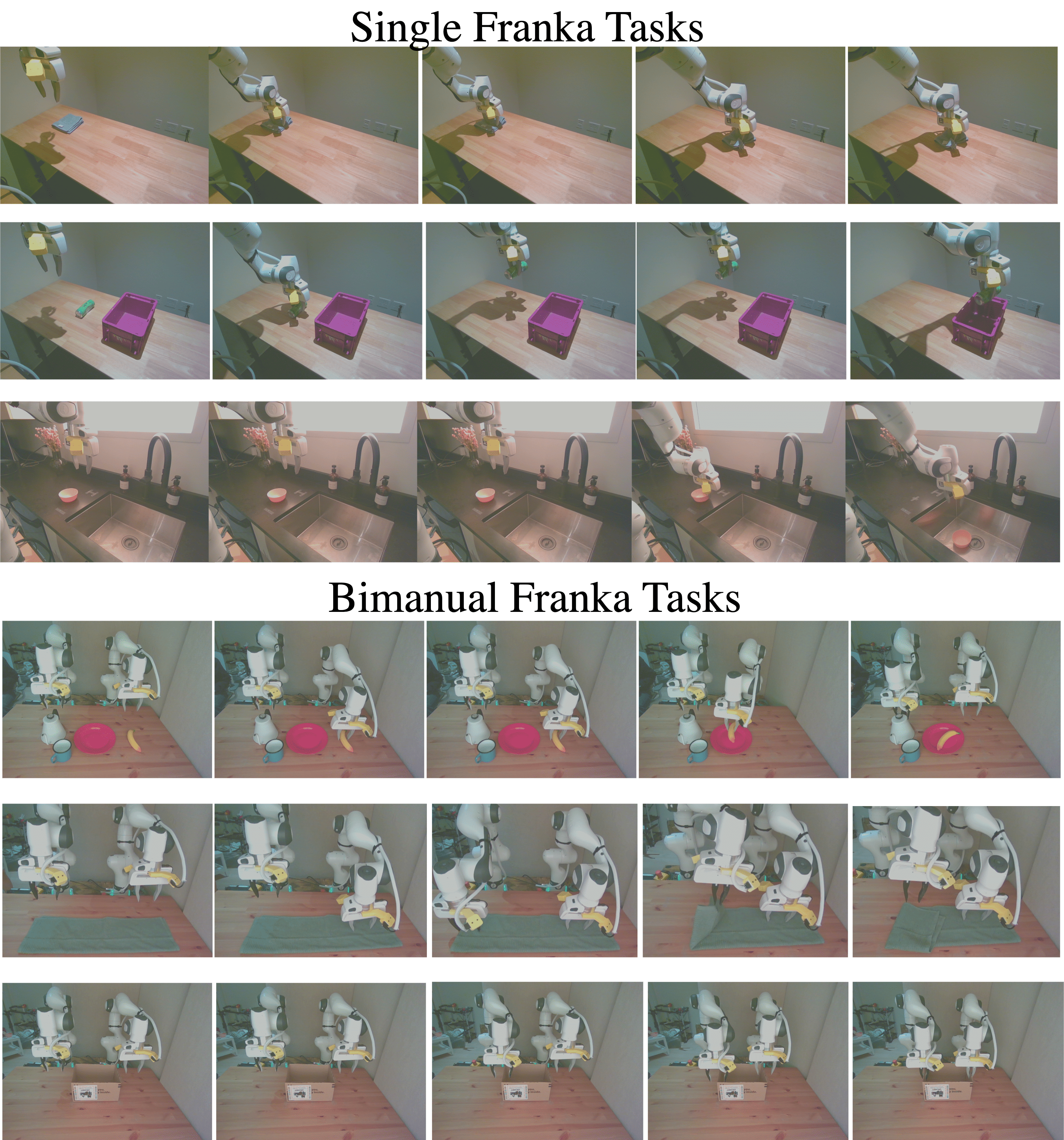}
  \caption{\textbf{Examples of Single-arm and Bimanual Tasks.} We list the observation breakdown to show how the robot performs each task.}
  \label{fig:singlearm_bimanual_example}
\end{figure}

\begin{figure}[t]
  \centering
  \includegraphics[width=\linewidth]{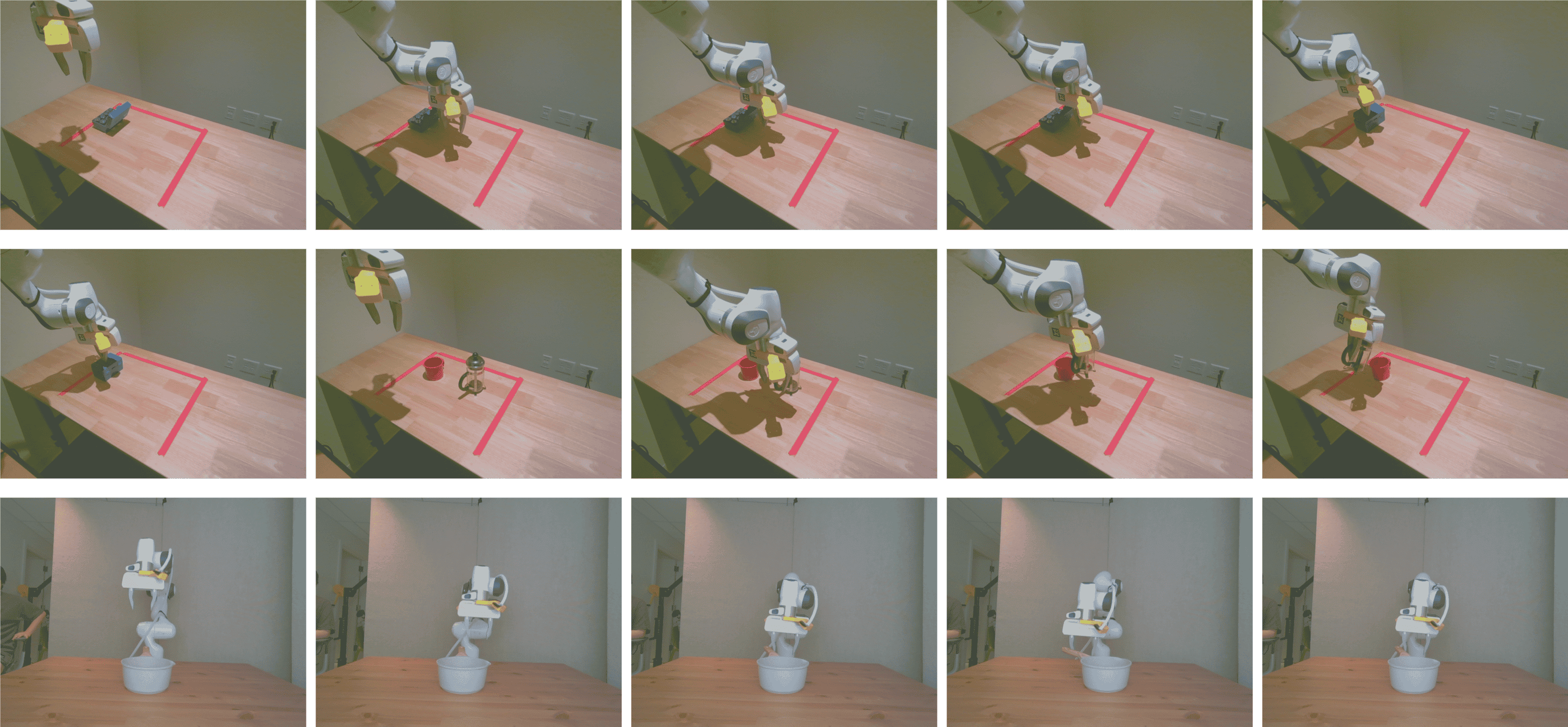}
  \caption{\textbf{Examples of \molmoactdata ablation experiments.}}
  \label{fig:ablation_example}
\end{figure}

\subsection{Evaluation on \simpler (Google Robot)}
\label{supp:eval:simpler}
We evaluate on \simpler \cite{li2024evaluating} simulation to test \molmoact's out-of-the-box performance on the Google Robot setup. The simulation evaluation consists of a Google Robot arm, front-view camera image (640 x 480 px, resized to 320 x 240 px for our case), task language instructions, and delta end-effector pose actions. \simpler evaluation consists of two components -- Visual Matching and Variant Aggregation.

\begin{figure}[t]
  \centering
  \includegraphics[width=\linewidth]{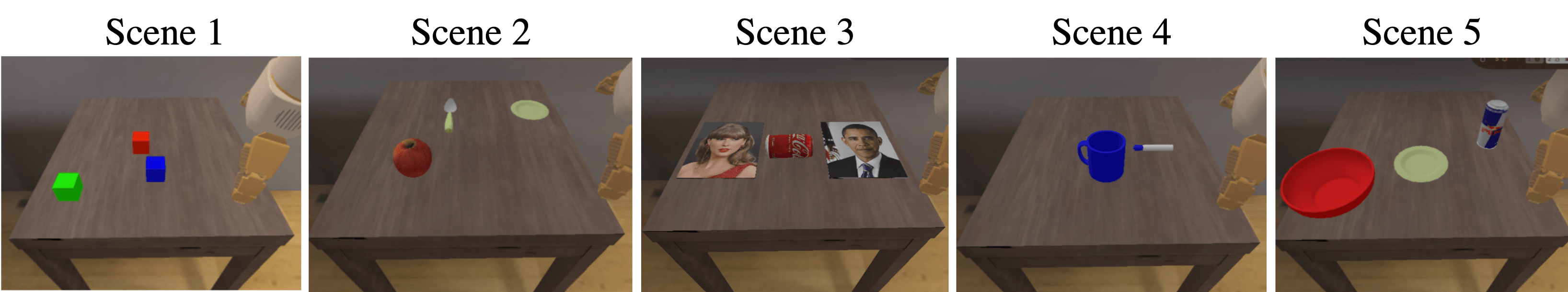}
  \caption{\textbf{Language instruction following.} These are the customized scenes curated for open-ended prompting by users.}
  \label{fig:line_generalize_example}
\end{figure}
\subsection{Evaluation on \libero}
\label{supp:eval:libero}
We evaluate on the \libero simulation benchmark \citep{liu2023libero}, which consists of a Franka Emika Panda arm in simulation with demonstrations containing front and wrist view camera images (256 x 256px), tasks language instructions, and delta end-effector pose actions. We follow prior works \citep{kim2024openvla} and evaluate on the four task suites -- \libero-Spatial, \libero-Object, \libero-Goal, and \libero-Long -- each with 500 expert demonstration across 10 tasks. Following \citep{kim2024openvla}, we trained on a modified dataset which filtered out no-ops actions and unsuccessful demonstrations. Moreover, we set action chunk size to $K = 8$ for evaluation on each task suites and execute full chunks before replanning. We report details of our post-training hyperparameters for \libero in \autoref{tab:hyperparams_libero}.

\begin{table}[!htbp]
\centering
\footnotesize
\setlength\tabcolsep{4pt}
\renewcommand\arraystretch{1.1}

\begin{tabular}{l*{4}{>{\centering\arraybackslash}p{2.3cm}}}
\toprule
\multicolumn{1}{c}{} &
\multicolumn{4}{c}{\cellcolor{olive!5}\textbf{LIBERO Task Suite}} \\[-0.2em]
\cmidrule(lr){2-5}
\textbf{Parameter} & Spatial & Object & Goal & Long \\
\midrule
% Learning Rate      & \multicolumn{4}{c}{5e-4} \\
Steps              & 50K & 50K & 40K & 80K \\
Global Batch Size  & \multicolumn{4}{c}{128} \\
GPUs (H100s)       & \multicolumn{4}{c}{64} \\
Time (Hours)       & 23 & 23 & 18 & 36 \\
GPU Hours          & 1472 & 1472 & 1152 & 2304 \\
Input Images            & \multicolumn{4}{c}{1 Third-person + 1 Wrist-mounted} \\
Image Size              & \multicolumn{4}{c}{256$\times$256 px} \\
DoF                     & \multicolumn{4}{c}{7 (3 Translations + 3 Rotations + 1 Gripper State)} \\
Observation History     & \multicolumn{4}{c}{No (Single-step Inputs)} \\
Use Proprioception & \multicolumn{4}{c}{No} \\
Action Chunk Size       & \multicolumn{4}{c}{8 Steps (Predict 8; Execute All 8 Open-loop)} \\
% LoRA Rank               & \multicolumn{4}{c}{32} \\
% LoRA Alpha              & \multicolumn{4}{c}{16} \\
% LoRA Dropout            & \multicolumn{4}{c}{0} \\
% LoRA Bias               & \multicolumn{4}{c}{None} \\
\# Trainable Params     & \multicolumn{4}{c}{97 M LoRA adapter} \\
Image Augmentations     &
\multicolumn{4}{c}{%
  \parbox[t]{10.2cm}{\raggedright
\texttt{import torchvision.transforms as T}\\
\texttt{transform = T.Compose([}\\
\texttt{\hspace*{1em}T.RandomResizedCrop(size=(height, width), scale=(0.9, 0.9), ratio=(width/height, width/height)),}\\
\texttt{\hspace*{1em}T.Resize((height, width)),}\\
\texttt{\hspace*{1em}T.ColorJitter(}\\
\texttt{\hspace*{2em}brightness=0.2,}\\
\texttt{\hspace*{2em}contrast=(0.8, 1.2),}\\
\texttt{\hspace*{2em}saturation=(0.8, 1.2),}\\
\texttt{\hspace*{2em}hue=0.05}\\
\texttt{\hspace*{1em}),}\\
\texttt{])}
}} \\
\addlinespace[4pt]  
\bottomrule
\end{tabular}
\caption{\textbf{\molmoact's Post-training Hyperparameters for \libero.} We specify the hyperparameters for \molmoact post-training. Note that we conduct all our post-training experiments on \molmoactd, with a fixed learning rate of 5e-4, LoRA rank of 32, LoRA alpha of 16, LoRA dropout of 0, and no LoRA bias. Note that for \libero-Goal, we train the model with 50K steps, but it reaches better performance at 40K steps.}
\label{tab:hyperparams_libero}
\end{table}

\subsection{Evaluation on Real-world Post-training}
\label{supp:eval:indist}
To evaluate \molmoact’s efficiency in fine-tuning, we curated six tasks: three for a single-arm Franka setup—\texttt{put bowl in sink}, \texttt{wipe table}, and \texttt{table bussing}—and three for a bimanual Franka setup—\texttt{set table}, \texttt{lift tray}, and \texttt{fold towel}. We benchmarked against OpenVLA and \pizerofast by training each model until convergence. In the single-arm setup, the Franka was mounted on a movable platform to allow relocation across different positions, whereas the bimanual setup was fixed to a tabletop configuration. For sensing, we employed an Intel RealSense D405 for the wrist-mounted camera and an Intel RealSense D435 for the front-facing view. In each efficiency fine-tuning task evaluation, we pre-marked the locations of all task objects in the scene to ensure that the evaluation conditions matched the distribution of the demonstrations used for fine-tuning. We defined the task, its description, the corresponding language instruction, and the task progression metric ratings. Refer to the complete results in Table \ref{tab:eval_per_trial_fold_towel} to \ref{tab:eval_per_trial_clean_table}.

\begin{enumerate}[leftmargin=0.5cm, label=\textbf{\arabic*.}]
    \taskitem
        {\texttt{put\_bowl\_in\_sink}}
        {The robot picks up the orange bowl next to the sink and place it all the way into the sink.}
        {\emph{Put the bowl into the sink.}}
        {grasp bowl (0.25), move into the sink (0.4), open gripper (0.7), drop bowl at target location (1).}

    \taskitem
        {\texttt{wipe\_table}}
        {The robot grasp onto the table cloth, and move across the surface in one direction.}
        {\emph{Wipe the table.}}
        {Grasp the towel (0.25), Move in the right direction (0.5), Complete the wipe (1).}

    \taskitem
        {\texttt{table\_bussing}}
        {The robot grasp onto the green tea can and place it into the purple bin.}
        {\emph{Clean the trash into the bin}}
        {Grasp onto the can (0.25), Lift up the can (0.5), Move to above the bin (0.75), Drop the can into the bin (1).}

    \taskitem
        {\texttt{set\_table}}
        {The right arm grasp onto the banana and place it onto the plate, and the left arm grasp onto the teapot to pour.}
        {\emph{Set the table}}
        {Put banana on plate (0.25), Grasp onto the teapot (0.75), Pour the tea (1).}

    \taskitem
        {\texttt{lift\_tray}}
        {The left and right arm approaches the box and grasp onto it, and lift up the box together.}
        {\emph{Lift up the box}}
        {Left arm grasp onto the tray (0.3), Right arm grasp onto the tray (0.6), Both arms lift up the tray (1).}

    \taskitem
        {\texttt{fold\_towel}}
        {The right arm press down on the centre of the towel, while the left arm grasp onto the towel to fold.}
        {\emph{Fold the towel}}
        {Grasp onto the towel (0.25), Put the towel over the right location for folding (0.75), Drop the towel so that it is folded (1).}
\end{enumerate}

We report details of \molmoact's post-training hyperparameters for this evaluation in \autoref{tab:hyperparams_real_indist_singlearm} (single-arm) and \autoref{tab:hyperparams_real_indist_bimanual} (bimanual). For all other baseline models, we follow their official model and training implementation and use their default configurations. We also make sure that they are all fully converged. Image examples of each task are shown in \autoref{fig:singlearm_bimanual_example}.

\subsection{Evaluation on Generalization in Real-world}
\label{supp:eval:generalize}
We collected a multi-task set that contains the full permutation of the scene:
\texttt{put\_green\_can\_in\_yellow\_plate} (\emph{put the green can into the yellow plate}), 
\texttt{put\_green\_can\_in\_blue\_plate} (\emph{put the green can into the blue plate}), 
\texttt{put\_red\_cup\_in\_yellow\_plate} (\emph{put the red cup into the yellow plate}), and 
\texttt{put\_red\_cup\_in\_blue\_plate} (\emph{put the red cup into the blue plate}), and 
\texttt{put\_banana\_in\_yellow\_plate} (\emph{put the banana into the yellow plate}). 
\texttt{put\_banana\_in\_blue\_plate} (\emph{put the banana into the blue plate}). 
And all models are trained on all tasks under a multi-task setting. 

We evaluated generalization across four perturbations and one in-distribution setting on three tasks drawn from the previous multi-task set: 
\texttt{put\_green\_can\_in\_yellow\_plate} (\emph{put the green can into the yellow plate}), 
\texttt{put\_red\_cup\_in\_yellow\_plate} (\emph{put the red cup into the yellow plate}), and 
\texttt{put\_banana\_in\_blue\_plate} (\emph{put the banana into the blue plate}). 
The perturbations were: 
(1) \textbf{Language variation} – modified instructions to \emph{put the green tea into the yellow plate}, \emph{put the fruit into the blue plate}, and \emph{put the red cylinder into the yellow plate}; 
(2) \textbf{Spatial variation} – altered the positions of objects in each task; 
(3) \textbf{Distractors} – added unrelated distractor objects to the scene; and 
(4) \textbf{Novel objects} – replaced the green can with a sponge, the red cup with a coke can, and the banana with a bowl.

\begin{itemize}
    \taskitem
        {\texttt{put\_<object>\_in\_(yellow/blue)\_plate}}
        {The robot first pickup the <object>, then put it into the yellow/blue plate.}
        {\emph{Put the <object> into the yellow/blue plate}}
        {Move towards the correct <object> (0.25). Pick up the correct <object> (0.5). Move towards the correct plate (0.75). Put the correct <object> into the correct plate (1).}
\end{itemize}

We report details of \molmoact's post-training hyperparameters for this evaluation in \autoref{tab:hyperparams_real_outdist}. For all other baseline models, we follow their official model and training implementation and use their default configurations. We also make sure that they are all fully converged. The full details of this evaluation are listed in \autoref{tab:eval_generalization_detail}.

\subsection{Evaluation on the Effect of \molmoactdata for \molmoact Mid-training}
\label{supp:eval:ablation}
To evaluate the effectiveness of mid-training with the \molmoactdata, we curated three real-world tasks: \texttt{close\_lid}, \texttt{rotate\_pot}, and \texttt{pour\_tea}. For each task, we collected 50 demonstrations and pre-marked object locations to ensure repeatability in evaluating \molmoact, \molmoact without the \molmoactdata, OpenVLA, and \pizerofast. We conducted 10 evaluation trials per task for each model. Refer to the complete results in Table \ref{tab:eval_per_trial_molmoactdata}

\begin{enumerate}[leftmargin=0.5cm, label=\textbf{\arabic*.}]
    \taskitem
        {\texttt{close\_lid}}
        {The robot goes to the back of the lid, closes its gripper and push the lid to close.}
        {\emph{Close the lid}}
        {Move the lid towards the closing direction (0.5). Close the lid (1).}

    \taskitem
        {\texttt{rotate\_pot}}
        {The robot goes to a target position to the handle, and rotate it by 90 degree.}
        {\emph{Rotate the pot}}
        {Go target position of pot handle (0.3). Rotate the pot by 45 degree (0.6). Close the 90 degree rotation (1).}

    \taskitem
        {\texttt{pour\_tea}}
        {The robot grasp onto the teapot handle, and lift it up to above the cup to pour.}
        {\emph{Pour tea into cup}}
        {Grasp onto the teapot (0.5). Move the teapot on top of cup (0.8). Pour tea into cup (1).}

\end{enumerate}

We report details of \molmoact's post-training hyperparameters for this evaluation in \autoref{tab:hyperparams_real_ablation}. For all other baseline models, we follow their official model and training implementation and use their default configurations. We also make sure that they are all fully converged. Image examples of each task are shown in \autoref{fig:ablation_example}.

\subsection{Evaluation of \molmoact on Instruction-following}
\label{supp:eval:follow}
For the evaluation of language-instruction following, we curated five customized scenes using \simpler \citep{li2024evaluating} and asked participants to provide open-ended prompts for each scene. After filtering, we obtained 29 prompts in total, which were executed by \molmoact, OpenVLA, and SpatialVLA for 200 steps to generate robot rollouts. These rollouts were then rated by 100 participants in an arena-style interface. Images of different scenes are shown in \autoref{fig:line_generalize_example}, and language prompts are shown in \autoref{tab:line_generalize_prompt}.

\begin{table}[h]
\centering
\renewcommand{\arraystretch}{1.2}
\begin{tabular}{p{0.15\linewidth} p{0.8\linewidth}}
\hline
\textbf{Scene} & \textbf{Prompts} \\
\hline
Scene 1 & 
\begin{itemize}[leftmargin=*]
    \item Pick up the green cube
    \item Pick up the red cube
    \item Pick up the blue cube
    \item Put the green cube on the blue cube
    \item Put the green cube on the red cube
    \item Put the blue cube onto the red cube
    \item Put the blue cube onto the green cube
    \item Put the red cube onto the green cube
    \item Put the red cube onto the blue cube
    \item Put the green cube onto the blue cube and then the red cube onto the green cube
    \item Move the blue cube next to the green cube
\end{itemize} \\
\hline
Scene 2 & 
\begin{itemize}[leftmargin=*]
    \item Pick up the apple
    \item Pick up the spoon
    \item Put the apple onto the plate
    \item Put the spoon onto the plate
    \item Put the spoon next to the plate
    \item Move the spoon to the right of the apple
    \item Put the apple onto the plate and move the spoon nearer to the plate
    \item Put the blue cube onto the green cube
    \item Put the red cube onto the green cube
    \item Put the red cube onto the blue cube
    \item Put the green cube onto the blue cube and then the red cube onto the green cube
\end{itemize} \\
\hline
Scene 3 & 
\begin{itemize}[leftmargin=*]
    \item Put the coke can onto the former president's image
    \item Put the coke can on the image of Obama
    \item Move the coke can to the image of Taylor Swift
\end{itemize} \\
\hline
Scene 4 & 
\begin{itemize}[leftmargin=*]
    \item Pick up the cup
    \item Pick up the marker
    \item Put the marker into the mug
    \item Pick up the mug by the handle
\end{itemize} \\
\hline
Scene 5 & 
\begin{itemize}[leftmargin=*]
    \item Pick up the bowl
    \item Pick up the Red Bull
    \item Put the Red Bull onto the plate
    \item Put the Red Bull in the red bowl
\end{itemize} \\
\hline
\end{tabular}
\caption{Open-ended prompts provided by users grouped by scene for language instruction following.}
\label{tab:line_generalize_prompt}
\end{table}

\subsection{Evaluation of \molmoact on Action Steerability}

\label{supp:eval:steer}

We curated the task \texttt{pick\_up\_bowl}, featuring one dirty and one clean bowl. As shown in Figure \ref{fig:fig9_steerability}, we built a web interface that enables users to modify the language instruction or sketch five points on the image for visual trace steering at test time, which are then passed to the model to generate actions. We evaluate task progression for this task based on: correct direction of the target bowel (0.5), grasp onto the correct bowl (0.85), lift up the bowl (1).

Unlike the usual straightforward way of collecting tele-operated real-world demonstrations, where we control the robot to directly complete the task, we collected half of the number of demonstrations in the regular way and the other half exploring alternative paths towards the same target conditioned on the language. Thus, in total, we have 50 demonstrations picking up the dirty bowl, 50 demonstrations picking up the clean bowl, 50 demonstrations picking up the dirty bowl while exploring other paths, and 50 demonstrations picking up the clean bowl while exploring other paths. We believe that this helps the model to learn more about how visual traces correlate with physical actions.

During test time, we collected open-ended instructions from 10 participants to steer the robot through language. We restrict the variation of the open-ended instructions only to verbs, nouns, or adjectives. The collected and used instructions are shown in \autoref{tab:open_instruc}.

We report details of \molmoact's post-training hyperparameters for this evaluation in \autoref{tab:hyperparams_real_steer}. For all other baseline models, we follow their official model and training implementation and use their default configurations. We also make sure that they are all fully converged. All results are reported Table \ref{tab:eval_bowl_tasks}.

\begin{table}[!htbp]
\centering
\footnotesize
\setlength\tabcolsep{6pt}
\renewcommand\arraystretch{1.15}
\begin{tabular}{c p{0.78\linewidth}}
\hline
\# & Instruction \\
\hline
1  & pick up the orange bowl \\
2  & lift up the dirty bowl \\
3  & pick up the bowl on the left \\
4  & pick up the empty bowl \\
5  & pick up the dirty container \\
6  & pick up the bowl with object inside \\
7  & pick up the left bowl \\
8  & pick up the bowl that is pink \\
9  & pick up the bowl that is pink \\
10 & pick up the bowl further \\
11 & pick up the bowl nearer to the camera \\
12 & pick up the right bowl \\
13 & pick up the bowl without tissue \\
14 & pick up the bowl with tissue \\
15 & pick up the bowl that is dirty \\
\hline
\end{tabular}
\caption{\textbf{Open-ended Language Instructions.} These are the collected open-ended instructions from 10 participants, where they were only allowed to make changes to verbs, nouns, or adjectives from the ground-truth instructions (i.e, "<verb> the <adj.> <noun.>").}
\label{tab:open_instruc}
\end{table}

% \begin{table}[!htbp]
% \centering
% \caption{Open-ended Language Instructions}
% \label{tab:open_instruc}
% \footnotesize
% \setlength\tabcolsep{6pt}
% \renewcommand\arraystretch{1.15}
% \begin{tabular}{\linewidth}{c p{0.78\linewidth}}
% \toprule
% \# & Instruction \\
% \midrule
% 1  & pick up the orange bowl \\
% 2  & lift up the dirty bowl \\
% 3  & pick up the bowl on the left \\
% 4  & pick up the empty bowl \\
% 5  & pick up the dirty container \\
% 6  & pick up the bowl with object inside \\
% 7  & pick up the left bowl \\
% 8  & pick up the bowl that is pink \\
% 9  & pick up the bowl that is pink \\
% 10 & pick up the bowl further \\
% 11 & pick up the bowl nearer to the camera \\
% 12 & pick up the right bowl \\
% 13 & pick up the bowl without tissue \\
% 14 & pick up the bowl with tissue \\
% 15 & pick up the bowl that is dirty \\
% \bottomrule
% \end{tabular}
% \end{table}

\begin{table}[!htbp]
\centering
\footnotesize
\setlength\tabcolsep{4pt}
\renewcommand\arraystretch{1.1}

\begin{tabular}{l*{3}{>{\centering\arraybackslash}p{2.8cm}}}
\toprule
\multicolumn{1}{c}{} &
\multicolumn{3}{c}{\cellcolor{olive!5}\textbf{Task Name}} \\[-0.2em]
\cmidrule(lr){2-4}
\textbf{Parameter} & \texttt{put\_bowl\_in\_sink} & \texttt{wipe\_table} & \texttt{table\_bussing} \\
\midrule
% Learning Rate      & \multicolumn{3}{c}{5e-4} \\
Steps              & 9K & 7K & 5K \\
Global Batch Size  & \multicolumn{3}{c}{64} \\
GPUs (H100s)       & \multicolumn{3}{c}{32} \\
Time (Hours)       & 5 & 4 & 3 \\
GPU Hours          & 160 & 128 & 96 \\
Multi-task Training     & \multicolumn{3}{c}{No} \\
Input Images            & \multicolumn{3}{c}{1 Third-person + 1 Wrist-mounted} \\
Image Size              & \multicolumn{3}{c}{640$\times$320 px (Resized to 320$\times$240 px)} \\
DoF                     & \multicolumn{3}{c}{7 (3 Translations + 3 Rotations + 1 Gripper State)} \\
Observation History     & \multicolumn{3}{c}{No (Single-step Inputs)} \\
Use Proprioception & \multicolumn{3}{c}{No} \\
Action Chunk Size       & \multicolumn{3}{c}{8 Steps (Naive Action Chunking with Close-loop Prediction)} \\
% LoRA Rank               & \multicolumn{3}{c}{32} \\
% LoRA Alpha              & \multicolumn{3}{c}{16} \\
% LoRA Dropout            & \multicolumn{3}{c}{0} \\
% LoRA Bias               & \multicolumn{3}{c}{None} \\
\# Trainable Params     & \multicolumn{3}{c}{97M LoRA adapter} \\
Image Augmentations     &
\multicolumn{3}{c}{%
  \parbox[t]{8.4cm}{\raggedright
\texttt{import torchvision.transforms as T}\\
\texttt{transform = T.Compose([}\\
\texttt{\hspace*{1em}T.RandomResizedCrop(size=(height, width), scale=(0.9, 0.9), ratio=(width/height, width/height)),}\\
\texttt{\hspace*{1em}T.Resize((height, width)),}\\
\texttt{\hspace*{1em}T.ColorJitter(}\\
\texttt{\hspace*{2em}brightness=0.2,}\\
\texttt{\hspace*{2em}contrast=(0.8, 1.2),}\\
\texttt{\hspace*{2em}saturation=(0.8, 1.2),}\\
\texttt{\hspace*{2em}hue=0.05}\\
\texttt{\hspace*{1em}),}\\
\texttt{])}
}} \\
\addlinespace[4pt]  
\bottomrule
\end{tabular}
\caption{\textbf{\molmoact's Post-training Hyperparameters for In-distribution Single-arm Tasks.} We specify the hyperparameters for \molmoact post-training. Note that we conduct all our post-training experiments on \molmoactd, with a fixed learning rate of 5e-4, LoRA rank of 32, LoRA alpha of 16, LoRA dropout of 0, and no LoRA bias.}
\label{tab:hyperparams_real_indist_singlearm}
\end{table}
\begin{table}[!htbp]
\centering
\footnotesize
\setlength\tabcolsep{4pt}
\renewcommand\arraystretch{1.1}

\begin{tabular}{l*{3}{>{\centering\arraybackslash}p{2.8cm}}}
\toprule
\multicolumn{1}{c}{} &
\multicolumn{3}{c}{\cellcolor{olive!5}\textbf{Task Name}} \\[-0.2em]
\cmidrule(lr){2-4}
\textbf{Parameter} & \texttt{set\_table} & \texttt{lift\_tray} & \texttt{fold\_towel} \\
\midrule
% Learning Rate      & \multicolumn{3}{c}{5e-4} \\
Steps              & 9K & 6K & 7K \\
Global Batch Size  & \multicolumn{3}{c}{64} \\
GPUs (H100s)       & \multicolumn{3}{c}{32} \\
Time (Hours)       & 5 & 3 & 4 \\
GPU Hours          & 160 & 96 & 128 \\
Multi-task Training     & \multicolumn{3}{c}{No} \\
Input Images            & \multicolumn{3}{c}{1 Third-person + 2 Wrist-mounted} \\
Image Size              & \multicolumn{3}{c}{640$\times$320 px (Resized to 320$\times$240 px)} \\
DoF                     & \multicolumn{3}{c}{14 (6 Translations + 6 Rotations + 2 Gripper States)} \\
Observation History     & \multicolumn{3}{c}{No (Single-step Inputs)} \\
Use Proprioception & \multicolumn{3}{c}{No} \\
Action Chunk Size       & \multicolumn{3}{c}{8 Steps (Naive Action Chunking with Close-loop Prediction)} \\
% LoRA Rank               & \multicolumn{3}{c}{32} \\
% LoRA Alpha              & \multicolumn{3}{c}{16} \\
% LoRA Dropout            & \multicolumn{3}{c}{0} \\
% LoRA Bias               & \multicolumn{3}{c}{None} \\
\# Trainable Params     & \multicolumn{3}{c}{97M LoRA adapter} \\
Image Augmentations     &
\multicolumn{3}{c}{%
  \parbox[t]{8.4cm}{\raggedright
\texttt{import torchvision.transforms as T}\\
\texttt{transform = T.Compose([}\\
\texttt{\hspace*{1em}T.RandomResizedCrop(size=(height, width), scale=(0.9, 0.9), ratio=(width/height, width/height)),}\\
\texttt{\hspace*{1em}T.Resize((height, width)),}\\
\texttt{\hspace*{1em}T.ColorJitter(}\\
\texttt{\hspace*{2em}brightness=0.2,}\\
\texttt{\hspace*{2em}contrast=(0.8, 1.2),}\\
\texttt{\hspace*{2em}saturation=(0.8, 1.2),}\\
\texttt{\hspace*{2em}hue=0.05}\\
\texttt{\hspace*{1em}),}\\
\texttt{])}
}} \\
\addlinespace[4pt]  
\bottomrule
\end{tabular}
\caption{\textbf{\molmoact's Post-training Hyperparameters for In-distribution Bimanual Tasks.} We specify the hyperparameters for \molmoact post-training. Note that we conduct all our post-training experiments on \molmoactd, with a fixed learning rate of 5e-4, LoRA rank of 32, LoRA alpha of 16, LoRA dropout of 0, and no LoRA bias.}
\label{tab:hyperparams_real_indist_bimanual}
\end{table}
\begin{table}[!htbp]
\centering
\footnotesize
\setlength\tabcolsep{4pt}
\renewcommand\arraystretch{1.1}

\begin{tabular}{l*{1}{>{\centering\arraybackslash}p{8.4cm}}}
\toprule
\multicolumn{1}{c}{} &
\multicolumn{1}{c}{\cellcolor{olive!5}\textbf{Task Name}} \\[-0.2em]
\cmidrule(lr){2-2}
\textbf{Parameter} & \texttt{put\_(green\_can/red\_cup/banana)\_in\_(yellow/blue)\_plate} \\
\midrule
% Learning Rate      & \multicolumn{1}{c}{5e-4} \\
Steps              & 44K \\
Global Batch Size  & \multicolumn{1}{c}{64} \\
GPUs (H100s)       & \multicolumn{1}{c}{32} \\
Time (Hours)       & 23 \\
GPU Hours          & 736 \\
Multi-task Training     & \multicolumn{1}{c}{Yes} \\
Input Images            & \multicolumn{1}{c}{1 Third-person + 1 Wrist-mounted} \\
Image Size              & \multicolumn{1}{c}{640$\times$320 px (Resized to 320$\times$240 px)} \\
DoF                     & \multicolumn{1}{c}{7 (3 Translations + 3 Rotations + 1 Gripper State)} \\
Observation History     & \multicolumn{1}{c}{No (Single-step Inputs)} \\
Use Proprioception & \multicolumn{1}{c}{No} \\
Action Chunk Size       & \multicolumn{1}{c}{8 Steps (Naive Action Chunking with Close-loop Prediction)} \\
% LoRA Rank               & \multicolumn{1}{c}{32} \\
% LoRA Alpha              & \multicolumn{1}{c}{16} \\
% LoRA Dropout            & \multicolumn{1}{c}{0} \\
% LoRA Bias               & \multicolumn{1}{c}{None} \\
\# Trainable Params     & \multicolumn{1}{c}{97M LoRA adapter} \\
Image Augmentations     &
\multicolumn{1}{c}{%
  \parbox[t]{8.4cm}{\raggedright
\texttt{import torchvision.transforms as T}\\
\texttt{transform = T.Compose([}\\
\texttt{\hspace*{1em}T.RandomResizedCrop(size=(height, width), scale=(0.9, 0.9), ratio=(width/height, width/height)),}\\
\texttt{\hspace*{1em}T.Resize((height, width)),}\\
\texttt{\hspace*{1em}T.ColorJitter(}\\
\texttt{\hspace*{2em}brightness=0.2,}\\
\texttt{\hspace*{2em}contrast=(0.8, 1.2),}\\
\texttt{\hspace*{2em}saturation=(0.8, 1.2),}\\
\texttt{\hspace*{2em}hue=0.05}\\
\texttt{\hspace*{1em}),}\\
\texttt{])}
}} \\
\addlinespace[4pt]  
\bottomrule
\end{tabular}
\caption{\textbf{\molmoact's Post-training Hyperparameters for Out-of-distribution Single-arm Tasks.} We specify the hyperparameters for \molmoact post-training. Note that we conduct all our post-training experiments on \molmoactd, with a fixed learning rate of 5e-4, LoRA rank of 32, LoRA alpha of 16, LoRA dropout of 0, and no LoRA bias.}
\label{tab:hyperparams_real_outdist}
\end{table}
\begin{table}[!htbp]
\centering
\footnotesize
\setlength\tabcolsep{4pt}
\renewcommand\arraystretch{1.1}

\begin{tabular}{l*{3}{>{\centering\arraybackslash}p{2.8cm}}}
\toprule
\multicolumn{1}{c}{} &
\multicolumn{3}{c}{\cellcolor{olive!5}\textbf{Task Name}} \\[-0.2em]
\cmidrule(lr){2-4}
\textbf{Parameter} & \texttt{close\_lip} & \texttt{rotate\_pot} & \texttt{pour\_tea} \\
\midrule
% Learning Rate      & \multicolumn{3}{c}{5e-4} \\
Steps              & 8K & 6K & 15K \\
Global Batch Size  & \multicolumn{3}{c}{64} \\
GPUs (H100s)       & \multicolumn{3}{c}{32} \\
Time (Hours)       & 4 & 3 & 8 \\
GPU Hours          & 128 & 96 & 256 \\
Multi-task Training     & \multicolumn{3}{c}{No} \\
Input Images            & \multicolumn{3}{c}{1 Third-person + 1 Wrist-mounted} \\
Image Size              & \multicolumn{3}{c}{640$\times$320 px (Resized to 320$\times$240 px)} \\
DoF                     & \multicolumn{3}{c}{7 (3 Translations + 3 Rotations + 1 Gripper State)} \\
Observation History     & \multicolumn{3}{c}{No (Single-step Inputs)} \\
Use Proprioception & \multicolumn{3}{c}{No} \\
Action Chunk Size       & \multicolumn{3}{c}{8 Steps (Naive Action Chunking with Close-loop Prediction)} \\
% LoRA Rank               & \multicolumn{3}{c}{32} \\
% LoRA Alpha              & \multicolumn{3}{c}{16} \\
% LoRA Dropout            & \multicolumn{3}{c}{0} \\
% LoRA Bias               & \multicolumn{3}{c}{None} \\
\# Trainable Params     & \multicolumn{3}{c}{97M LoRA adapter} \\
Image Augmentations     &
\multicolumn{3}{c}{%
  \parbox[t]{8.4cm}{\raggedright
\texttt{import torchvision.transforms as T}\\
\texttt{transform = T.Compose([}\\
\texttt{\hspace*{1em}T.RandomResizedCrop(size=(height, width), scale=(0.9, 0.9), ratio=(width/height, width/height)),}\\
\texttt{\hspace*{1em}T.Resize((height, width)),}\\
\texttt{\hspace*{1em}T.ColorJitter(}\\
\texttt{\hspace*{2em}brightness=0.2,}\\
\texttt{\hspace*{2em}contrast=(0.8, 1.2),}\\
\texttt{\hspace*{2em}saturation=(0.8, 1.2),}\\
\texttt{\hspace*{2em}hue=0.05}\\
\texttt{\hspace*{1em}),}\\
\texttt{])}
}} \\
\addlinespace[4pt]  
\bottomrule
\end{tabular}
\caption{\textbf{\molmoact's Post-training Hyperparameters for Evaluation on \molmoactdata.} We specify the hyperparameters for \molmoact post-training. Note that we conduct all our post-training experiments on \molmoactd, with a fixed learning rate of 5e-4, LoRA rank of 32, LoRA alpha of 16, LoRA dropout of 0, and no LoRA bias.}
\label{tab:hyperparams_real_ablation}
\end{table}
\begin{table}[!htbp]
\centering
\footnotesize
\setlength\tabcolsep{4pt}
\renewcommand\arraystretch{1.1}

\begin{tabular}{l*{1}{>{\centering\arraybackslash}p{8.4cm}}}
\toprule
\multicolumn{1}{c}{} &
\multicolumn{1}{c}{\cellcolor{olive!5}\textbf{Task Name}} \\[-0.2em]
\cmidrule(lr){2-2}
\textbf{Parameter} & \texttt{pick\_up\_bowl} \\
\midrule
% Learning Rate      & \multicolumn{1}{c}{5e-4} \\
Steps              & 11K \\
Global Batch Size  & \multicolumn{1}{c}{64} \\
GPUs (H100s)       & \multicolumn{1}{c}{32} \\
Time (Hours)       & 6 \\
GPU Hours          & 192 \\
Multi-task Training     & \multicolumn{1}{c}{No} \\
Input Images            & \multicolumn{1}{c}{1 Third-person + 1 Wrist-mounted} \\
Image Size              & \multicolumn{1}{c}{640$\times$320 px (Resized to 320$\times$240 px)} \\
DoF                     & \multicolumn{1}{c}{7 (3 Translations + 3 Rotations + 1 Gripper State)} \\
Observation History     & \multicolumn{1}{c}{No (Single-step Inputs)} \\
Use Proprioception & \multicolumn{1}{c}{No} \\
Action Chunk Size       & \multicolumn{1}{c}{8 Steps (Naive Action Chunking with Close-loop Prediction)} \\
% LoRA Rank               & \multicolumn{1}{c}{32} \\
% LoRA Alpha              & \multicolumn{1}{c}{16} \\
% LoRA Dropout            & \multicolumn{1}{c}{0} \\
% LoRA Bias               & \multicolumn{1}{c}{None} \\
\# Trainable Params     & \multicolumn{1}{c}{97M LoRA adapter} \\
Image Augmentations     &
\multicolumn{1}{c}{%
  \parbox[t]{8.4cm}{\raggedright
\texttt{import torchvision.transforms as T}\\
\texttt{transform = T.Compose([}\\
\texttt{\hspace*{1em}T.RandomResizedCrop(size=(height, width), scale=(0.9, 0.9), ratio=(width/height, width/height)),}\\
\texttt{\hspace*{1em}T.Resize((height, width)),}\\
\texttt{\hspace*{1em}T.ColorJitter(}\\
\texttt{\hspace*{2em}brightness=0.2,}\\
\texttt{\hspace*{2em}contrast=(0.8, 1.2),}\\
\texttt{\hspace*{2em}saturation=(0.8, 1.2),}\\
\texttt{\hspace*{2em}hue=0.05}\\
\texttt{\hspace*{1em}),}\\
\texttt{])}
}} \\
\addlinespace[4pt]  
\bottomrule
\end{tabular}
\caption{\textbf{\molmoact's Post-training Hyperparameters for Steerability Evaluation.} We specify the hyperparameters for \molmoact post-training. Note that we conduct all our post-training experiments on \molmoactd, with a fixed learning rate of 5e-4, LoRA rank of 32, LoRA alpha of 16, LoRA dropout of 0, and no LoRA bias.}
\label{tab:hyperparams_real_steer}
\end{table}

\begin{table}[t]
\centering
\scriptsize
\setlength{\tabcolsep}{4pt}
\renewcommand{\arraystretch}{1.1}
\begin{tabular}{|p{0.15\linewidth}|p{0.05\linewidth}|p{0.12\linewidth}|p{0.12\linewidth}|p{0.10\linewidth}|}
\hline
\textbf{Task} & \textbf{Trial} & \textbf{MolmoAct} & \textbf{\pizerofast} & \textbf{OpenVLA} \\
\hline
\multirow{26}{*}{Fold Towel} 
& 0  & 0.25 & 0.25 & 0.25 \\ \cline{2-5}
& 1  & 1.00 & 0.50 & 0.25 \\ \cline{2-5}
& 2  & 1.00 & 0.25 & 0.25 \\ \cline{2-5}
& 3  & 1.00 & 1.00 & 0.50 \\ \cline{2-5}
& 4  & 1.00 & 0.25 & 0.25 \\ \cline{2-5}
& 5  & 1.00 & 0.25 & 0.25 \\ \cline{2-5}
& 6  & 1.00 & 0.25 & 1.00 \\ \cline{2-5}
& 7  & 1.00 & 0.25 & 0.25 \\ \cline{2-5}
& 8  & 0.25 & 0.25 & 0.25 \\ \cline{2-5}
& 9  & 1.00 & 1.00 & 0.25 \\ \cline{2-5}
& 10 & 1.00 & 1.00 & 0.25 \\ \cline{2-5}
& 11 & 1.00 & 1.00 & 0.25 \\ \cline{2-5}
& 12 & 1.00 & 1.00 & 0.25 \\ \cline{2-5}
& 13 & 0.75 & 0.25 & 0.25 \\ \cline{2-5}
& 14 & 0.25 & 0.25 & 0.25 \\ \cline{2-5}
& 15 & 0.25 & 0.00 & 0.25 \\ \cline{2-5}
& 16 & 1.00 & 0.25 & 0.25 \\ \cline{2-5}
& 17 & 1.00 & 1.00 & 0.25 \\ \cline{2-5}
& 18 & 0.25 & 0.50 & 0.25 \\ \cline{2-5}
& 19 & 0.75 & 1.00 & 0.25 \\ \cline{2-5}
& 20 & 0.25 & 0.25 & 0.25 \\ \cline{2-5}
& 21 & 1.00 & 0.25 & 0.25 \\ \cline{2-5}
& 22 & 1.00 & 0.25 & 0.25 \\ \cline{2-5}
& 23 & 1.00 & 0.75 & 0.25 \\ \cline{2-5}
& 24 & 1.00 & 1.00 & 1.00 \\ \hline
\multicolumn{2}{|c|}{\textbf{Average}} & \textbf{0.80} & \textbf{0.52} & \textbf{0.32} \\ \hline
\end{tabular}
\caption{Detailed per-trial performance for \textit{Fold Towel} for Bimanual tasks. Each row shows the task progress score for a specific trial.}
\label{tab:eval_per_trial_fold_towel}
\end{table}

\begin{table}[t]
\centering
\scriptsize
\setlength{\tabcolsep}{4pt}
\renewcommand{\arraystretch}{1.1}
\begin{tabular}{|p{0.15\linewidth}|p{0.05\linewidth}|p{0.12\linewidth}|p{0.12\linewidth}|p{0.10\linewidth}|}
\hline
\textbf{Task} & \textbf{Trial} & \textbf{MolmoAct} & \textbf{\pizerofast} & \textbf{OpenVLA} \\
\hline
\multirow{25}{*}{Lift Tray} 
& 0  & 1.00 & 1.00 & 1.00 \\ \cline{2-5}
& 1  & 1.00 & 0.00 & 1.00 \\ \cline{2-5}
& 2  & 1.00 & 0.60 & 1.00 \\ \cline{2-5}
& 3  & 1.00 & 1.00 & 1.00 \\ \cline{2-5}
& 4  & 1.00 & 0.60 & 1.00 \\ \cline{2-5}
& 5  & 1.00 & 0.00 & 1.00 \\ \cline{2-5}
& 6  & 1.00 & 1.00 & 1.00 \\ \cline{2-5}
& 7  & 1.00 & 1.00 & 1.00 \\ \cline{2-5}
& 8  & 1.00 & 1.00 & 1.00 \\ \cline{2-5}
& 9  & 1.00 & 1.00 & 1.00 \\ \cline{2-5}
& 10 & 1.00 & 1.00 & 1.00 \\ \cline{2-5}
& 11 & 1.00 & 1.00 & 1.00 \\ \cline{2-5}
& 12 & 1.00 & 1.00 & 1.00 \\ \cline{2-5}
& 13 & 1.00 & 1.00 & 1.00 \\ \cline{2-5}
& 14 & 1.00 & 1.00 & 1.00 \\ \cline{2-5}
& 15 & 1.00 & 0.00 & 1.00 \\ \cline{2-5}
& 16 & 1.00 & 1.00 & 1.00 \\ \cline{2-5}
& 17 & 1.00 & 1.00 & 1.00 \\ \cline{2-5}
& 18 & 1.00 & 0.00 & 1.00 \\ \cline{2-5}
& 19 & 1.00 & 1.00 & 1.00 \\ \cline{2-5}
& 20 & 1.00 & 1.00 & 1.00 \\ \cline{2-5}
& 21 & 1.00 & 0.00 & 1.00 \\ \cline{2-5}
& 22 & 1.00 & 0.30 & 1.00 \\ \cline{2-5}
& 23 & 1.00 & 1.00 & 1.00 \\ \cline{2-5}
& 24 & 1.00 & 1.00 & 1.00 \\ \hline
\multicolumn{2}{|c|}{\textbf{Average}} & \textbf{1.00} & \textbf{0.74} & \textbf{1.00} \\ \hline
\end{tabular}
\caption{Detailed per-trial performance for \textit{Lift Tray} for Bimanual tasks. Each row shows the task progress score for a specific trial.}
\label{tab:eval_per_trial_lift_tray}
\end{table}

\begin{table}[t]
\centering
\scriptsize
\setlength{\tabcolsep}{4pt}
\renewcommand{\arraystretch}{1.1}
\begin{tabular}{|p{0.15\linewidth}|p{0.05\linewidth}|p{0.12\linewidth}|p{0.12\linewidth}|p{0.10\linewidth}|}
\hline
\textbf{Task} & \textbf{Trial} & \textbf{MolmoAct} & \textbf{\pizerofast} & \textbf{OpenVLA} \\
\hline
\multirow{26}{*}{Set up Table} 
& 0  & 1.00 & 0.50 & 0.25 \\ \cline{2-5}
& 1  & 1.00 & 0.00 & 0.25 \\ \cline{2-5}
& 2  & 0.25 & 0.25 & 0.00 \\ \cline{2-5}
& 3  & 1.00 & 0.50 & 0.00 \\ \cline{2-5}
& 4  & 1.00 & 0.25 & 0.25 \\ \cline{2-5}
& 5  & 1.00 & 0.25 & 0.75 \\ \cline{2-5}
& 6  & 1.00 & 0.00 & 0.25 \\ \cline{2-5}
& 7  & 1.00 & 0.25 & 0.25 \\ \cline{2-5}
& 8  & 0.75 & 0.00 & 0.25 \\ \cline{2-5}
& 9  & 0.25 & 0.00 & 1.00 \\ \cline{2-5}
& 10 & 1.00 & 0.25 & 0.25 \\ \cline{2-5}
& 11 & 0.75 & 0.25 & 0.25 \\ \cline{2-5}
& 12 & 1.00 & 0.75 & 0.25 \\ \cline{2-5}
& 13 & 0.75 & 0.25 & 0.25 \\ \cline{2-5}
& 14 & 1.00 & 0.25 & 0.25 \\ \cline{2-5}
& 15 & 0.25 & 0.50 & 1.00 \\ \cline{2-5}
& 16 & 0.00 & 0.25 & 0.25 \\ \cline{2-5}
& 17 & 1.00 & 0.25 & 0.25 \\ \cline{2-5}
& 18 & 1.00 & 0.00 & 0.25 \\ \cline{2-5}
& 19 & 0.25 & 0.50 & 0.25 \\ \cline{2-5}
& 20 & 1.00 & 0.00 & 0.25 \\ \cline{2-5}
& 21 & 1.00 & 0.50 & 0.00 \\ \cline{2-5}
& 22 & 0.50 & 0.00 & 0.25 \\ \cline{2-5}
& 23 & 0.50 & 0.00 & 0.25 \\ \cline{2-5}
& 24 & 1.00 & 0.25 & 0.25 \\ \cline{2-5}
& \multicolumn{1}{c|}{\textbf{Average}} & \textbf{0.77} & \textbf{0.24} & \textbf{0.30} \\ \hline
\end{tabular}
\caption{Detailed per-trial performance for \textit{Set up Table} for Bimanual tasks. Each row shows the task progress score for a specific trial.}
\label{tab:eval_per_trial_setup_table}
\end{table}

\begin{table}[t]
\centering
\scriptsize
\setlength{\tabcolsep}{4pt}
\renewcommand{\arraystretch}{1.1}
\begin{tabular}{|p{0.18\linewidth}|p{0.05\linewidth}|p{0.12\linewidth}|p{0.12\linewidth}|p{0.10\linewidth}|}
\hline
\textbf{Task} & \textbf{Trial} & \textbf{MolmoAct} & \textbf{\pizerofast} & \textbf{OpenVLA} \\
\hline
\multirow{26}{*}{Put bowl in the sink} 
& 0  & 1.00 & 1.00 & 0.25 \\ \cline{2-5}
& 1  & 1.00 & 1.00 & 0.25 \\ \cline{2-5}
& 2  & 1.00 & 0.00 & 0.25 \\ \cline{2-5}
& 3  & 0.25 & 1.00 & 0.25 \\ \cline{2-5}
& 4  & 1.00 & 1.00 & 0.25 \\ \cline{2-5}
& 5  & 1.00 & 0.40 & 0.25 \\ \cline{2-5}
& 6  & 1.00 & 1.00 & 0.25 \\ \cline{2-5}
& 7  & 0.25 & 0.25 & 0.25 \\ \cline{2-5}
& 8  & 1.00 & 0.25 & 0.25 \\ \cline{2-5}
& 9  & 0.40 & 0.40 & 0.25 \\ \cline{2-5}
& 10 & 1.00 & 1.00 & 0.25 \\ \cline{2-5}
& 11 & 0.25 & 1.00 & 0.25 \\ \cline{2-5}
& 12 & 1.00 & 1.00 & 0.25 \\ \cline{2-5}
& 13 & 1.00 & 0.25 & 0.25 \\ \cline{2-5}
& 14 & 1.00 & 0.25 & 0.25 \\ \cline{2-5}
& 15 & 1.00 & 0.75 & 0.25 \\ \cline{2-5}
& 16 & 1.00 & 1.00 & 0.25 \\ \cline{2-5}
& 17 & 1.00 & 1.00 & 0.25 \\ \cline{2-5}
& 18 & 0.25 & 0.40 & 0.25 \\ \cline{2-5}
& 19 & 1.00 & 0.75 & 0.25 \\ \cline{2-5}
& 20 & 0.25 & 0.25 & 0.25 \\ \cline{2-5}
& 21 & 1.00 & 1.00 & 0.25 \\ \cline{2-5}
& 22 & 1.00 & 1.00 & 0.25 \\ \cline{2-5}
& 23 & 1.00 & 1.00 & 0.25 \\ \cline{2-5}
& 24 & 1.00 & 0.75 & 0.25 \\ \cline{2-5}
& \multicolumn{1}{c|}{\textbf{Average}} & \textbf{0.826} & \textbf{0.708} & \textbf{0.25} \\ \hline
\end{tabular}
\caption{Detailed per-trial performance for \textit{Put bowl in the sink} for Single arm tasks. Each row shows the task progress score for a specific trial.}
\label{tab:eval_per_trial_put_bowl}
\end{table}

\begin{table}[t]
\centering
\scriptsize
\setlength{\tabcolsep}{4pt}
\renewcommand{\arraystretch}{1.1}
\begin{tabular}{|p{0.18\linewidth}|p{0.05\linewidth}|p{0.12\linewidth}|p{0.12\linewidth}|p{0.10\linewidth}|}
\hline
\textbf{Task} & \textbf{Trial} & \textbf{MolmoAct} & \textbf{\pizerofast} & \textbf{OpenVLA} \\
\hline
\multirow{26}{*}{Wipe Table} 
& 0  & 1.00 & 1.00 & 0.25 \\ \cline{2-5}
& 1  & 1.00 & 0.50 & 0.25 \\ \cline{2-5}
& 2  & 1.00 & 1.00 & 0.25 \\ \cline{2-5}
& 3  & 1.00 & 1.00 & 0.25 \\ \cline{2-5}
& 4  & 1.00 & 1.00 & 0.25 \\ \cline{2-5}
& 5  & 1.00 & 1.00 & 0.25 \\ \cline{2-5}
& 6  & 1.00 & 0.25 & 0.25 \\ \cline{2-5}
& 7  & 1.00 & 1.00 & 0.25 \\ \cline{2-5}
& 8  & 1.00 & 1.00 & 0.25 \\ \cline{2-5}
& 9  & 1.00 & 1.00 & 0.25 \\ \cline{2-5}
& 10 & 1.00 & 1.00 & 0.25 \\ \cline{2-5}
& 11 & 1.00 & 1.00 & 0.25 \\ \cline{2-5}
& 12 & 1.00 & 1.00 & 0.25 \\ \cline{2-5}
& 13 & 1.00 & 0.50 & 0.25 \\ \cline{2-5}
& 14 & 1.00 & 1.00 & 0.25 \\ \cline{2-5}
& 15 & 1.00 & 1.00 & 0.25 \\ \cline{2-5}
& 16 & 1.00 & 1.00 & 0.25 \\ \cline{2-5}
& 17 & 1.00 & 1.00 & 0.25 \\ \cline{2-5}
& 18 & 1.00 & 1.00 & 0.25 \\ \cline{2-5}
& 19 & 1.00 & 1.00 & 1.00 \\ \cline{2-5}
& 20 & 1.00 & 1.00 & 0.25 \\ \cline{2-5}
& 21 & 1.00 & 1.00 & 0.25 \\ \cline{2-5}
& 22 & 1.00 & 1.00 & 0.25 \\ \cline{2-5}
& 23 & 1.00 & 1.00 & 0.25 \\ \cline{2-5}
& 24 & 1.00 & 1.00 & 0.25 \\ \cline{2-5}
& \multicolumn{1}{c|}{\textbf{Average}} & \textbf{1.000} & \textbf{0.817} & \textbf{0.265} \\ \hline
\end{tabular}
\caption{Detailed per-trial performance for \textit{Wipe Table} for Single arm tasks. Each row shows the task progress score for a specific trial.}
\label{tab:eval_per_trial_wipe_table}
\end{table}

\begin{table}[t]
\centering
\scriptsize
\setlength{\tabcolsep}{4pt}
\renewcommand{\arraystretch}{1.1}
\begin{tabular}{|p{0.18\linewidth}|p{0.05\linewidth}|p{0.12\linewidth}|p{0.12\linewidth}|p{0.10\linewidth}|}
\hline
\textbf{Task} & \textbf{Trial} & \textbf{MolmoAct} & \textbf{\pizerofast} & \textbf{OpenVLA} \\
\hline
\multirow{26}{*}{Clean the table} 
& 0  & 1.00 & 1.00 & 0.50 \\ \cline{2-5}
& 1  & 1.00 & 1.00 & 1.00 \\ \cline{2-5}
& 2  & 0.50 & 1.00 & 0.50 \\ \cline{2-5}
& 3  & 1.00 & 0.25 & 0.50 \\ \cline{2-5}
& 4  & 1.00 & 0.75 & 0.00 \\ \cline{2-5}
& 5  & 0.50 & 1.00 & 0.75 \\ \cline{2-5}
& 6  & 1.00 & 0.75 & 0.50 \\ \cline{2-5}
& 7  & 1.00 & 0.50 & 0.50 \\ \cline{2-5}
& 8  & 1.00 & 1.00 & 0.50 \\ \cline{2-5}
& 9  & 1.00 & 1.00 & 0.75 \\ \cline{2-5}
& 10 & 1.00 & 1.00 & 0.50 \\ \cline{2-5}
& 11 & 1.00 & 1.00 & 0.75 \\ \cline{2-5}
& 12 & 0.50 & 0.25 & 0.25 \\ \cline{2-5}
& 13 & 1.00 & 1.00 & 0.50 \\ \cline{2-5}
& 14 & 0.50 & 1.00 & 1.00 \\ \cline{2-5}
& 15 & 0.50 & 1.00 & 0.75 \\ \cline{2-5}
& 16 & 1.00 & 0.25 & 0.75 \\ \cline{2-5}
& 17 & 1.00 & 1.00 & 0.25 \\ \cline{2-5}
& 18 & 1.00 & 1.00 & 0.25 \\ \cline{2-5}
& 19 & 0.50 & 0.50 & 0.25 \\ \cline{2-5}
& 20 & 0.50 & 1.00 & 0.25 \\ \cline{2-5}
& 21 & 1.00 & 1.00 & 1.00 \\ \cline{2-5}
& 22 & 1.00 & 1.00 & 0.25 \\ \cline{2-5}
& 23 & 1.00 & 1.00 & 0.25 \\ \cline{2-5}
& 24 & 0.50 & 1.00 & 0.75 \\ \cline{2-5}
& \multicolumn{1}{c|}{\textbf{Average}} & \textbf{0.84} & \textbf{0.85} & \textbf{0.53} \\ \hline
\end{tabular}
\caption{Detailed per-trial performance for \textit{Clean the table} for Single arm tasks. Each row shows the task progress score for a specific trial.}
\label{tab:eval_per_trial_clean_table}
\end{table}

\begin{table}[t]
\centering
\scriptsize
\setlength{\tabcolsep}{4pt}
\renewcommand{\arraystretch}{1.1}
\begin{tabular}{|p{0.26\linewidth}|p{0.30\linewidth}|p{0.10\linewidth}|p{0.10\linewidth}|p{0.10\linewidth}|}
\hline
\textbf{Category} & \textbf{Task} & \textbf{OpenVLA} & \textbf{\pizerofast} & \textbf{MolmoAct} \\
\hline
In Distribution & put the green can into the yellow plate & 0.375 & 0.8125 & 1.0 \\ \hline
In Distribution & put the red cup into the yellow plate & 0.5 & 0.5 & 0.625 \\ \hline
In Distribution & put the banana into the blue plate & 0.25 & 0.625 & 0.75 \\ \hline
Language Variation & put the green tea into the yellow plate & 0.375 & 0.8125 & 0.625 \\ \hline
Language Variatiion & put the fruit into the blue plate & 0.0 & 0.0625 & 0.625 \\ \hline
Language Variatiion & put the red cylinder into the yellow plate & 0.3125 & 0.0 & 0.75 \\ \hline
Spatial Variation & put the green can into the yellow plate & 0.4375 & 0.5625 & 0.625 \\ \hline
Spatial Variation & put the red cup into the yellow plate & 0.5 & 0.375 & 0.4375 \\ \hline
Spatial Variation & put the banana into the blue plate & 0.25 & 0.4375 & 0.5625 \\ \hline
Distractor (Coke Can, Sponge) & put the green can into the yellow plate & 0.125 & 0.875 & 0.9375 \\ \hline
Distractor (Coke Can, Sponge) & put the red cup into the yellow plate & 0.5 & 0.3125 & 0.6875 \\ \hline
Distractor (Coke Can, Sponge) & put the banana into the blue plate & 0.25 & 0.4375 & 0.625 \\ \hline
Novel Object & put the sponge into the yellow plate & 0.25 & 0.0 & 0.875 \\ \hline
Novel Object & put the coke can into the yellow plate & 0.375 & 0.5625 & 0.625 \\ \hline
Novel Object & put the bowl into the yellow plate & 0.25 & 0.3125 & 0.4375 \\ \hline
\end{tabular}
\caption{Detailed results of real-world evaluation. The first column indicates the variation category while the second column presents the language instruction. For each task, the detailed task progress score used to evaluate each model are detailed at section \ref{supp:eval:generalize}}
\label{tab:eval_generalization_detail}
\end{table}

\begin{table}[t]
\centering
\scriptsize
\setlength{\tabcolsep}{4pt}
\renewcommand{\arraystretch}{1.1}
\begin{tabular}{|p{0.15\linewidth}|p{0.05\linewidth}|p{0.12\linewidth}|p{0.18\linewidth}|p{0.12\linewidth}|p{0.10\linewidth}|}
\hline
\textbf{Task} & \textbf{Trial} & \textbf{MolmoAct} & \textbf{MolmoAct (W/o MolmoAct Data)} & \textbf{\pizerofast} & \textbf{OpenVLA} \\
\hline
\multirow{11}{*}{Pour Tea} 
& 0 & 0.8 & 0.5 & 0.5 & 1.0 \\ \cline{2-6}
& 1 & 0.8 & 0.8 & 0.0 & 0.5 \\ \cline{2-6}
& 2 & 0.5 & 0.5 & 0.0 & 0.0 \\ \cline{2-6}
& 3 & 1.0 & 1.0 & 0.0 & 0.5 \\ \cline{2-6}
& 4 & 1.0 & 1.0 & 0.8 & 0.5 \\ \cline{2-6}
& 5 & 0.8 & 0.5 & 1.0 & 0.5 \\ \cline{2-6}
& 6 & 1.0 & 1.0 & 1.0 & 0.5 \\ \cline{2-6}
& 7 & 0.5 & 0.5 & 1.0 & 0.0 \\ \cline{2-6}
& 8 & 0.5 & 1.0 & 0.0 & 0.0 \\ \cline{2-6}
& 9 & 1.0 & 0.5 & 0.0 & 0.5 \\ \cline{2-6}
& 10 & 0.8 & 0.8 & 0.0 & 0.5 \\ \hline

\multirow{11}{*}{Close Lid} 
& 0 & 0.5 & 0.0 & 0.5 & 0.0 \\ \cline{2-6}
& 1 & 0.5 & 0.5 & 0.0 & 0.5 \\ \cline{2-6}
& 2 & 0.5 & 0.0 & 0.5 & 1.0 \\ \cline{2-6}
& 3 & 0.5 & 0.5 & 0.5 & 0.5 \\ \cline{2-6}
& 4 & 0.5 & 0.0 & 1.0 & 0.0 \\ \cline{2-6}
& 5 & 1.0 & 0.5 & 0.5 & 0.0 \\ \cline{2-6}
& 6 & 0.5 & 0.0 & 0.0 & 0.5 \\ \cline{2-6}
& 7 & 0.5 & 1.0 & 0.0 & 0.0 \\ \cline{2-6}
& 8 & 0.5 & 1.0 & 1.0 & 0.0 \\ \cline{2-6}
& 9 & 0.0 & 1.0 & 0.5 & 0.5 \\ \cline{2-6}
& 10 & 0.5 & 0.0 & 0.5 & 0.5 \\ \hline

\multirow{11}{*}{Rotate Pot} 
& 0 & 1.0 & 1.0 & 0.6 & 1.0 \\ \cline{2-6}
& 1 & 0.6 & 1.0 & 1.0 & 1.0 \\ \cline{2-6}
& 2 & 1.0 & 1.0 & 0.0 & 1.0 \\ \cline{2-6}
& 3 & 1.0 & 1.0 & 1.0 & 1.0 \\ \cline{2-6}
& 4 & 1.0 & 1.0 & 0.6 & 1.0 \\ \cline{2-6}
& 5 & 1.0 & 1.0 & 1.0 & 1.0 \\ \cline{2-6}
& 6 & 1.0 & 1.0 & 0.6 & 1.0 \\ \cline{2-6}
& 7 & 0.6 & 1.0 & 1.0 & 1.0 \\ \cline{2-6}
& 8 & 1.0 & 1.0 & 1.0 & 1.0 \\ \cline{2-6}
& 9 & 1.0 & 1.0 & 1.0 & 1.0 \\ \cline{2-6}
& 10 & 1.0 & 0.0 & 0.6 & 1.0 \\ \hline
\end{tabular}
\caption{Detailed per-trial performance for three tasks (\textit{Pour Tea}, \textit{Close Lid}, and \textit{Rotate Pot}). Each row shows the task progress score for a specific trial.}
\label{tab:eval_per_trial_molmoactdata}
\end{table}

\begin{table}[t]
\centering
\scriptsize
\setlength{\tabcolsep}{4pt}
\renewcommand{\arraystretch}{1.1}
\begin{tabular}{|p{0.22\linewidth}|p{0.19\linewidth}|p{0.08\linewidth}|p{0.16\linewidth}|p{0.16\linewidth}|p{0.12\linewidth}|}
\hline
\textbf{Task} & \textbf{Task Detail} & \textbf{Episode} & \textbf{Open instruction (MolmoAct)} & \textbf{Open instruction (\pizerofast)} & \textbf{Visual Trace (MolmoAct)} \\
\hline
pick up the orange bowl & steer from dirty to clean & 0 & 0.00 & 0.50 & 1.00 \\ \hline
lift up the dirty bowl & steer from clean to dirty & 1 & 1.00 & 0.00 & 1.00 \\ \hline
pick up the bowl on the left & steer from clean to dirty & 2 & 0.00 & 0.00 & 1.00 \\ \hline
pick up the empty bowl & steer from dirty to clean & 3 & 0.85 & 0.50 & 0.85 \\ \hline
pick up the dirty container & steer from clean to dirty & 4 & 0.50 & 0.00 & 1.00 \\ \hline
pick up the bowl with object inside & steer from clean to dirty & 5 & 0.00 & 0.00 & 0.50 \\ \hline
pick up the left bowl & steer from clean to dirty & 6 & 0.00 & 0.50 & 0.50 \\ \hline
pick up the bowl that is pink & steer from clean to dirty & 7 & 0.00 & 0.00 & 0.00 \\ \hline
pick up the bowl that is pink & steer from clean to dirty & 8 & 0.50 & 0.00 & 1.00 \\ \hline
pick up the bowl further & steer from dirty to clean & 9 & 0.85 & 0.50 & 0.85 \\ \hline
pick up the bowl nearer to the camera & steer from dirty to clean & 10 & 0.50 & 0.00 & 1.00 \\ \hline
pick up the right bowl & steer from dirty to clean & 11 & 0.50 & 0.00 & 0.50 \\ \hline
pick up the bowl without tissue & steer from clean to dirty & 12 & 0.50 & 0.00 & 0.50 \\ \hline
pick up the bowl with tissue & steer from clean to dirty & 13 & 0.00 & 0.00 & 0.50 \\ \hline
pick up the bowl that is dirty & steer from clean to dirty & 14 & 1.00 & 0.00 & 1.00 \\ \hline
\end{tabular}
\caption{Per-episode evaluation results for bowl-picking tasks with different steering conditions. Scores indicate task progression for each model configuration.}
\label{tab:eval_bowl_tasks}
\end{table}

\section{Data Details}
\label{supp:data}

\subsection{\molmoactdata}

\molmoactdata has two external camera views and a single wrist camera view. In the home environment data, the camera view configuration may vary between tasks, whereas for the tabletop data it remains the same for all tasks. For each task, we first rank the two external camera views based on scene clarity (i.e., how well the robot and objects are visible) and whether the view is occluded by the robot during task execution. Based on this ranking, we label them as the \textit{primary} and \textit{secondary} camera views. All home environment data is recorded at 15 Hz, while all tabletop data is recorded at 20 Hz. The tabletop data additionally includes extrinsic and intrinsic camera calibration matrices for both external cameras available on . We list details of the tasks we collected for \molmoact Dataset in Table \ref{tab:home_task} and \ref{tab:tabletop_task}.

\begin{table}[t]
\centering
\begingroup
\setlength{\tabcolsep}{1pt}
\setlength{\arrayrulewidth}{0.1pt}
\renewcommand{\arraystretch}{1.1}
\scriptsize
\begin{tabular}{|p{0.11\linewidth}|p{0.26\linewidth}|p{0.34\linewidth}|p{0.25\linewidth}|}
\hline
\textbf{Scene} & \textbf{Task} & \textbf{Language Instruction} & \textbf{Object(s)} \\
\hline
\csvreader[
  separator=comma,
  head to column names,
  late after line=\\\hline
]{\detokenize{tables/appendix/MolmoAct_Dataset_home.csv}}{}%
{\csvcoliii & \texttt{\csvcoli} & \csvcolii & \csvcoliv}
\end{tabular}
\caption{Tasks details of \molmoactdata Home Environemnt including scene, task name, language instruction and all objects used for data collection.}
\label{tab:home_task}
\endgroup
\end{table}

\begin{table}[t]
\centering
\begingroup
\setlength{\tabcolsep}{1pt}
\setlength{\arrayrulewidth}{0.1pt}
\renewcommand{\arraystretch}{1.1}
\scriptsize
\begin{tabular}{|p{0.11\linewidth}|p{0.26\linewidth}|p{0.34\linewidth}|p{0.25\linewidth}|}
\hline
\textbf{Scene} & \textbf{Task} & \textbf{Language Instruction} & \textbf{Object(s)} \\
\hline
\csvreader[
  separator=comma,
  head to column names,
  late after line=\\\hline
]{\detokenize{tables/appendix/MolmoAct_Dataset_tabletop.csv}}{}%
{\csvcoliii & \texttt{\csvcoli} & \csvcolii & \csvcoliv}
\end{tabular}
\caption{Tasks details of \molmoactdata Tabletop Environemnt including scene, task name, language instruction and all objects used for data collection.}
\label{tab:tabletop_task}
\endgroup
\end{table}

% \begin{table}[t]
% \centering
% \begingroup
% \setlength{\tabcolsep}{2pt}        % was 4pt; tighter padding, same column widths
% \setlength{\arrayrulewidth}{0.3pt} % slightly thinner vertical lines
% \renewcommand{\arraystretch}{1.1}
% \scriptsize
% \begin{tabular}{|p{0.08\linewidth}|p{0.26\linewidth}|p{0.34\linewidth}|p{0.28\linewidth}|}
% \hline
% \textbf{Scene} & \textbf{Task} & \textbf{Language Instruction} & \textbf{Object(s)} \\
% \hline
% \csvreader[
%   separator=comma,
%   head to column names,
%   late after line=\\\hline
% ]{figures/Molmo-ACT_Dataset.csv}{}%
% {\csvcoliii & \texttt{\csvcoli} & \csvcolii & \csvcoliv}
% \end{tabular}
% \caption{Prompts from \texttt{figures/Molmo-ACT_Dataset.csv} grouped by scene order.}
% \label{tab:molmoact_task_detail}
% \endgroup
% \end{table}

\section{Data Examples}
\label{supp:dataexamples}

This section include \textbf{randomly selected} examples from \molmoact's Action Reasoning Data and Multimodal web data used in pre-training, as well as \molmoactdata used in mid-training, and demonstrations collected for post-training. Prompts are shown in bold and Visual Reasoning Trace are annotated with a yellow line.

\begin{figure*}[t]  
  \centering
  \includegraphics[width=\textwidth]{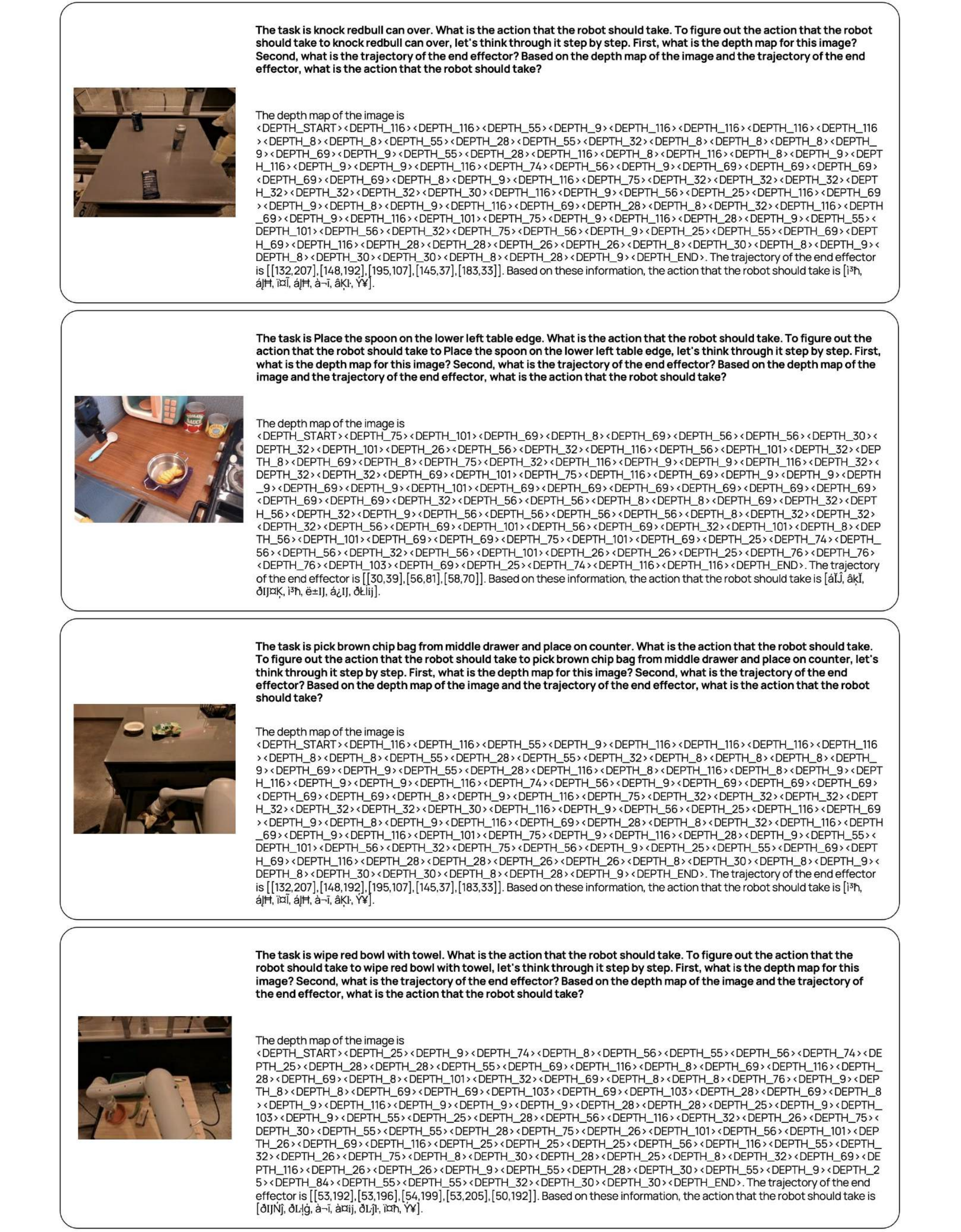}
  \caption{Randomly selected examples from \textbf{Action Reasoning Data} used in the pre-training stage.}

  \label{fig:appendix_action_reasoning_data_1}
\end{figure*}

\begin{figure*}[t]  
  \centering
  \includegraphics[width=\textwidth]{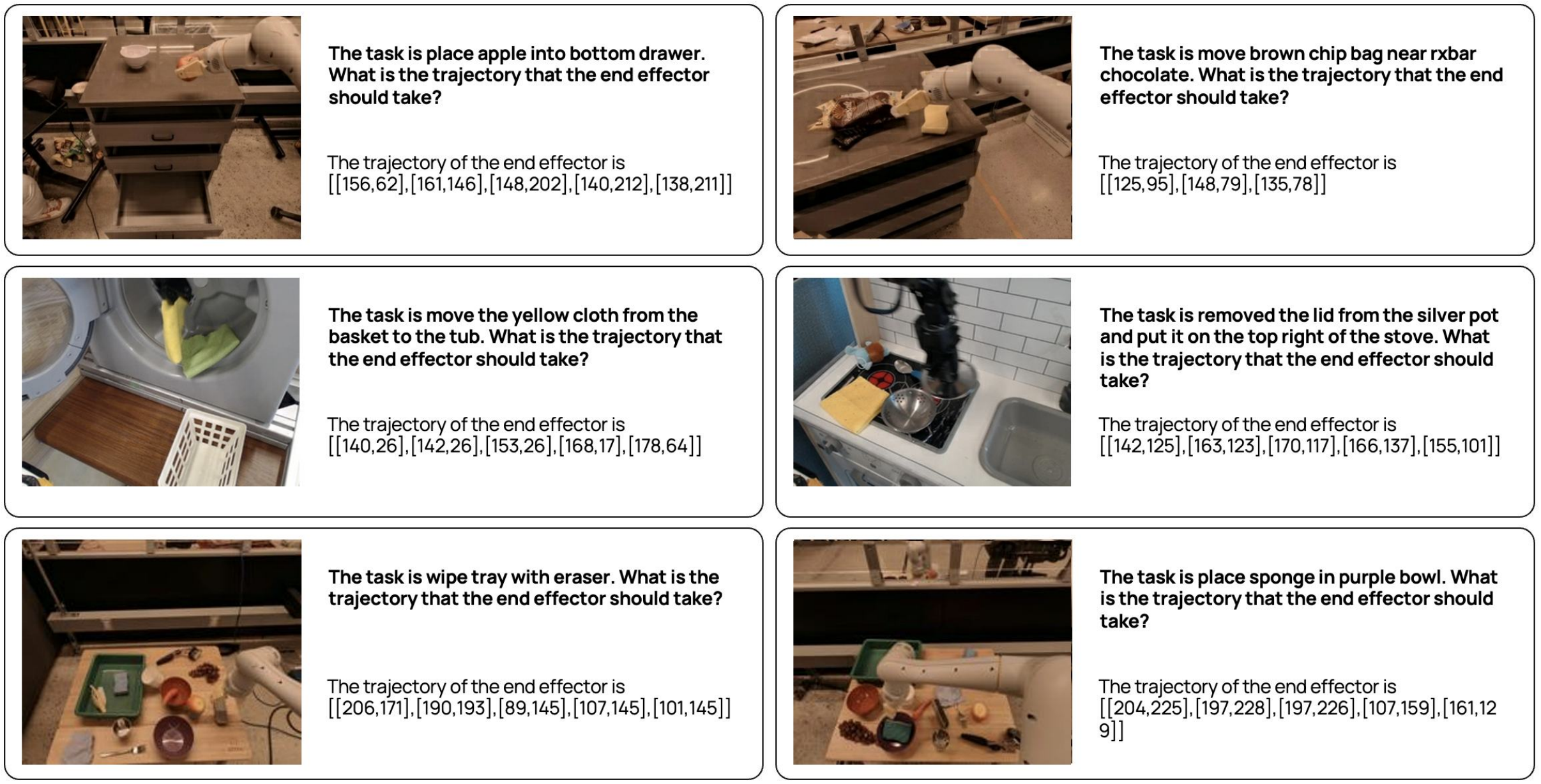}
  \caption{Randomly selected examples from \textbf{Auxiliary Visual Reasoning Trace} data used in the pre-training stage.}

  \label{fig:appendix_line}
\end{figure*}

\begin{figure*}[t]  
  \centering
  \includegraphics[width=\textwidth]{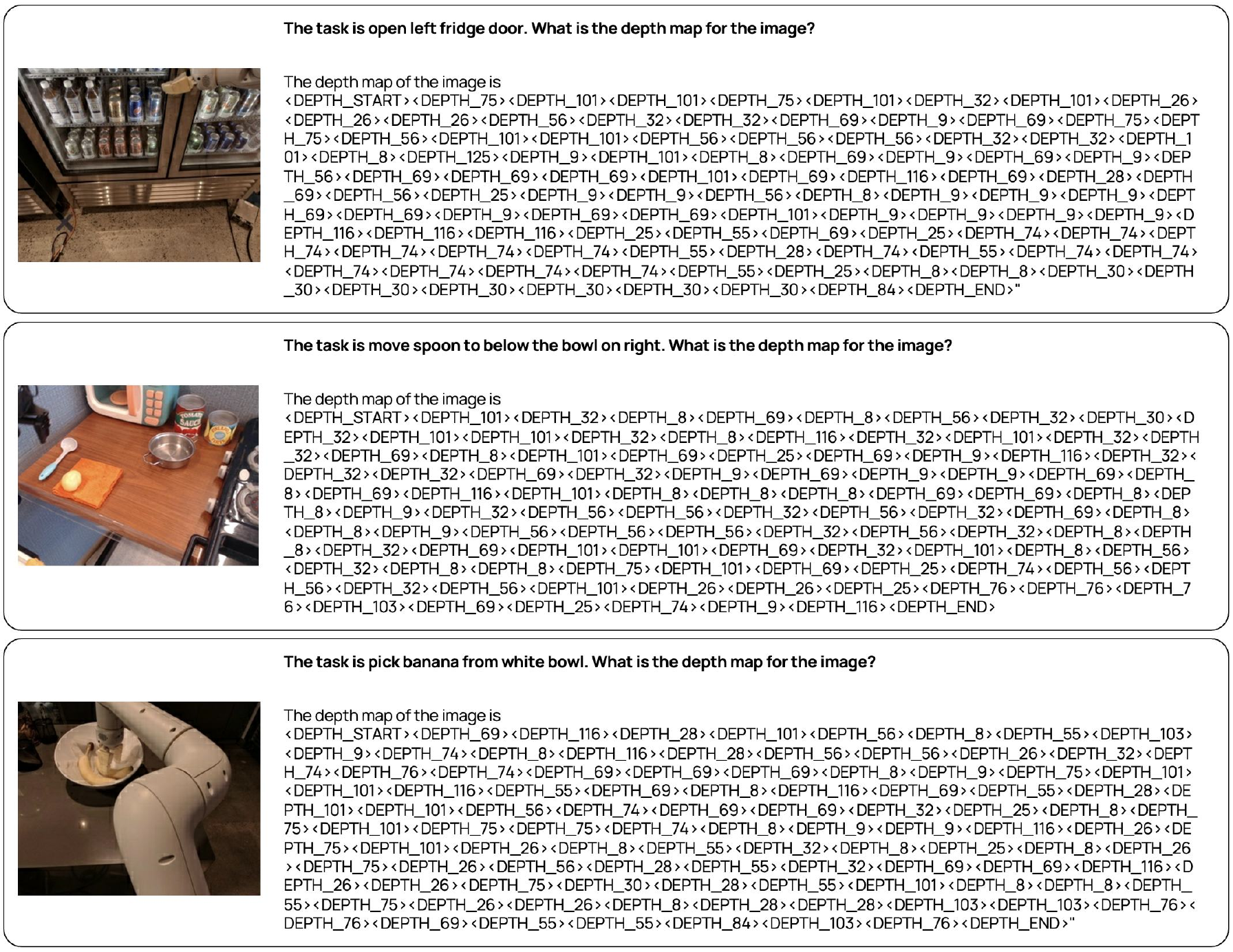}
  \caption{Randomly selected examples from \textbf{Auxiliary Depth Perception Tokens} data used in the pre-training stage.}

  \label{fig:appendix_depth}
\end{figure*}

\begin{figure*}[t]  
  \centering
  \includegraphics[width=\textwidth]{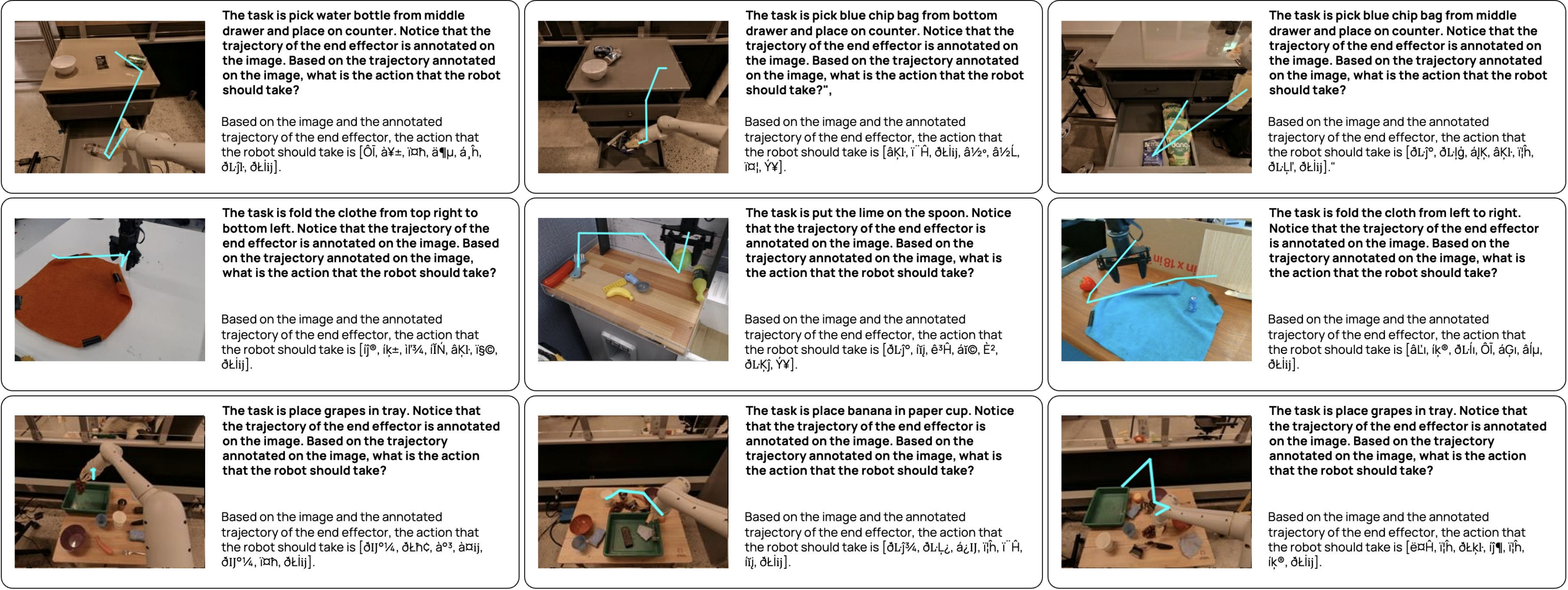}
  \caption{Randomly selected examples from \textbf{Trajectory-conditioned Action Data} used in the pre-training stage.}

  \label{fig:appendix_trajectory}
\end{figure*}

\begin{figure*}[t]  
  \centering
  \includegraphics[width=\textwidth]{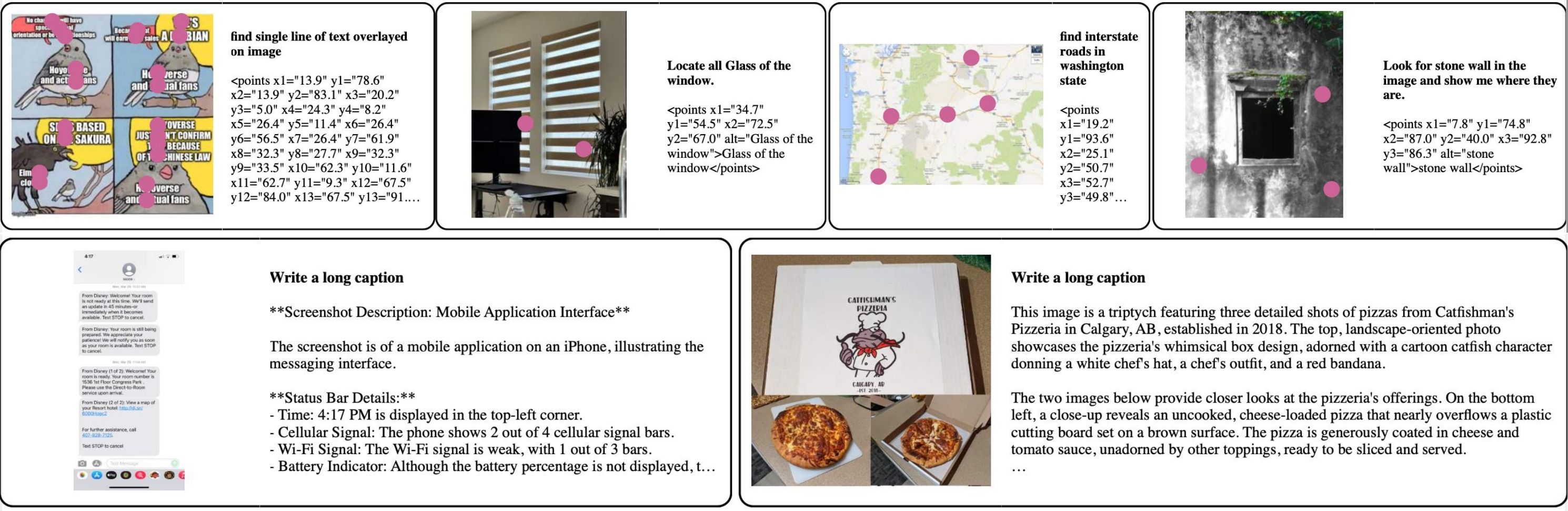}
  \caption{Randomly selected examples from \textbf{Multimodal Web Data} used in the pre-training stage.}

  \label{fig:appendix_vqa}
\end{figure*}

\begin{figure*}[t]  
  \centering
  \includegraphics[width=\textwidth]{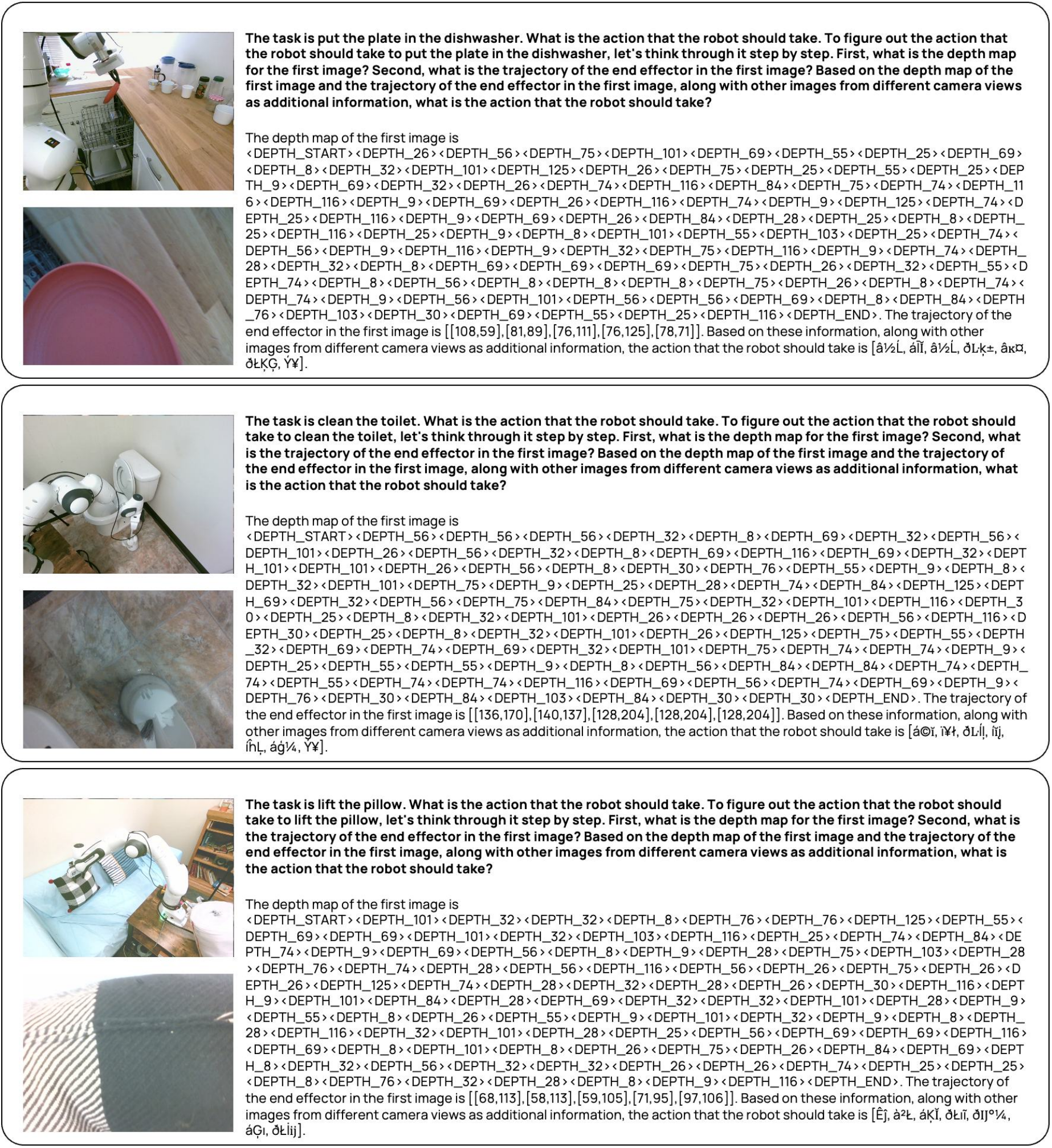}
  \caption{Randomly selected examples from \textbf{\molmoactdata (Home Environment)} used in the mid-training stage.}

  \label{fig:appendix_molmoactdataset_home}
\end{figure*}

\begin{figure*}[t]  
  \centering
  \includegraphics[width=\textwidth]{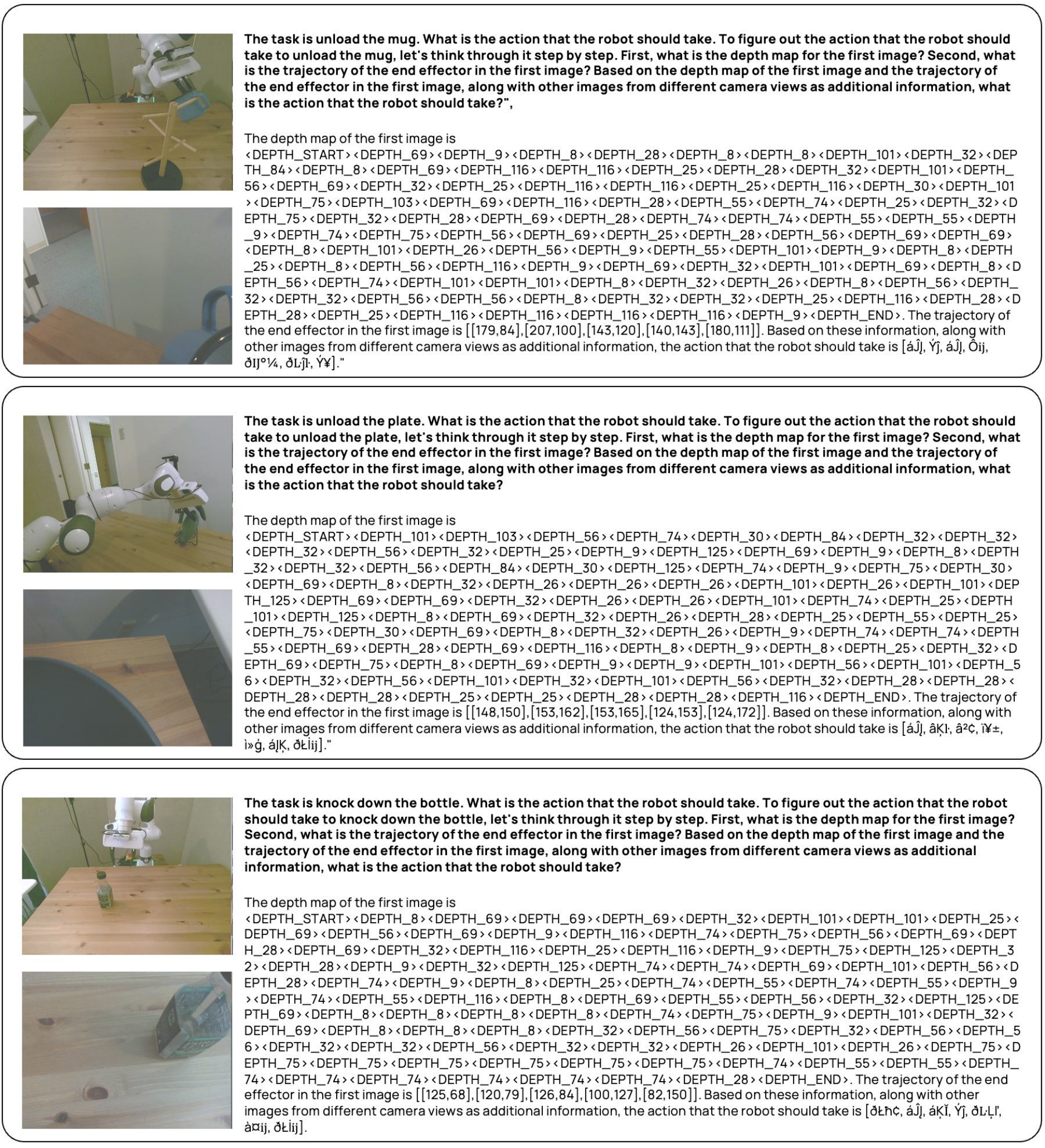}
  \caption{Randomly selected examples from \textbf{\molmoactdata (Tabletop Environment)} used in the mid-training stage.}

  \label{fig:appendix_molmoactdataset_tabletop}
\end{figure*}

\begin{figure*}[t]  
  \centering
  \includegraphics[width=\textwidth]{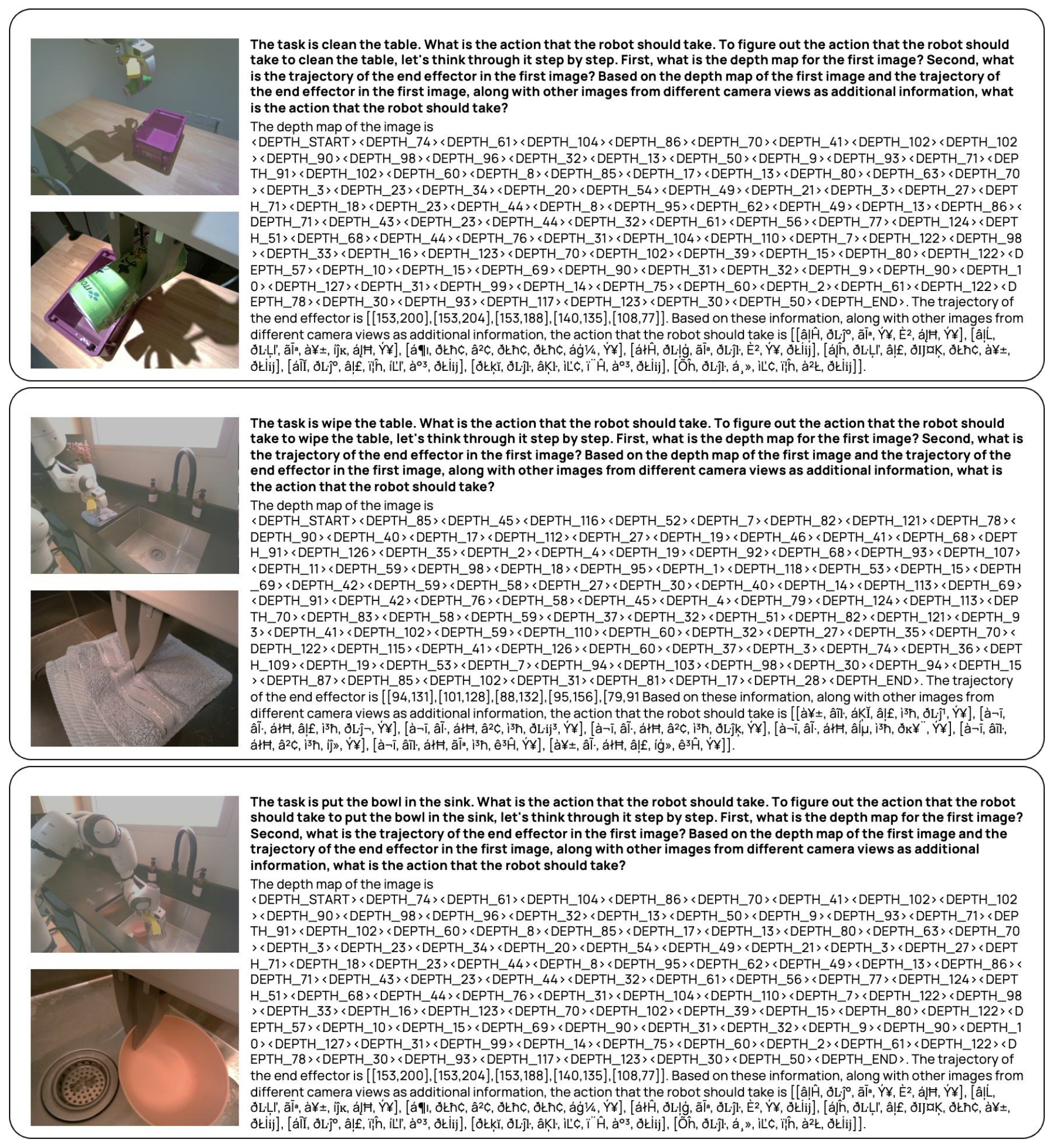}
  \caption{Randomly selected examples from \textbf{Single Arm Franka} demonstrations used in the post-training stage.}

  \label{fig:appendix_singlearm}
\end{figure*}

\begin{figure*}[t]  
  \centering
  \includegraphics[width=\textwidth]{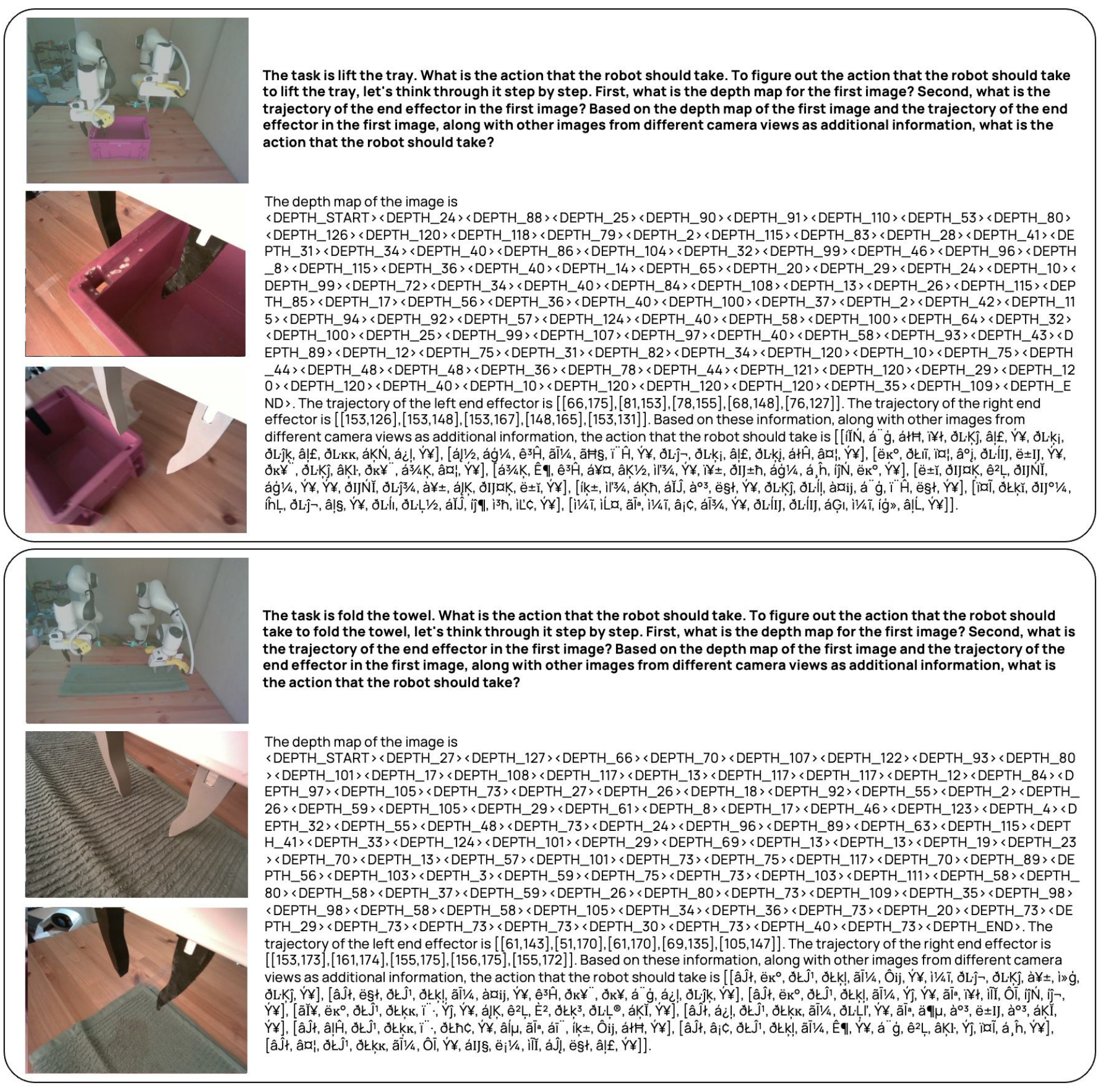}
  \caption{Randomly selected examples from \textbf{Bimanual Franka} demonstrations used in the post-training stage.}

  \label{fig:appendix_bimanual}
\end{figure*}

\begin{figure*}[t]  
  \centering
  \includegraphics[width=\textwidth]{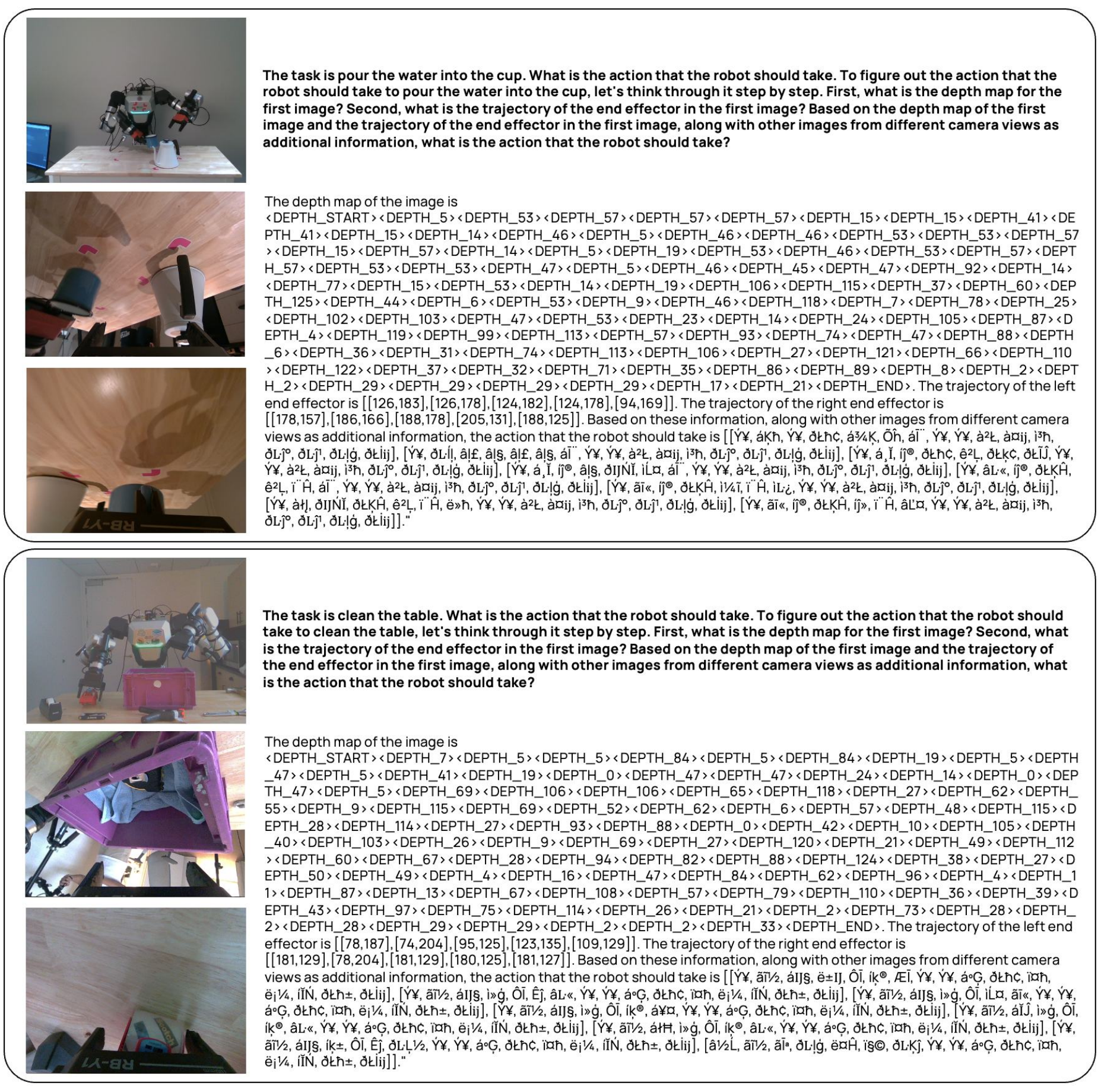}
  \caption{Randomly selected examples from \textbf{Rainbow} demonstrations used in the post-training stage.}

  \label{fig:appendix_rainbow}
\end{figure*}

\begin{itemize}
    \item \textbf{Action Reasoning Data - Figure \ref{fig:appendix_action_reasoning_data_1}}
    \item \textbf{Auxiliary Visual Reasoning Trace - Figure \ref{fig:appendix_line} }
    \item \textbf{Auxiliary Depth Perception Tokens - Figure \ref{fig:appendix_depth}}
    \item \textbf{Trajectory-conditioned Action Data - Figure \ref{fig:appendix_trajectory}}
    \item \textbf{Multimodal Web Data - Figure \ref{fig:appendix_vqa}}
    \item \textbf{\molmoactdata (Home Environment) - Figure \ref{fig:appendix_molmoactdataset_home}}
    \item \textbf{\molmoactdata (Tabletop) - Figure \ref{fig:appendix_molmoactdataset_tabletop}}
    \item \textbf{Post-Training Single Arm Franka - Figure \ref{fig:appendix_singlearm}}
    \item \textbf{Post-Training Bimanual Franka - Figure \ref{fig:appendix_bimanual}}
    \item \textbf{Post-Training Rainbow - Figure \ref{fig:appendix_rainbow}}
    
\end{itemize}

% \clearpage

\section{Limitations and Potential Solutions}
\label{supp:limit}

While \molmoact is all quite capable as a general-purpose action reasoning model, it is not without limitations. In the following sections, we discuss some of these limitations and potential solutions.

\textbf{Camera Occlusion of End-effector.} During post-training, \molmoact can process multiple camera views (e.g., front and wrist cameras), but its spatial reasoning primarily relies on the front camera, which typically provides a full view of the end-effector. This visibility is crucial for accurate visual reasoning trace prediction. However, if the end-effector is occluded in the front camera’s view, visual trace prediction—and thus overall performance—can degrade. A potential solution is to use a wide field-of-view camera (e.g., fisheye lens) and generate visual traces via SLAM, enabling temporal rather than purely spatial reasoning.

\textbf{Robustness of Steerability via Visual Traces.}
Robust action steerability relies on two factors: (i) precise yet diverse 2D visual traces during pre- and mid-training, and (ii) abundant, high-quality post-training data
\begin{itemize}
\item \textit{Trace Quality and Diversity.} For trajectory-conditioned action data, bounding-box–based detectors (e.g., Detectron) are problematic: predicted points collapse toward box centers, reducing spatial variation. They also require task-specific fine-tuning to localize robot grippers and transfer poorly across embodiments. In contrast, VLM-based point annotations (e.g., Molmo \citep{deitke2024molmo}, RoboPoint \citep{yuan2024robopoint}) yield accurate, non-degenerate traces and markedly improve steerability.
\item \textit{Coverage of Action Compositions.} To achieve steerability in real-world settings, post-training data should span as many action compositions as possible. Practically, this means inducing the robot to explore motion variants while still completing tasks, so the model learns rich correspondences between image-space traces and resulting actions.
\end{itemize}

\molmoact only learns to directly predict action simply based on the trace-overlaid image. So when we steer actions with a 2D visual trace, we are not leveraging the capability of \molmoact to perform action reasoning in space. Thus, we observe that this form of action steering still cannot enable the model to follow more complicated tasks. In particular, because the cue is purely 2D, the model lacks an explicit notion of depth: it often follows the intended path within the image plane (in-plane motion) but exhibits unintended or imprecise translation along the camera’s depth axis (out-of-plane). We hypothesize this could be mitigated by conditioning on—or reusing—the model’s predicted depth-perception tokens to lift the trace into 3D, which we leave for future exploration. Despite these limitations, our scheme demonstrates the feasibility of action steerability based on pure visual cues, and offers a simple, practical insight that the robotics community can build upon.

\textbf{Speed of Action Reasoning Model prediction.} Similar to many existing VLAs, our model exhibits a mismatch between its control inference frequency and the control frequency used during data collection. This gap may stem from server-to-robot communication latency and the additional time required to predict a larger number of reasoning tokens. Future work could explore techniques to reduce inference time, as seen in VLM optimization, or develop smaller parameter models optimized for efficient execution on edge or local devices.

\textbf{Precision in Depth Perception Token.} For depth perception token prediction, we follow \citep{bigverdi2025perception} and use a fixed set of 100 tokens to represent depth. However, fine-grained manipulation tasks require higher-resolution depth estimation. Increasing the number of depth perception tokens could enhance spatial reasoning and improve performance on such tasks.

\end{document}